%% file: main.tex
\documentclass[conference]{cls/IEEEtran}
\let\labelindent\relax
\usepackage{times}

\usepackage[numbers]{natbib}
\usepackage[table,usenames,dvipsnames]{xcolor}
\usepackage{enumitem}
\usepackage{amsmath,amssymb,amsfonts,amsthm,amscd,dsfont} % math
\usepackage{algorithm,algorithmicx,listings}        % algorithms
\usepackage[noend]{algpseudocode}			        % necessary for algorithmicx

\usepackage{graphicx,tabularx,adjustbox}
\usepackage{multicol}
\usepackage[font={small}]{caption}   %onehalfspacing
\usepackage{subcaption}

\usepackage[bookmarks=true, breaklinks=true, colorlinks, citecolor=Black, urlcolor=Violet,linkcolor=Black]{hyperref}

\newtheorem*{proposition*}{Proposition}

\newtheorem*{corollary*}{Corollary}

\theoremstyle{definition}

\newtheorem*{assumption*}{Assumption}
\newtheorem*{problem*}{Problem}
\newtheorem{problem}{Problem}
\theoremstyle{remark}

\newtheorem*{solution*}{Solution}

\newcommand{\prl}[1]{\left(#1\right)}
\newcommand{\brl}[1]{\left[#1\right]}

\DeclareMathOperator{\tr}{tr}

\input{cls/sym.tex}

\begin{document}

\title{Hamiltonian-based Neural ODE Networks on the SE(3) Manifold For Dynamics Learning and Control}

% You will get a Paper-ID when submitting a pdf file to the conference system
%\author{Thai Duong \and Nikolay Atanasov% <-this % stops a space
%\thanks{The authors are with the Department of Electrical and Computer Engineering, University of California, San Diego, La Jolla, CA 92093 USA {\tt\small \{tduong, natanasov\}@ucsd.edu}}%
%}

\author{\authorblockN{Thai Duong and Nikolay Atanasov}
\authorblockA{Department of Electrical and Computer Engineering\\
University of California, San Diego\\
La Jolla, CA 92093 USA\\
Email: \{tduong, natanasov\}@ucsd.edu}}
%\and
%\authorblockA{Twentieth Century Fox\\
%\and
%\authorblockN{James Kirk\\ and Montgomery Scott}
%\authorblockA{Starfleet Academy\\
%San Francisco, California 96678-2391\\

% avoiding spaces at the end of the author lines is not a problem with
% conference papers because we don't use \thanks or \IEEEmembership

%Homer Simpson\authorrefmark{2},
%Montgomery Scott\authorrefmark{3} and
%Eldon Tyrell\authorrefmark{4}}
%Telephone: (800) 555--1212, Fax: (888) 555--1212}
%\authorblockA{\authorrefmark{4}Tyrell Inc., 123 Replicant Street, Los Angeles, California 90210--4321}}

\maketitle

\begin{abstract}
Accurate models of robot dynamics are critical for safe and stable control and generalization to novel operational conditions. Hand-designed models, however, may be insufficiently accurate, even after careful parameter tuning. This motivates the use of machine learning techniques to approximate the robot dynamics over a training set of state-control trajectories. The dynamics of many robots, including ground, aerial, and underwater vehicles, are described in terms of their $SE(3)$ pose and generalized velocity, and satisfy conservation of energy principles. This paper proposes a Hamiltonian formulation over the $SE(3)$ manifold of the structure of a neural ordinary differential equation (ODE) network to approximate the dynamics of a rigid body. In contrast to a black-box ODE network, our formulation guarantees total energy conservation by construction. We develop energy shaping and damping injection control for the learned, potentially under-actuated $SE(3)$ Hamiltonian dynamics to enable a unified approach for stabilization and trajectory tracking with various platforms, including pendulum, rigid-body, and quadrotor systems.
\end{abstract}

% an energy-shaping controller. Our design relies on 

%\begin{abstract}
%An accurate dynamics model of a robotics system is crucial for planning, control of a robot. To improve accuracy, prior knowledge, e.g. state constraints and basic laws of physics, needs to be incorporated into learning models. For a wide range of robotics platforms ranging from pendulum to ground robots to unmanned aerial vehicles, the robot's pose is often considered as part of the state in the robot's dynamics and lie on the SE(3) manifold. As a physical system, the dynamics of such robots also has to satisfy the basic law of conservation of energy. In this paper, we propose a Hamiltonian-based dynamics learning framework that conserves the total energy of the system and guarantees state transitions on SE(3). Based on the learned Hamiltonian-based dynamics, we develop energy-based controllers for fully- and under-actuated systems (e.g. quadrotors) by energy shaping and damping injection. The effectiveness of the proposed learning framework and the energy-based controllers is verified on various platforms including pendulums, fully actuated rigid bodies (e.g. fully-actuated hexarotors), and under-actuated quadrotors. \TD{TODO: Talk a bit more about the experiments results after we finish that section.}
%\end{abstract}

\IEEEpeerreviewmaketitle

\input{tex/Introduction.tex}
\input{tex/RelatedWork.tex}

\input{tex/ProblemFormulation.tex}

\input{tex/Prelim.tex}

\input{tex/TechnicalApproach.tex}
\input{tex/ControllerDesign.tex}

%\input{tex/NikolaysAddon.tex}
\input{tex/ExperimentalResults.tex}
\input{tex/Conclusion.tex}

\input{tex/Appendix.tex}

\section*{Acknowledgments}
We gratefully acknowledge support from NSF RI IIS-2007141 and ARL DCIST CRA W911NF-17-2-0181.

%% Use plainnat to work nicely with natbib. 

%\newpage
%\bibliographystyle{plainnat}
%\bibliographystyle{unsrtnat}
\bibliographystyle{cls/abbrvplainnat}
\bibliography{bib/thai_ref}
%\newpage
%\input{tex/Appendix.tex}

\end{document}

%% file: cls/sym.tex
\newcommand{\calD}{{\cal D}}

\newcommand{\calG}{{\cal G}}
\newcommand{\calH}{{\cal H}}

\newcommand{\calJ}{{\cal J}}

\newcommand{\calL}{{\cal L}}

\newcommand{\calR}{{\cal R}}

\newcommand{\frakp}{{\mathfrak{p}}}
\newcommand{\frakq}{{\mathfrak{q}}}

\newcommand{\bfa}{\mathbf{a}}
\newcommand{\bfb}{\mathbf{b}}

\newcommand{\bfd}{\mathbf{d}}
\newcommand{\bfe}{\mathbf{e}}
\newcommand{\bff}{\mathbf{f}}
\newcommand{\bfg}{\mathbf{g}}

\newcommand{\bfp}{\mathbf{p}}
\newcommand{\bfq}{\mathbf{q}}
\newcommand{\bfr}{\mathbf{r}}
\newcommand{\bfs}{\mathbf{s}}

\newcommand{\bfu}{\mathbf{u}}
\newcommand{\bfv}{\mathbf{v}}
\newcommand{\bfw}{\mathbf{w}}
\newcommand{\bfx}{\mathbf{x}}

\newcommand{\bfzeta}{\boldsymbol{\zeta}}

\newcommand{\bftheta}{\boldsymbol{\theta}}

\newcommand{\bfpi}{\boldsymbol{\pi}}

\newcommand{\bftau}{\boldsymbol{\tau}}

\newcommand{\bfpsi}{\boldsymbol{\psi}}
\newcommand{\bfomega}{\boldsymbol{\omega}}

\newcommand{\bfI}{\mathbf{I}}
\newcommand{\bfJ}{\mathbf{J}}
\newcommand{\bfK}{\mathbf{K}}
\newcommand{\bfL}{\mathbf{L}}
\newcommand{\bfM}{\mathbf{M}}

\newcommand{\bfR}{\mathbf{R}}

\newcommand{\bfT}{\mathbf{T}}

\newcommand{\bbR}{\mathbb{R}}
\newcommand{\bbS}{\mathbb{S}}

%% file: tex/Introduction.tex
\section*{Supplementary Material}
Software and videos supplementing this paper:\\ 
\centerline{\url{https://thaipduong.github.io/SE3HamDL/}}
\section{Introduction}
\label{sec:intro}

Motion planning and optimal control algorithms depend on the availability of accurate system dynamics models \cite{LynchParkBook,mellinger2011minimum, loquercio2021autotune}. Models obtained from first principles and calibrated over a small set of parameters via system identification~\cite{ljung1999system} are widely used for unmanned ground vehicles (UGVs), unmanned aerial vehicles (UAVs), and unmanned underwater vehicles (UUVs). Such models often over-simplify or even incorrectly describe the underlying structure of the dynamical system, leading to bias and modeling errors that cannot be corrected by optimization over few parameters. Data-driven techniques \cite{nguyen2011model,deisenroth2011pilco,williams2017information, raissi2018multistep,chua2018deep} have emerged as a powerful approach to approximate the function describing the system dynamics with an over-parameterized machine learning model, trained over a dataset of system state and control trajectories. Neural networks are especially expressive function approximation models, capable of identifying and generalizing dynamics interaction patterns from the training data. Training neural network models, however, typically requires large amounts of data and computation time, which is impractical in robotics applications. Recent works \cite{lutter2019deeplagrangian,gupta2019general,cranmer2020lagrangian,greydanus2019hamiltonian,chen2019symplectic,roehrl2020modeling} have considered a hybrid approach to this problem, where prior knowledge of the physics, governing the system dynamics, is used to assist the learning process. The dynamics of physical systems obey kinematic constraints and energy conservation laws. While these laws are known to be universally true, a black-box machine learning model might struggle to infer them for the training data, causing poor generalization. Instead, prior knowledge may be encoded into the learning model, e.g., using a prior distribution \cite{deisenroth2011pilco}, a graph-network forward kinematic model \cite{sanchez2018graph}, or a network architecture reflecting the structure of Lagrangian \cite{lutter2019deeplagrangian} or Hamiltonian \cite{greydanus2019hamiltonian} mechanical systems. Moreover, many physical robot platforms are composed of rigid-body interconnections and their state evolution respects the position and orientation kinematics over the $SE(3)$ manifold \cite{LynchParkBook}. The goal of this paper is to incorporate both the $SE(3)$ kinematics and energy conservation constraints in the structure of the dynamics learning model. We only focus on learning the dynamics of a single rigid body but this is already sufficient to model many UGV, UAV, and UUV robots. We also aim to design a unified control approach that attempts to stabilize the learned model without depending on any prior knowledge of its parameters or level of under-actuation.

Lagrangian and Hamiltonian mechanics \cite{lurie2013analytical,HolmBook} provide physical system descriptions that can be integrated into the structure of a neural network \cite{greydanus2019hamiltonian, bertalan2019learning, chen2019symplectic, finzi2020simplifying, zhong2020symplectic, willard2020integrating}. Prior work, however, has only considered vector-valued states, when designing Lagrangian- or Hamiltonian-structured neural networks. This limits the applicability of these techniques as most interesting robot systems have states on the $SE(3)$ manifold. Hamiltonian equations of motion are available for orientation states but existing formulations rely predominantly on $3$ dimensional vector parametrizations, such as Euler angles \cite{maciejewski1985hamiltonian,shivarama2004hamilton}, which suffer from singularities. We design a neural ordinary differential equation (ODE) network \cite{chen2018neural}, whose structure captures Hamiltonian dynamics over the $SE(3)$ manifold \cite{lee2017global}. Our model guarantees by construction that its long-term trajectory predictions satisfy $SE(3)$ constraints and total energy conservation.  Inspired by \cite{zhong2020symplectic, zhong2020dissipative}, we model kinetic energy and potential energy by separate neural networks, each governed by a set of Hamiltonian equations on $SE(3)$. The Hamiltonian formulation can be generalized to a Port-Hamiltonian one \cite{van2014port}, enabling us to design an energy-based controller for trajectory tracking. In summary, this paper makes the following \textbf{contributions}.
\begin{itemize}

  \item We design a neural ODE model that respects the structure of Hamiltonian dynamics over the $SE(3)$ manifold to enable data-driven learning of rigid-body system dynamics.
  
  \item We develop a unified controller for (Port-)Hamiltonian $SE(3)$ dynamics that achieves trajectory tracking if permissible by the system's degree of underactuation.
  
  \item We demonstrate our dynamics learning and control approach for fully-actuated pendulum and rigid-body systems and an under-actuated quadrotor system.
  
\end{itemize}

%% file: tex/RelatedWork.tex
\section{Related Work}
\label{sec:related_work}

Physics-guided dynamics learning, in which prior knowledge about a physical system is integrated into the design of a machine learning model, has received significant attention recently \cite{willard2020integrating}. Models designed with structure respecting kinematic constraints \cite{sanchez2018graph}, symmetry \cite{ruthotto2019deep,wang2020incorporating}, Lagrangian mechanics \cite{roehrl2020modeling, lutter2019deeplagrangian,gupta2019general,cranmer2020lagrangian,lutter2019deepunderactuated} or Hamiltonian mechanics \cite{greydanus2019hamiltonian, bertalan2019learning, chen2019symplectic, finzi2020simplifying, zhong2020symplectic, willard2020integrating} guarantee that the laws of physics are satisfied by construction, regardless of the training data. Sanchez-Gonzalez et al. \cite{sanchez2018graph} design graph neural networks to represent the kinematic structure of complex dynamical systems and demonstrate forward model learning and online planning via gradient-based trajectory optimization. Ruthotto et al. \cite{ruthotto2019deep} propose a partial differential equation (PDE) interpretation of convolutional neural networks and derive new parabolic and hyperbolic ResNet architectures guided by PDE theory. Wang et al. \cite{wang2020incorporating} design symmetry equivariant neural network models, encoding rotation, scaling, and uniform motion, to learn physical dynamics that are robust to symmetry group distributional shifts. Lagrangian-based methods \cite{roehrl2020modeling, lutter2019deeplagrangian,gupta2019general,cranmer2020lagrangian,lutter2019deepunderactuated} design neural network models for physical systems, based on the Euler-Lagrange differential equations of motion \cite{lurie2013analytical,HolmBook}, in terms of generalized coordinates $\mathbf\frakq$, their velocity $\dot{\mathbf\frakq}$ and a Lagrangian function $\mathcal{L}(\mathbf\frakq, \dot{\mathbf\frakq})$, defined as the difference between the kinetic and potential energies. The energy terms are modeled by neural networks, either separately \cite{lutter2019deeplagrangian,lutter2019deepunderactuated} or together \cite{cranmer2020lagrangian}. Hamiltonian-based methods \cite{greydanus2019hamiltonian, bertalan2019learning, chen2019symplectic, finzi2020simplifying, zhong2020symplectic, willard2020integrating} use a Hamiltonian formulation \cite{lurie2013analytical,HolmBook} of the system dynamics, instead, in terms of generalized coordinates $\mathbf\frakq$, generalized momenta $\mathbf\frakp$, and a Hamiltonian function, $\mathcal{H}(\mathbf\frakq, \mathbf\frakp)$, representing the total energy of the system.
%This approach guarantees that the total energy is conserved if there is no external control input applied to the system. 
Greydanus et al. \cite{greydanus2019hamiltonian} model the Hamiltonian as a neural network and update its parameters by minimizing the discrepancy between its symplectic gradients and the time derivatives of the states $(\mathbf\frakq, \mathbf\frakp)$. This approach, however, requires that the state time derivatives are available in the training data set. Chen et al. \cite{chen2019symplectic} and Zhong et al. \cite{zhong2020symplectic} relax this assumption by using differentiable leapfrog integrators \cite{leimkuhler2004simulating} and differentiable ODE solvers \cite{chen2018neural}, respectively. The need for time derivatives of the states is eliminated by back-propagating a loss function measuring state discrepancy through the ODE solvers via the adjoint method. Toth et al. \cite{toth2019hamiltonian} show that, instead of from state trajectories, the Hamiltonian function can be learned from high-dimensional image observations. Finzi et al. \cite{finzi2020simplifying} show that using Cartesian coordinates with explicit constraints improves both the accuracy and data efficiency for the Lagrangian- and Hamiltonian-based approaches. In a recent closely related work, Zhong et al. \cite{zhong2020dissipative} showed that dissipating elements, such as friction or air drag, can be incorporated in a Hamiltonian-based neural ODE network by reformulating the system dynamics in Port-Hamiltonian form \cite{van2014port}.

% \cite{lee2017global}
In addition to learning the dynamics of a physical system from training data, this paper considers designing control methods for stabilization and trajectory tracking, relying only on the Hamiltonian dynamics structure rather than a particular system realization. The Hamiltonian formulation and its Port-Hamiltonian generalization \cite{van2014port} are built around the notion of system energy and, hence, are naturally related to control techniques for stabilization aiming to minimize the total energy. Since the minimum point of the Hamiltonian might not correspond to a desired regulation point, the control design needs to inject additional energy to ensure that the minimum of the total energy is at the desired equilibrium. For fully-actuated (Port-)Hamiltonian systems, it is sufficient to shape the potential energy only using an energy-shaping and damping-injection (ES-DI) controller \cite{van2014port}. For under-actuated systems, both the kinetic and potential energies needs to be shaped, e.g., via interconnection and damping assignment passivity-based control (IDA-PBC) \cite{van2014port,ortega2002stabilization,acosta2014robust,cieza2019ida}. Wang and Goldsmith \cite{wang2008modified} extend the IDA-PBC controller to trajectory tracking problem. Closely related to our controller design, Souza et. al. \cite{souza2014passivity} apply this technique to design controller for an underactuated quadrotor but use Euler angles as the orientation representation.

While existing Hamiltonian-based learning methods are designed for generalized coordinates in $\mathbb{R}^n$, we develop a neural ODE network for learning Hamiltonian dynamics over the $SE(3)$ manifold \cite{lee2017global}. %The reformulated Port-Hamiltonian dynamics also enables us to design energy-based controllers for the learned fully- and under-actuated systems such as quadrotors.
We connect Hamiltonian-dynamics learning with the idea of IDA-PBC control to allow stabilization of any rigid-body robot without relying on its model parameters a priori. We design the trajectory tracking controller for under-actuated systems based on the IDA-PBC approach and show how to construct desired pose and momentum trajectories given only a desired position trajectory. We demonstrate the tight integration of dynamics learning and control to achieve closed-loop trajectory tracking with fully- and under-actuated quadrotor robots.

%% file: tex/ProblemFormulation.tex
\section{Problem Statement}
\label{sec:problem_statement}

Consider a robot with state $\bfx$ consisting of its position $\bfp \in \mathbb{R}^3$, orientation $\bfR\in SO(3)$, body-frame linear velocity $\bfv \in \bbR^3$, and body-frame angular velocity $\bfomega\in \mathbb{R}^3$. Let $\dot{\bfx} = \bff(\bfx, \bfu)$ characterize the robot dynamics with control input $\bfu$. For example, the control input of an Ackermann-drive UGV may include its linear acceleration and steering angle rate, and that of a quadrotor UAV may include the total thrust and moment generated by the propellers.

We assume that the function $\bff$ specifying the robot dynamics is unknown and aim to approximate it using a dataset $\mathcal{D}$ of state and control trajectories. Specifically, let $\calD = \{t_{0:N}^{(i)}, \bfx^{(i)}_{0:N}, \bfu^{(i)}\}_{i=1}^D$ consist of $D$ state sequences $\bfx^{(i)}_{0:N}$, obtained by applying a constant control input $\bfu^{(i)}$ to the system with initial condition $\bfx^{(i)}_0$ at time $t_0^{(i)}$ and sampling its state $\bfx^{(i)}(t_n^{(i)}) =: \bfx^{(i)}_n$ at times $t_0^{(i)} < t_1^{(i)} < \ldots < t_N^{(i)}$. We aim to find a function $\bar{\bff}_{\bftheta}$ with parameters $\bftheta$ that approximates the true dynamics $\bff$ well on the dataset $\mathcal{D}$. To optimize the parameters $\bftheta$, we roll out the approximate dynamics $\bar{\bff}_{\bftheta}$ with initial state $\bfx_0^{(i)}$ and constant control $\bfu^{(i)}$ and minimize the discrepancy between the computed state sequence $\bar{\bfx}^{(i)}_{1:N}$ and the true state sequence $\bfx^{(i)}_{1:N}$ in $\calD$. The dynamics learning problem is summarized below.

\begin{problem}
\label{problem:dynamics_learning}
Given a dataset $\calD = \{t_{0:N}^{(i)}, \bfx^{(i)}_{0:N}, \bfu^{(i)}\}_{i=1}^D$ and a function $\bar{\bff}_{\bftheta}$, find the parameters $\bftheta$ that minimize:
\begin{equation}
\label{problem_formulation_unknown_env_equation}
\begin{aligned}
\min_{\bftheta} \;&\sum_{i = 1}^D \sum_{n = 1}^N \ell(\bfx^{(i)}_j,\bar{\bfx}^{(i)}_j)\\
\text{s.t.} \;\; & \dot{\bar{\bfx}}^{(i)}(t) = \bar{\bff}_{\bftheta}(\bar{\bfx}^{(i)}(t), \bfu^{(i)}), \;\;\bar{\bfx}^{(i)}(t_0) = \bfx^{(i)}_0,\\
& \bar{\bfx}^{(i)}_n = \bar{\bfx}^{(i)}(t_n), \;\;\forall n = 1, \ldots, N,\;\;\forall i = 1, \ldots, D,
\end{aligned}
\end{equation}
where $\ell$ is a distance metric on the state space.
\end{problem}

Further, we aim to design a feedback controller that attempts to track a desired state trajectory $\bfx^*(t)$, $t \geq t_0$, for any learned realization $\bar{\bff}_{\bftheta}$ of the robot dynamics.

\begin{problem}
\label{problem:setpoint_reg}
Given an initial condition $\bfx_0$ at time $t_0$, desired state trajectory $\bfx^*(t)$, $t \geq t_0$, and learned dynamics $\bar{\bff}_{\bftheta}$, design a feedback control law $\bfu = \bfpi(\bfx, \bftheta, \bfx^*(t))$ such that $\limsup_{t \to \infty} \ell(\bfx(t),\bfx^*(t))$ is bounded.
\end{problem}

In our setting, the robot kinematics need to evolve over the $SE(3)$ manifold and the dynamics $\bff(\bfx, \bfu)$ respect the law of energy conservation, when there is no control input, $\bfu = \bf0$. We embed these constraints in the structure of the parametric function $\bar{\bff}_{\bftheta}$. We review the $SE(3)$ kinematics and Hamiltonian dynamics equations next.

%% file: tex/Prelim.tex
\section{Preliminaries}
\label{sec:prelim}

\subsection{SE(3) Kinematics}
\label{subsec:SO3_SE3_kinematics}

Consider a fixed world inertial frame of reference $\{W\}$ and a rigid body with a body-fixed frame $\{B\}$ attached to its center of mass. The pose of $\{B\}$ in $\{W\}$ is determined by the position $\bfp = [x,y,z]^\top \in \bbR^3$ of the center of mass and the orientation of the coordinate axes of $\{B\}$:
\begin{equation}
\label{eq:rotmat_def}
\bfR = \begin{bmatrix}
\bfr_1 & \bfr_2 & \bfr_3
\end{bmatrix}^\top \in SO(3),
\end{equation}
where $\bfr_1, \bfr_2, \bfr_3 \in \mathbb{R}^3$ are the rows of the rotation matrix $\bfR$, which is an element of the special orthogonal group:
\begin{equation}
\label{eq:SO3_def}
SO(3) = \left\{\bfR \in \bbR^{3\times 3} : \bfR^\top \bfR = \bfI, \det(\bfR) = 1\right\}.
\end{equation}
The rigid-body position and orientation can be combined in a single pose matrix $\bfT \in SE(3)$, which is an element of the special Euclidean group:
\begin{equation}
\label{eq:SE3_def}
SE(3) = \left\{ \begin{bmatrix}
\bfR & \bfp \\
\bf0^\top & 1
\end{bmatrix} \in \mathbb{R}^{4\times 4}: \bfR \in \text{SO(3)}, \bfp \in \mathbb{R}^3\right\}.
\end{equation}
The kinematic equations of motion of the rigid body are determined by the linear velocity $\bfv \in \bbR^3$ and angular velocity $\bfomega \in \bbR^3$ of frame $\{B\}$ with respect to frame $\{W\}$, expressed in body-frame coordinates. The generalized velocity $\bfzeta = [\bfv^\top\!,\, \bfomega^\top]^\top \in \mathbb{R}^6$ determines the rate of change of the rigid-body pose according to the $SE(3)$ kinematics:
\begin{equation}
\label{eq:pose_kinematics}
\dot{\bfT} = \bfT \hat{\bfzeta} =: \bfT\begin{bmatrix}
\hat{\bfomega} & \bfv\\
\bf0^\top & 0
\end{bmatrix},
\end{equation}
where we overload $\hat{\cdot}$ to denote the mapping from a vector $\bfzeta \in \bbR^6$ to a $4 \times 4$ twist matrix $\hat{\bfzeta}$ in the Lie algebra $\mathfrak{se}(3)$ of $SE(3)$ and from a vector $\bfomega \in \bbR^3$ to a $3 \times 3$ skew-symmetric matrix $\hat{\bfomega}$ in the Lie algebra $\mathfrak{so}(3)$ of $SO(3)$:
\begin{equation}
\label{eq:hatmap}
\hat{\bfomega} = \begin{bmatrix}
0 & -\omega_3 & \omega_2 \\
\omega_3 & 0 & -\omega_1 \\
-\omega_2 & \omega_1 & 0 
\end{bmatrix}.
\end{equation}
Please refer to \cite{BarfootBook} for an excellent introduction to the use of matrix Lie groups in robot state estimation problems.

\subsection{Hamiltonian and Port-Hamiltonian Dynamics}

There are three predominant formulations of classical mechanics \cite{HolmBook} for describing the motion of macroscopic objects: Newtonian, Lagrangian, and Hamiltonian. Newtonian mechanics models the dynamics of mobile objects using forces and Cartesian coordinates according to the Newton's laws of motion. Lagrangian and Hamiltonian mechanics use generalized coordinates and energy in their formulations, which simplifies the equations of motion and reveals conserved quantities and their symmetries. Lagrangian mechanics considers generalized coordinates $\mathbf\frakq\in \mathbb{R}^n$ and velocity $\dot{\mathbf\frakq}\in \mathbb{R}^n$, and defines a Lagrangian function $\mathcal{L(\mathbf\frakq, \dot{\mathbf\frakq})}$ as the difference between kinetic energy $\frac{1}{2}\dot{\mathbf\frakq}^\top \bfM(\mathbf\frakq) \dot{\mathbf\frakq}$ and the potential energy $V(\mathbf\frakq)$:
\begin{equation}
\mathcal{L}(\mathbf\frakq, \dot{\mathbf\frakq}) =  \frac{1}{2}\dot{\mathbf\frakq}^\top \bfM(\mathbf\frakq) \dot{\mathbf\frakq} - V(\mathbf\frakq),
\end{equation}
where the symmetric positive definite matrix $\bfM(\mathbf\frakq) \in \bbS_{\succ 0}^{n\times n}$ represents the generalized mass of the system. Starting from the Lagrangian formulation, Hamiltonian mechanics expresses the system dynamics in terms of the generalized coordinates $\mathbf\frakq\in \mathbb{R}^n$ and generalized momenta $\mathbf\frakp\in \mathbb{R}^n$, defined as:
\begin{equation}
\mathbf\frakp = \frac{\partial \mathcal{L}(\mathbf\frakq, \dot{\mathbf\frakq})}{\partial \dot{\mathbf\frakq}} = \bfM(\mathbf\frakq) \dot{\mathbf\frakq}.
\end{equation}
Instead of the Lagrangian function, a Hamiltonian function $\mathcal{H}(\mathbf\frakq, \mathbf\frakp)$, representing the total energy of the dynamical systems, is obtained by a Legendre transformation of $\mathcal{L}(\mathbf\frakq, \dot{\mathbf\frakq})$:
\begin{equation}
\label{eq:hamiltonian_def}
\mathcal{H}(\mathbf\frakq, \mathbf\frakp) = \mathbf\frakp^\top \dot{\mathbf\frakq} - \mathcal{L}(\mathbf\frakq, \dot{\mathbf\frakq}) = \frac{1}{2}\mathbf\frakp^\top \bfM(\mathbf\frakq)^{-1} \mathbf\frakp + V(\mathbf\frakq).
\end{equation}
The Hamiltonian characterizes the system dynamics according to the equations:
\begin{equation}
\label{eq:hamiltonian_evolution_control}
\dot{\mathbf\frakq} = \frac{\partial \mathcal{H}}{\partial \mathbf\frakp}, \quad \dot{\mathbf\frakp} = -\frac{\partial \mathcal{H}}{\partial \mathbf\frakq} + \bfg(\mathbf\frakq)\bfu,
\end{equation}
where $\bfu \in \mathbb{R}^n$ is an external affine generalized control input with a coefficient matrix $\bfg(\mathbf\frakq) \in \bbR^{n \times n}$ that only affects the generalized momenta $\mathbf\frakp$. Without any external control, i.e., $\bfu = \bf0$, the total energy of the system is conserved, $\frac{d}{dt} \mathcal{H}(\mathbf\frakq, \mathbf\frakp) = 0$.

The notion of energy in dynamical systems is shared across multiple domains, including mechanical, electrical, and thermal. A Port-Hamiltonian generalization \cite{van2014port} of Hamiltonian mechanics is used to model systems with energy-storing elements (e.g., kinetic and potential energy), energy-dissipating elements (e.g., friction or resistors), and external energy sources (e.g., control inputs), connected via energy ports. An input-state-output Port-Hamiltonian system is expressed in the following form:
\begin{equation}
\label{eq:port_Hal_dyn}
\begin{bmatrix}
\dot{\mathbf\frakq} \\
\dot{\mathbf\frakp} \\
\end{bmatrix}
= (\mathbf\calJ(\mathbf\frakq, \mathbf\frakp) - \mathbf\calR(\mathbf\frakq, \mathbf\frakp))
\begin{bmatrix}
\frac{\partial \mathcal{H}}{\mathbf\frakq} \\
\frac{\partial \mathcal{H}}{\mathbf\frakp} 
\end{bmatrix} + \mathbf\calG(\mathbf\frakq, \mathbf\frakp)\bfu,
\end{equation}
%\begin{equation}
%\mathfrak{y} = \mathfrak{G(\mathbf\frakq, \mathbf\frakp)}^\top 
%\begin{bmatrix}
%\frac{\partial \mathcal{H}}{\mathbf\frakq} \\
%\frac{\partial \mathcal{H}}{\mathbf\frakp} 
%\end{bmatrix}
%\end{equation}
where $\mathbf\calJ(\mathbf\frakq, \mathbf\frakp)$ is a skew-symmetric inter-connection matrix, representing the energy-storing elements, $\mathbf\calR(\mathbf\frakq, \mathbf\frakp)$ is a positive semi-definite dissipation matrix, representing the energy-dissipating elements, and $\calG(\mathbf\frakq, \mathbf\frakp)$ is an input matrix such that $\mathbf\calG(\mathbf\frakq, \mathbf\frakp)\bfu$ represents the external energy sources. Intuitively, without the energy-dissipating elements and external energy sources, the skew-symmetry of $\mathbf\calJ(\mathbf\frakq, \mathbf\frakp)$ guarantees the energy conservation of the system. The Hamiltonian dynamics \eqref{eq:hamiltonian_evolution_control} are a special case of the Port-Hamiltonian dynamics \eqref{eq:port_Hal_dyn} with:
\begin{equation}
\mathbf\calJ(\mathbf\frakq, \mathbf\frakp) = \begin{bmatrix}
\phantom{-}\bf0 & \bfI \\
-\bfI & \bf0 
\end{bmatrix}, \; \mathbf\calR(\mathbf\frakq, \mathbf\frakp) = \bf0, \; \calG(\mathbf\frakq, \mathbf\frakp) = \begin{bmatrix}
\bf0 \\ \bfg(\mathfrak{q})
\end{bmatrix}.
\end{equation}

\subsection{Symplectic Neural ODE Networks}
\label{subsec:symplectic_neuralode}

The previous section described a system with Hamiltonian dynamics of the form $\dot{\bfx} = \bff(\bfx,\bfu)$ in \eqref{eq:hamiltonian_evolution_control} with $\bfx~=~[\mathbf\frakq^\top \quad \mathbf\frakp^\top]^\top$. Now, we consider approximating the function $\bff(\bfx,\bfu)$ when its elements, generalized mass $\bfM(\mathbf\frakq)$, potential energy $V(\mathbf\frakq)$, input matrix $\bfg(\mathbf\frakq)$, are unknown. Chen et al. \cite{chen2018neural} proposed a neural ODE formulation to approximate the closed-loop dynamics $\dot{\bfx} = \bff(\bfx,\bfpi(\bfx))$ for some unknown control policy $\bfu = \bfpi(\bfx)$ by a neural network $\bar{\bff}_{\bftheta}(\bfx)$. The parameters of the network $\bar{\bff}_{\bftheta}(\bfx)$ are trained using a data set $\calD = \{t^{(i)}_{0:N}, \bfx_{0:N}^{(i)}\}_i$ of state trajectory samples $\bfx_n^{(i)} = \bfx^{(i)}(t_n^{(i)})$ via forward and backward passes through a differentiable ODE solver. Given an initial state $\bfx_0^{(i)}$ at time $t_0^{(i)}$, a forward pass returns predicted states at times $t_1^{(i)}, \ldots, t_N^{(i)}$:
\begin{equation}
\label{eq:ode_solver}
\{\bar{\bfx}_1^{(i)}, \ldots, \bar{\bfx}_N^{(i)}\} = \text{ODESolver}(\bfx_0^{(i)}, \bar{\bff}_{\bftheta}, t_1^{(i)}, \ldots, t_N^{(i)}).
\end{equation}
The gradient of a loss function, $\sum_{i=1}^D\sum_{j=1}^N \ell(\bfx_j^{(i)},\bar{\bfx}_j^{(i)})$, is back-propagated by solving another ODE with adjoint states. The parameters $\bftheta$ are updated by gradient descent to minimize the loss.

For physical systems, a symplectic neural ODE formulation, proposed by Zhong et al. \cite{zhong2020symplectic}, extends the neural ODE by integrating the structure of the Hamiltonian dynamics in \eqref{eq:hamiltonian_def} and  \eqref{eq:hamiltonian_evolution_control} into the neural network model $\bar{\bff}_{\bftheta}(\bfx)$. Three neural networks are used to approximate the three unknown functions: generalized mass inverse $\bfM_{\bftheta_1}(\bfq)^{-1}$, potential energy $V_{\bftheta_2}(\bfq)$, and input matrix $\bfg_{\bftheta_3}(\bfq)$, where $\bftheta_1$, $\bftheta_2$, $\bftheta_3$ are the network parameters. A constant control input $\bfu$ is also considered, leading to the approximated dynamics:
\begin{equation}
\label{eq:gen_dyn_state_input}
\begin{bmatrix}
\dot{\bfx}\\
\dot{\bfu}
\end{bmatrix} = \begin{bmatrix}
\bar{\bff}_{\bftheta}(\bfx, \bfu)\\
\bf0
\end{bmatrix}
\end{equation}
where $\bar{\bff}_{\bftheta}(\bfx, \bfu)$ has the form of \eqref{eq:hamiltonian_evolution_control}.

%% file: tex/TechnicalApproach.tex
\section{Hamiltonian Dynamics on the $SE(3)$ Manifold}
\label{sec:hamiltonian_dynamics_pose}

The symplectic neural ODE formulation \cite{zhong2020symplectic} is designed for generalized coordinates $\mathbf\frakq$ and generalized momenta $\mathbf\frakp$, both in $\mathbb{R}^n$. For dynamical systems with states evolving on a manifold, such as $SE(3)$, it is necessary to enforce the manifold constraints (Sec. \ref{subsec:SO3_SE3_kinematics}) on the Hamiltonian dynamics. In this section, we describe the Hamiltonian equations of motion on the $SE(3)$ manifold and reformulate the dynamics in the input-state-output Port-Hamiltonian form.

Let $\mathbf\frakq = [\bfp^\top \quad \bfr_1^\top \quad \bfr_2^\top \quad \bfr_3^\top]^\top \in \mathbb{R}^{12}$ be the generalized coordinates, and $\bfzeta = [\bfv^\top\!\quad \bfomega^\top]^\top \in \mathbb{R}^6$ be the generalized velocity. Note that $\mathbf\frakq$ and $\bfzeta$ have different dimensions and satisfy the $SE(3)$ kinematics constraints in \eqref{eq:pose_kinematics}, re-expressed in vectorized form as:
\begin{equation}
\label{eq:pose_kinematics_rewritten}
\dot{\mathbf\frakq} = \mathbf\frakq^{\times}\mathbf\bfzeta, \qquad \mathbf\frakq^{\times} = \begin{bmatrix}
\bfR^\top\!\!\!\! & \bf0 & \bf0 & \bf0 \\
\bf0 & \hat{\bfr}_1^\top & \hat{\bfr}_2^\top & \hat{\bfr}_3^\top
\end{bmatrix}^\top.
\end{equation}
The Lagrangian function on $SE(3)$ is expressed in terms of $\mathbf\frakq$ and $\bfzeta$, instead of $\mathbf\frakq$ and $\dot{\mathbf\frakq}$:
\begin{equation} \label{eq:lagrangian_angvel_linvel}
\mathcal{L}(\mathbf\frakq, \bfzeta) = \frac{1}{2} \bfzeta^\top \bfM(\mathbf\frakq)\bfzeta -V(\mathbf\frakq).
\end{equation}
The generalized mass matrix has a block-diagonal form when the body frame is attached to the center of mass \cite{lee2017global}:
\begin{equation}
\label{eq:mass_matrix_hamiltonian}
\bfM(\mathbf\frakq) = \begin{bmatrix}
\bfM_1(\mathbf\frakq) & \bf0 \\
\bf0 & \bfM_2(\mathbf\frakq)
\end{bmatrix} \in \bbS_{\succ0}^{6 \times 6},
\end{equation}
where $\bfM_1(\mathbf\frakq), \bfM_2(\mathbf\frakq) \in \bbS_{\succ0}^{3 \times 3}$.
%where $c_{m}^{\{B\}} \in \mathbb{R}^3$ is the coordinates of the center of mass in the body-fix frame $\{B\}$, $\hat{c}_{m} \in \mathfrak{so(3)}$ is its hat map defined in Eq. \eqref{eq:hatmap} and the matrix $\bfM$ is of the following form:
%\begin{equation}
%\label{eq:mass_matrix_hamiltonian}
%\bfM = \begin{bmatrix}
%m\bfI_{3\times 3} & -m\hat{c_{m}} \\
%m\hat{c_{m}} & \bfJ
%\end{bmatrix}.
%\end{equation}
%Note that if the body-fixed frame is attached to the center of mass, then $c_{m} = 0$ and $\bfM$ is a block-diagonal matrix. 
The generalized momenta are defined, as before, via the partial derivative of the Lagrangian with respect to the velocity:
\begin{equation}
\label{eq:momenta_Mtwist}
\mathbf\frakp = \begin{bmatrix} {\mathbf\frakp}_{\bfv} \\ {\mathbf\frakp}_{\bfomega} \end{bmatrix} = \frac{\partial \mathcal{L}(\mathbf\frakq, \bfzeta)}{\partial \bfzeta} = \bfM(\mathbf\frakq)\bfzeta \in \mathbb{R}^6.
\end{equation}
While the Hamiltonian function, $\mathcal{H}(\mathbf\frakq, \mathbf\frakp)$, is the same as \eqref{eq:hamiltonian_def}, the system dynamics do not satisfy \eqref{eq:hamiltonian_evolution_control} due to the $SE(3)$ kinematics constraints in \eqref{eq:pose_kinematics_rewritten}. However, the dynamics can be specified in a Port-Hamiltonian form \cite{lee2017global, forni2015port, rashad2019port} as in \eqref{eq:port_Hal_dyn} with inter-connection matrix:
\begin{equation}
\label{eq:SE3_PH_J}
\bf\mathcal{J}(\mathbf\frakq, \mathbf\frakp) = \begin{bmatrix}
\bf0 & \mathbf\frakq^{\times} \\
-\mathbf\frakq^{\times\top} & \mathbf\frakp^{\times} 
\end{bmatrix}, \qquad \mathbf\frakp^{\times} = \begin{bmatrix}
\bf0 & \hat{\mathbf\frakp}_{\bfv}\\
\hat{\mathbf\frakp}_{\bfv} & \hat{\mathbf\frakp}_{\bfomega}
\end{bmatrix},
\end{equation}
and dissipation and input matrices:
\begin{equation}
\label{eq:SE3_PH_RG}
\bf\mathcal{R}(\mathbf\frakq, \mathbf\frakp) = \bf0, \qquad \mathcal{G}(\mathbf\frakq, \mathbf\frakp) = \begin{bmatrix} \bf0^\top & \bfg(\mathbf\frakq)^\top \end{bmatrix}^\top.
\end{equation}
Note that the dissipation matrix may not be necessarily zero and can, in fact, be used to model the effects of friction or drag forces \cite{zhong2020dissipative}. However, for the clarity of the model learning approach and unified control design, we leave this for future work.  %Such effects are often modeled \cite{faessler2017differential} as a linear transformation $\bfD(\mathbf\frakq, \mathbf\frakp) \in \bbS_{\succeq 0}^{6\times 6}$ of the generalized velocity $\bfzeta$ and only affect the generalized momenta $\mathbf\frakp$:
%
%\begin{equation}
%\bf\mathcal{R}(\mathbf\frakq, \mathbf\frakp) = \begin{bmatrix}
%\bf0 & \bf0\\
%\bf0 & \bfD(\mathbf\frakq, \mathbf\frakp)
%\end{bmatrix}.
%\end{equation}
%
We consider a system with unknown generalized mass $\bfM(\mathbf\frakq)$, potential energy $V(\mathbf\frakq)$, input matrix $\bfg(\mathbf\frakq)$, and design a structured neural ODE network to learn these terms from state-control trajectories.

%%%%%%%%%%%%%%%%%%%%%%%%%%%%%%%%%%%%%%%%%%%%%%%%%%%%%%%%%%%%%%%%%%%%%%%%%%%%%%%%%%%%%%%%%%%%%%%%%%%%%%%%

\section{Hamiltonian $SE(3)$ Dynamics Learning}
\label{sec:data_gen_net_design}

This section describes a neural ODE network design incorporating $SE(3)$ kinematics and energy conservation constraints. We discuss data generation, embedding the constraints into the model architecture, and the training process.

\subsection{Training Data Generation}
\label{subsec:data_gen}

We collect a data set $\calD = \{t_{0:N}^{(i)}, \bfx^{(i)}_{0:N}, \bfu^{(i)}\}_{i=1}^D$ consisting of state sequences $\bfx^{(i)}_{0:N}$, where $\bfx_n^{(i)}~=~[\mathbf\frakq_n^{(i)\top} \quad \bfzeta_n^{(i)\top}]^\top$, $\mathbf\frakq_n^{(i)}~=~[\bfp_n^{(i)\top} \quad \bfr_{1n}^{(i)\top} \quad \bfr_{2n}^{(i)\top} \quad \bfr_{3n}^{(i)\top}]^\top~\in~\mathbb{R}^{12}$, $\bfzeta_n^{(i)}~=~[\bfv_n^{(i)\top} \quad \bfomega_n^{(i)\top}]^\top \in \mathbb{R}^6$ for $n = 0, \ldots, N$. Such data are generated by applying a constant control input $\bfu^{(i)}$ to the system and sampling the state $\bfx_{n}^{(i)} = \bfx^{(i)}(t_n^{(i)})$ at times $t_n^{(i)}$ for $n = 0, \ldots, N$. The generalized coordinates $\mathbf\frakq$ and velocity $\bfzeta$ may be obtained from an odometry algorithm, such as \cite{vio_benchmark,OdometrySurvey}, or from a motion capture system. In physics-based simulation the data can be generated by applying random control inputs $\bfu^{(i)}$. In real-world applications, where safety is a concern, data may be collected by a human operator manually controlling the robot.

\subsection{Model Design}
\label{subsec:network_design}

To learn the dynamics $\bff(\bfx,\bfu)$ from the data set $\calD$, we design a neural ODE network (Sec. \ref{subsec:symplectic_neuralode}), approximating the dynamics via a parametric function $\bar{\bff}_{\bftheta}(\bfx,\bfu)$. To impose the $SE(3)$ Hamiltonian equations described in Sec. \ref{sec:hamiltonian_dynamics_pose} on the structure of the neural network $\bar{\bff}_{\bftheta}(\bfx,\bfu)$, we expand \eqref{eq:port_Hal_dyn} with the interconnection matrix $\calJ$ in \eqref{eq:SE3_PH_J}:
\begin{eqnarray}
\dot{\bfp} &=& \bfR\frac{\partial{\mathcal{H}(\mathbf\frakq, \mathbf\frakp)}}{\partial \mathbf\frakp_{\bfv}}, \label{eq:hamiltonian_pose_pdot} \\
\dot{\bfr_i} &=& \bfr_i \times \frac{\partial{\mathcal{H}(\mathbf\frakq, \mathbf\frakp)}}{\partial \mathbf\frakp_{\bfomega}}, \quad i = 1,2,3 \label{eq:hamiltonian_pose_rdot} \\
\dot{\mathbf\frakp}_{\bfv} &=& \mathbf\frakp_{\bfv}\times \frac{\partial{\mathcal{H}(\mathbf\frakq, \mathbf\frakp)}}{\partial \mathbf\frakp_{\bfomega}} - \bfR^\top \frac{\partial{\mathcal{H}(\mathbf\frakq, \mathbf\frakp)}}{\partial \bfp} + \bfg_{\bfv}(\mathbf\frakq)\bfu, \label{eq:hamiltonian_pose_pvdot}\\
\dot{\mathbf\frakp}_{\bfomega} &=& \mathbf\frakp_{\bfomega} \times \frac{\partial{\mathcal{H}(\mathbf\frakq, \mathbf\frakp)}}{\partial \mathbf\frakp_{\bfomega}} + \mathbf\frakp_{\bfv}\times \frac{\partial{\mathcal{H}(\mathbf\frakq, \mathbf\frakp)}}{\partial \mathbf\frakp_{\bfv}} + \nonumber \\
&& + \sum_{i = 1}^3 \bfr_i \times \frac{\partial{\mathcal{H}(\mathbf\frakq, \mathbf\frakp)}}{\partial \bfr_i} + \bfg_{\bfomega}(\mathbf\frakq)\bfu,  \label{eq:hamiltonian_pose_pomegadot}
\end{eqnarray}
where the input matrix $\bfg(\mathbf\frakq) = \begin{bmatrix} \bfg_{\bfv}(\mathbf\frakq)^\top & \bfg_{\bfomega}(\mathbf\frakq)^\top \end{bmatrix}^\top$ is decomposed into components corresponding to $\mathbf\frakp_\bfv$ and $\mathbf\frakp_{\bfomega}$ and the Hamiltonian function $\mathcal{H}(\mathbf\frakq, \mathbf\frakp)$ is defined in \eqref{eq:hamiltonian_def}. Since the generalized momenta $\mathbf\frakp$ are not directly available in the data set $\calD$ (Sec.~\ref{subsec:data_gen}), we use the time derivative of the generalized velocity, obtained from \eqref{eq:momenta_Mtwist}:
\begin{equation}
\label{eq:hamiltonian_zetadot}
\dot{\bfzeta} =  \prl{ \frac{d}{dt} \bfM^{-1}(\mathbf\frakq) }\mathbf\frakp + \bfM^{-1}(\mathbf\frakq)\dot{\mathbf\frakp}.
\end{equation}
%
%This equation is used instead of Eq. \eqref{eq:hamiltonian_pose_pomegadot} and \eqref{eq:hamiltonian_pose_pvdot} .
The approximated dynamics function $\bar{\bff}_{\bftheta}(\bfx,\bfu)$ is described by \eqref{eq:hamiltonian_pose_pdot}, \eqref{eq:hamiltonian_pose_rdot}, and \eqref{eq:hamiltonian_zetadot} with an internal state $\mathbf\frakp$ satisfying the ODEs in \eqref{eq:hamiltonian_pose_pvdot} and \eqref{eq:hamiltonian_pose_pomegadot}.

To integrate the Hamiltonian equations into the structure of $\bar{\bff}_{\bftheta}(\bfx,\bfu)$, we use four neural networks with parameters $\bftheta = (\bftheta_1, \bftheta_2, \bftheta_3, \bftheta_4)$ to approximate the blocks $\bfM^{-1}_{1}(\mathbf\frakq)$, $\bfM^{-1}_{2}(\mathbf\frakq)$ of the inverse generalized mass in \eqref{eq:mass_matrix_hamiltonian}, the potential energy $V(\mathbf\frakq)$, and the input matrix $\bfg(\mathbf\frakq)$, respectively. The blocks $\bfM^{-1}_{1}(\mathbf\frakq)$, $\bfM^{-1}_{2}(\mathbf\frakq)$ are forced to be positive definite using Cholesky decomposition:
\begin{equation}
\label{eq:M_cholesky}
\begin{aligned}
\bfM^{-1}_{1}(\mathbf\frakq) &= \bfL_{1}(\mathbf\frakq)\bfL_{1}^\top(\mathbf\frakq) + \varepsilon \bfI,\\
\bfM^{-1}_{2}(\mathbf\frakq) &= \bfL_{2}(\mathbf\frakq)\bfL_{2}^\top(\mathbf\frakq) + \varepsilon \bfI,
\end{aligned}
\end{equation}
where $\bfL_{1}(\mathbf\frakq)$, $\bfL_{2}(\mathbf\frakq)$ are lower-triangular matrices implemented as two neural networks with parameters $\bftheta_1$ and $\bftheta_2$, respectively, and $\varepsilon > 0$. The time derivative $\frac{d}{dt} \bfM(\mathbf\frakq)^{-1}$ and the partial derivatives $\frac{\partial \mathcal{H}}{\partial\mathbf\frakp}$ and $\frac{\partial \mathcal{H}}{\partial\mathbf\frakq}$ are calculated using automatic differentiation, e.g. by Pytorch.

In many applications, prior information is available about the generalized mass matrices $\bfM^{-1}_{1}(\mathbf\frakq)$, $\bfM^{-1}_{2}(\mathbf\frakq)$ and can be used to assist the training process. If a potentially inaccurate estimate of the masses is available, the neural networks modeling $\bfL_{1}(\mathbf\frakq)$, $\bfL_{2}(\mathbf\frakq)$ can be pre-trained to fit this estimate. For example, if we guess that the generalized mass matrix inverse is an identity matrix $\bfI$, we can sample $M$ random inputs $\mathbf\frakq$ and pre-train the network to output $\bfI$. The pre-trained networks are used as initialization for the full training process which may correct these parameters using the dataset $\calD$.

\subsection{Training Process}
\label{sec:training}

Let $\bar{\bfx}^{(i)}(t) \in \bbR^{18}$ denote the state trajectory predicted with control input $\bfu^{(i)}$ by the approximate dynamics $\bff_{\bftheta}$ initialized at $\bar{\bfx}^{(i)}(t_0^{(i)}) = \bfx_0^{(i)}$. For sequence $i$, forward passes through the ODE solver in \eqref{eq:ode_solver} return the predicted states $\bar{\bfx}^{(i)}_{0:N}$ at times $t^{(i)}_{0:N}$, where $\bar{\bfx}_n^{(i)} = [\bar{\mathbf\frakq}_n^{(i)\top} \quad \bar{\bfzeta}_n^{(i)\top}]^\top$, $\bar{\mathbf\frakq}_n^{(i)\top} = [\bar{\bfp}_n^{(i)\top} \quad \bar{\bfr}_{1n}^{(i)\top} \quad \bar{\bfr}_{2n}^{(i)\top} \quad \bar{\bfr}_{3n}^{(i)\top}]^\top$, $\bar{\bfzeta}_n^{(i)\top} = [\bar{\bfv}_n^{(i)\top} \quad \bar{\bfomega}_n^{(i)\top}]^\top$, for $n = 1, \ldots, N$. The predicted rotation matrix $\bar{\bfR}_n^{(i)} = [\bar{\bfr}_{1n}^{(i)} \quad \bar{\bfr}_{2n}^{(i)} \quad \bar{\bfr}_{3n}^{(i)}]^\top$ and the ground-truth one $\bfR_n^{(i)} = [\bar{\bfr}_{1n}^{(i)} \quad \bar{\bfr}_{2n}^{(i)} \quad \bar{\bfr}_{3n}^{(i)}]^\top$ are used to calculate an orientation loss:
\begin{equation}
\mathcal{L}_{\bfR}(\bftheta) = \sum_{i = 1}^D \sum_{n = 1}^N \left\|\left(\log (\bar{\bfR}_n^{(i)} \bfR_n^{(i)\top})\right)^{\vee} \right\|_2^2,
\end{equation}
where $\log : SE(3) \mapsto \mathfrak{so}(3)$ converts a rotation matrix to a skew-symmetric matrix and $(\cdot)^\vee : \mathfrak{so}(3) \mapsto \bbR^3$ is the inverse of the hat map in \eqref{eq:hatmap}, which extracts the components of a vector $\bfw \in \mathbb{R}^3$ from $\hat{\bfw} \in \mathfrak{so}(3)$. We use the squared Euclidean norm to calculate losses for the position and generalized velocity terms: 
%\begin{equation}
%\mathcal{L}_{\bfp}(\bftheta) = \sum_{i = 1}^D \sum_{n = 1}^N \Vert \bfp_n^{(i)} - \bar{\bfp}_n^{(i)} \Vert^2_2, \mathcal{L}_{\bfzeta}(\bftheta) = \sum_{i = 1}^D \sum_{n = 1}^N \Vert \bfzeta_n^{(i)} - \bar{\bfzeta}_n^{(i)} \Vert^2_2.
%\end{equation}
%
\begin{equation}
\begin{aligned}
\mathcal{L}_{\bfp}(\bftheta) &=& \sum_{i = 1}^D \sum_{n = 1}^N \Vert \bfp_n^{(i)} - \bar{\bfp}_n^{(i)} \Vert^2_2,\\
\mathcal{L}_{\bfzeta}(\bftheta) &=& \sum_{i = 1}^D \sum_{n = 1}^N \Vert \bfzeta_n^{(i)} - \bar{\bfzeta}_n^{(i)} \Vert^2_2.
\end{aligned}
\end{equation}
The total loss $\calL(\bftheta)$ is defined as:
\begin{equation}
\calL(\bftheta)= \mathcal{L}_{\bfR}(\bftheta) + \mathcal{L}_{\bfp}(\bftheta) + \mathcal{L}_{\bfzeta}(\bftheta).
\end{equation}

The gradient of the total loss function $\calL(\bftheta)$ is back-propagated by solving an ODE with adjoint states \cite{chen2018neural}. Specifically, let $\bfa = \frac{\partial \calL}{\partial \bar{\bfx}}$ be the adjoint state and $\bfs = (\bar{\bfx}, \bfa, \frac{\partial \calL}{\partial \bftheta})$ be the augmented state. The augmented state dynamics are \cite{chen2018neural}:
\begin{equation}
\dot{\bfs} = \bar{\bff}_{\bfs} = (\bar{\bff}_{\bftheta}, -\bfa^\top\frac{\partial \bar{\bff}_{\bftheta}}{\partial \bar{\bfx}}, -\bfa^\top\frac{\partial \bar{\bff}_{\bftheta}}{\partial \bftheta}).
\end{equation}
The predicted state $\bar{\bfx}$, the adjoint state $\bfa$, and the derivatives $\frac{\partial \calL}{\partial \bftheta}$ can be obtained by a single call to a reverse-time ODE solver starting from $\bfs_N = \bfs({t_N})$:
\begin{equation}
\bfs_0 = \left(\bar{\bfx}_0, \bfa_0, \frac{\partial \calL}{\partial \bftheta}\right) = \text{ODESolver}(\bfs_N, \bar{\bff}_\bfs, t_N),
\end{equation}
where at each time $t_k, k = 1, \ldots, N$, the adjoint state $\bfa_{k}$ at time $t_k$ is reset to $ \frac{\partial \calL}{\partial \bar{\bfx}_{k}}$. The resulting derivative $ \frac{\partial \calL}{\partial \bftheta}$ is used to update the parameters $\bftheta$ using gradient descent.

%%%%%%%%%%%%%%%%%%%%%%%%%%%%%%%%%%%%%%%%%%%%%%%%%%%%%%%%%%%%%%%%%%%%%%%%%%%%%%%%%%%%%%%%%%%%%%%%%%%%%%%%

%% file: tex/ControllerDesign.tex
\section{Energy-based Control Design}
\label{sec:controller_design}

The function $\bar{\bff}_{\bftheta}$ learned in Sec.~\ref{sec:data_gen_net_design} satisfies the input-state-output Port-Hamiltonian dynamics on the $SE(3)$ manifold in Sec. \ref{sec:hamiltonian_dynamics_pose} by design. This section extends the interconnection and damping assignment passivity-based control (IDA-PBC) \cite{van2014port, wang2008modified, souza2014passivity} to the $SE(3)$ manifold to achieve trajectory tracking (Problem \ref{problem:setpoint_reg}) based on the learned $SE(3)$ Port-Hamiltonian dynamics.

% state trajectory $\bfx^*(t) = (\mathbf\frakq^*(t), \mathbf\frakp^*(t))$ that the system should track

First, consider a desired regulation point $(\mathbf\frakq^*, \mathbf\frakp^*)$ that the system should be stabilized to. The Hamiltonian function $\mathcal{H}(\mathbf\frakq, \mathbf\frakp)$, representing the total energy of the system, generally does not have a minimum at $(\mathbf\frakq^*, \mathbf\frakp^*)$. An IDA-PBC controller is designed to inject additional energy $\calH_a(\mathbf\frakq, \mathbf\frakp)$ such that the desired total energy $\mathcal{H}_d(\mathbf\frakq, \mathbf\frakp)$ achieves its minimum at $(\mathbf\frakq^*, \mathbf\frakp^*)$:
\begin{equation}
\mathcal{H}_d(\mathbf\frakq, \mathbf\frakp) = \mathcal{H}(\mathbf\frakq, \mathbf\frakp) + \mathcal{H}_a(\mathbf\frakq, \mathbf\frakp).
\end{equation}
To drive the system towards the desired state $(\mathbf\frakq^*, \mathbf\frakp^*)$, the Port-Hamiltonian dynamics in \eqref{eq:port_Hal_dyn} should be shaped into a desired form \cite{van2014port,ortega2002stabilization,wang2008modified}:
\begin{equation}
\label{eq:desired_port_Hal_dyn}
\begin{bmatrix}
\dot{\mathbf\frakq} \\
\dot{\mathbf\frakp} \\
\end{bmatrix}
= (\mathcal{J}_d(\mathbf\frakq, \mathbf\frakp) - \mathcal{R}_d(\mathbf\frakq, \mathbf\frakp))
\begin{bmatrix}
\frac{\partial \mathcal{H}_d}{\partial \mathbf\frakq} \\
\frac{\partial \mathcal{H}_d}{\partial \mathbf\frakp} 
\end{bmatrix}.
\end{equation}
Specifically, the control input $\bfu$ should be chosen so that \eqref{eq:port_Hal_dyn} and \eqref{eq:desired_port_Hal_dyn} are equal. This usual matching equation design does not directly apply to trajectory tracking problems, especially for under-actuated systems \cite{souza2014passivity, wang2008modified}.

Consider a desired state trajectory $\bfx^*(t) = (\mathbf\frakq^*(t), \bfzeta^*(t))$ that the system should track where $\mathbf\frakq^*(t)$ is the desired generalized coordinate and $\bfzeta^*(t) = \begin{bmatrix}\bfv^*(t)^\top & \bfomega^*(t)^\top\end{bmatrix}^\top$ is the desired generalized velocity expressed in the desired frame. Let $\mathbf\frakp^*$ denote the desired momenta, which will be defined later using $\bfzeta^*$. Let $\bfx_e = (\mathbf\frakq_e, \mathbf\frakp_e)$ denote the time-varying error in the generalized coordinates and momenta, respectively. Let $\bfR_e = \bfR^{*\top} \bfR = \begin{bmatrix}
	\bfr_{e1} & \bfr_{e2} & \bfr_{e3}
	\end{bmatrix}^\top$ represent the rotation error between the current rotation matrix $\bfR$ and the desired one $\bfR^*$. The error $\mathbf\frakq_e$ in the generalized coordinates is:
	\begin{equation}
	\label{eq:qe_def}
	\mathbf\frakq_e = \begin{bmatrix} \bfp - \bfp^* \\
	\bfr_{e1} \\ \bfr_{e2} \\ \bfr_{e3} \end{bmatrix}.
\end{equation}
The error in the generalized momenta is $\mathbf\frakp_e = \mathbf\frakp - \mathbf\frakp^*$. For under-actuated trajectory tracking, the desired total energy should be defined in terms of the error state as: 
\begin{equation}
\label{eq:desired_Hamil_tracking}
\calH_d(\mathbf\frakq_e, \mathbf\frakp_e) = \frac{1}{2}\mathbf\frakp_e^\top \bfM_d^{-1}(\mathbf\frakq_e)\mathbf\frakp_e + V_d(\mathbf\frakq_e),
\end{equation}
where $\bfM_d(\mathbf\frakq_e)$ and $V_d(\mathbf\frakq_e)$ are the desired generalized mass and potential energy. The new states $\mathbf\frakq_e$ and $\mathbf\frakp_e$ satisfy the desired dynamics:
\begin{equation}
\label{eq:desired_port_Hal_dyn_tracking}
\begin{bmatrix}
\dot{\mathbf\frakq}_e \\
\dot{\mathbf\frakp}_e \\
\end{bmatrix}
= (\mathcal{J}_d(\mathbf\frakq_e, \mathbf\frakp_e) - \mathcal{R}_d(\mathbf\frakq_e, \mathbf\frakp_e))
\begin{bmatrix}
\frac{\partial \mathcal{H}_d}{\partial \mathbf\frakq_e} \\
\frac{\partial \mathcal{H}_d}{\partial \mathbf\frakp_e} 
\end{bmatrix},
\end{equation}
leading to the following matching equations for the control input design:
\begin{align}
\mathbf\calG(\mathbf\frakq, \mathbf\frakp)\bfu = (&\mathcal{J}_d(\mathbf\frakq_e, \mathbf\frakp_e) - \mathcal{R}_d(\mathbf\frakq_e, \mathbf\frakp_e))
\begin{bmatrix}
\frac{\partial \mathcal{H}_d}{\partial \mathbf\frakq_e} \\
\frac{\partial \mathcal{H}_d}{\partial \mathbf\frakp_e} 
\end{bmatrix} \label{eq:matching_eqn_tracking}\\
- (&\mathbf\calJ(\mathbf\frakq, \mathbf\frakp) - \mathbf\calR(\mathbf\frakq, \mathbf\frakp))
\begin{bmatrix}
\frac{\partial \mathcal{H}}{\partial \mathbf\frakq} \\
\frac{\partial \mathcal{H}}{\partial \mathbf\frakp} 
\end{bmatrix} + \begin{bmatrix}
\dot{\mathbf\frakq} \\
\dot{\mathbf\frakp}
\end{bmatrix} - \begin{bmatrix}
\dot{\mathbf\frakq}_e \\
\dot{\mathbf\frakp}_e
\end{bmatrix}. \notag
\end{align}
Choosing the following desired inter-connection matrix and dissipation matrix:
\begin{equation}
\mathcal{J}_d(\mathbf\frakq_e, \mathbf\frakp_e) = \begin{bmatrix}
\bf0 & \bfJ_1 \\
-\bfJ_1^\top & \bfJ_2\end{bmatrix}, \quad
\mathcal{R}_d(\mathbf\frakq_e, \mathbf\frakp_e) = \begin{bmatrix}
\bf0 & \bf0 \\
\bf0 & \bfK_\bfd 
\end{bmatrix},
\end{equation}
and plugging $\calJ(\mathbf\frakq, \mathbf\frakp)$ and $\calR(\mathbf\frakq, \mathbf\frakp)$ from \eqref{eq:SE3_PH_J} into the matching equations in \eqref{eq:matching_eqn_tracking}, leads to:
\begin{eqnarray}
\bf0 &=& \bfJ_1 \frac{\partial \calH_d}{\partial \mathbf\frakp_e} - \mathbf\frakq^{\times} \frac{\partial \calH}{\partial \mathbf\frakp} + \dot{\mathbf\frakq} - \dot{\mathbf\frakq}_e, \label{eq:tracking_cond1} \\
\bfg(\mathbf\frakq)\bfu &=& \mathbf\frakq^{\times\top}\frac{\partial \calH}{\partial \mathbf\frakq} - \bfJ_1^\top \frac{\partial \calH_d}{\partial \mathbf\frakq_e} + \bfJ_2\frac{\partial \calH_d}{\partial \mathbf\frakp_e} - \mathbf\frakp^{\times}\frac{\partial \calH}{\partial \mathbf\frakp} \nonumber \\
&& - \bfK_\bfd\frac{\partial \calH_d}{\partial \mathbf\frakp_e} + \dot{\mathbf\frakp} - \dot{\mathbf\frakp}_e. \label{eq:tracking_cond2} 
\end{eqnarray}
Assuming $\bfM_d(\mathbf\frakq_e) = \bfM(\mathbf\frakq)$ , the condition \eqref{eq:tracking_cond1} is satisfied if we choose $\bfJ_1 =\begin{bmatrix}
	\bfR^\top\!\!\!\! & \bf0 & \bf0 & \bf0 \\
	\bf0 & \hat{\bfr}_{e1}^\top & \hat{\bfr}_{e2}^\top & \hat{\bfr}_{e3}^\top
	\end{bmatrix}^\top$ and $\mathbf\frakp^* = \bfM \begin{bmatrix} \bfR^\top \bfR^*\bfv^* \\ \bfR^\top \bfR^* \bfomega^*\end{bmatrix}$, i.e., the desired momenta are defined based on \eqref{eq:momenta_Mtwist} with the desired velocity expressed in the body frame. The desired control input can be obtained from \eqref{eq:tracking_cond2} as the sum $\bfu = \bfu_{ES} + \bfu_{DI}$ of an energy-shaping component $\bfu_{ES}$ and a damping-injection component $\bfu_{DI}$:
\begin{eqnarray}
\bfu_{ES} &=& \bfg^{\dagger}(\mathbf\frakq)\left(\mathbf\frakq^{\times\top}\frac{\partial \calH}{\partial \mathbf\frakq} - \mathbf\bfJ_1^{\top}\frac{\partial \calH_d}{\partial \mathbf\frakq_e} + \bfJ_2\frac{\partial \calH_d}{\partial \mathbf\frakp_e} \right. \nonumber \\
&& \left. \qquad \qquad- \mathbf\frakp^{\times}\frac{\partial \calH}{\partial \mathbf\frakp} + \dot{\mathbf\frakp} - \dot{\mathbf\frakp}_e \right), \label{eq:general_u_ES} \\
\bfu_{DI} &=& -\bfg^{\dagger}(\mathbf\frakq)\bfK_\bfd\frac{\partial \calH_d}{\partial \mathbf\frakp_e}, \label{eq:general_u_DI}
\end{eqnarray}
where $\bfg^{\dagger}(\mathbf\frakq) = \left(\bfg^{\top}(\mathbf\frakq)\bfg(\mathbf\frakq)\right)^{-1}\bfg^{\top}(\mathbf\frakq)$ is the pseudo-inverse of $\bfg(\mathbf\frakq)$. The control input $\bfu_{ES}$ exists as long as the desired $\bfM_d(\mathbf\frakq_e)$ and $\bfJ_2$ satisfy the following matching condition:
\begin{eqnarray}
\bfg^\perp(\mathbf\frakq)\left(\mathbf\frakq^{\times\top}\frac{\partial \calH}{\partial \mathbf\frakq} - \bfJ_1^{\top}\frac{\partial \calH_d}{\partial \mathbf\frakq_e} + \bfJ_2\frac{\partial \calH_d}{\partial \mathbf\frakp_e}\right. &&\nonumber \\
- \left.\mathbf\frakp^{\times}\frac{\partial \calH}{\partial \mathbf\frakp} + \dot{\mathbf\frakp} - \dot{\mathbf\frakp}_e \right) &=& 0, \label{eq:matching_condition_tracking}
\end{eqnarray}
where $\bfg^{\perp}(\mathbf\frakq)$ is a maximal-rank left annihilator of $\bfg(\mathbf\frakq)$, i.e., $\bfg^{\perp}(\mathbf\frakq)\bfg(\mathbf\frakq) = \bf0$.
%Solving this matching partial differential equation \eqref{eq:matching_condition_tracking} for the desired mass  $\bfM_d(\mathbf\frakq)$ and $\bfJ_2$ is hard in general.

To avoid solving the PDE equations in \eqref{eq:matching_condition_tracking} needed for the IDA-PBC controller, the parameters of the desired Port-Hamiltonian dynamics can be further specified to satisfy $\bfJ_{2} \equiv 0$. With this choice, we add the following Hamiltonian energy term:
\begin{align}
\mathcal{H}_a&(\mathbf\frakq, \mathbf\frakp) = -\mathcal{H}(\mathbf\frakq, \mathbf\frakp) + \frac{1}{2}(\bfp - \bfp^*)^\top\bfK_\bfp(\bfp - \bfp^*) \\
& + \frac{1}{2} \tr(\bfK_{\bfR}(\bfI - \bfR^{*\top}\bfR)) + \frac{1}{2}(\mathbf\frakp-\mathbf\frakp^*)^\top\bfM^{-1}(\mathbf\frakq)(\mathbf\frakp-\mathbf\frakp^*)\notag
\end{align}
to reshape the open-loop Hamiltonian $\mathcal{H}(\mathbf\frakq, \mathbf\frakp)$ into a desired total energy $\mathcal{H}_d(\mathbf\frakq, \mathbf\frakp) = \mathcal{H}(\mathbf\frakq, \mathbf\frakp) + \mathcal{H}_a(\mathbf\frakq, \mathbf\frakp)$ that is minimized along the desired trajectory $\bfx^* = (\mathbf\frakq^*, \mathbf\frakp^*)$. For an $SE(3)$ rigid-body system with constant generalized mass matrix $\bfM_d = \bfM$ and $\bfJ_{2} = 0$, the energy-shaping term in \eqref{eq:general_u_ES} and the damping-injection term in \eqref{eq:general_u_DI} simplify as follows:
\begin{align}
\label{eq:idapbc_pose_twist_tracking}
\bfu_{ES}(\mathbf\frakq, \mathbf\frakp) &= \bfg^{\dagger}(\mathbf\frakq)\left(\mathbf\frakq^{\times\top} \frac{\partial V}{\partial \mathbf\frakq} - \mathbf\frakp^{\times}\bfM^{-1}\mathbf\frakp - \bfe(\mathbf\frakq,\mathbf\frakq^*) + \dot{\frakp}^*\right), \notag\\
\bfu_{DI}(\mathbf\frakq, \mathbf\frakp) &= -\bfg^{\dagger}(\mathbf\frakq) \bfK_\bfd  \bfM^{-1}(\mathbf\frakp-\mathbf\frakp^*),
\end{align}
where the generalized coordinate error between $\mathbf\frakq$ and $\mathbf\frakq^*$ is:
\begin{equation}
\label{eq:coordinate-error}
\begin{aligned}
\bfe(\mathbf\frakq,\mathbf\frakq^*) := \bfJ_1^{\top}\frac{\partial  V_d}{\partial \mathbf\frakq_e} = \begin{bmatrix} \bfR^\top\bfK_\bfp(\bfp - \bfp^*) \\ \frac{1}{2}\prl{\bfK_{\bfR}\bfR^{*\top}\bfR-\bfR^\top\bfR^{*}\bfK_{\bfR}^\top}^{\vee}\end{bmatrix}.
\end{aligned}
\end{equation}
Without requiring a priori knowledge of the system parameters, the control design in \eqref{eq:idapbc_pose_twist_tracking} and \eqref{eq:coordinate-error} offers a unified control approach for $SE(3)$ Hamiltonian systems that achieves trajectory tracking, if permissible by the system's degree of under-actuation.
%

%is decomposed into energy-shaping and damping-injection components: $\bfu_{ES}(\mathbf\frakq, \mathbf\frakp)$, and $\bfu_{DI}(\mathbf\frakq, \mathbf\frakp)$, respectively. It 

%%%%%%%%%%%%%%%%%%%%%%%%%%%%%%%%%%%%%%%%%%%%%%%%%%%%%%%%%%%%%%%%

%% file: tex/ExperimentalResults.tex
\section{Evaluation}
\label{sec:experimental_results}
We verify the effectiveness of our Hamiltonian-based neural ODE network for dynamics learning and control on the $SE(3)$ manifold using two fully-actuated systems (a pendulum and a rigid body) and one under-actuated system (a quadrotor). The implementation details for the experiments are provided in Appendix \ref{subsec:implement_details}.

\begin{figure*}[t]
\begin{subfigure}[t]{0.245\textwidth}
        \centering
        \includegraphics[width=\textwidth]{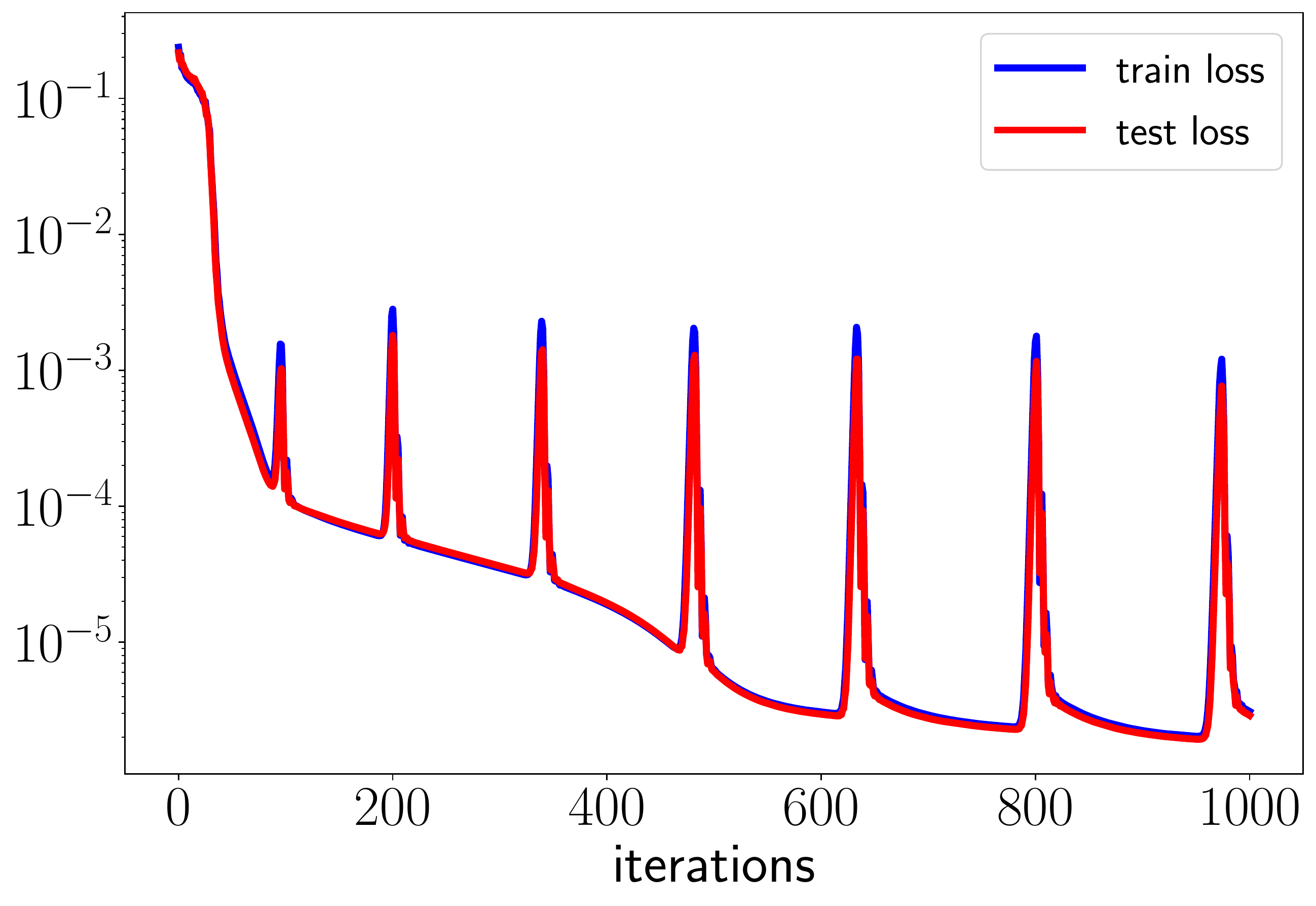}
        \caption{Loss (log scale)}
        \label{fig:pend_loss}
\end{subfigure}%
\hfill%
\begin{subfigure}[t]{0.232\textwidth}
        \centering
        \includegraphics[width=\textwidth]{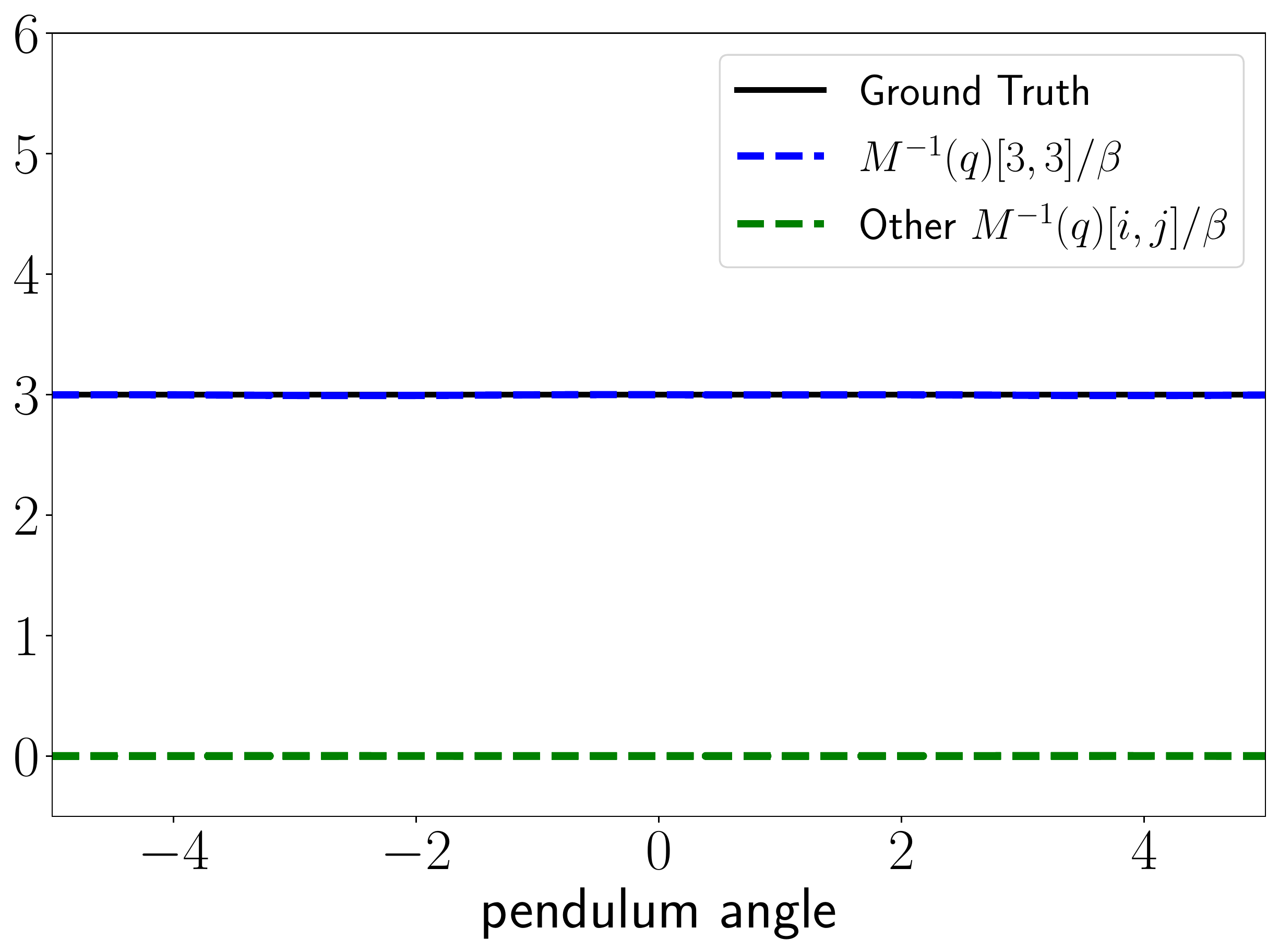}
        \caption{$\bfM^{-1}(q)/\beta$ versus $\varphi$.}
        \label{fig:pend_M_x_all}
\end{subfigure}%
\hfill%
\begin{subfigure}[t]{0.245\textwidth}
        \centering
\includegraphics[width=\textwidth]{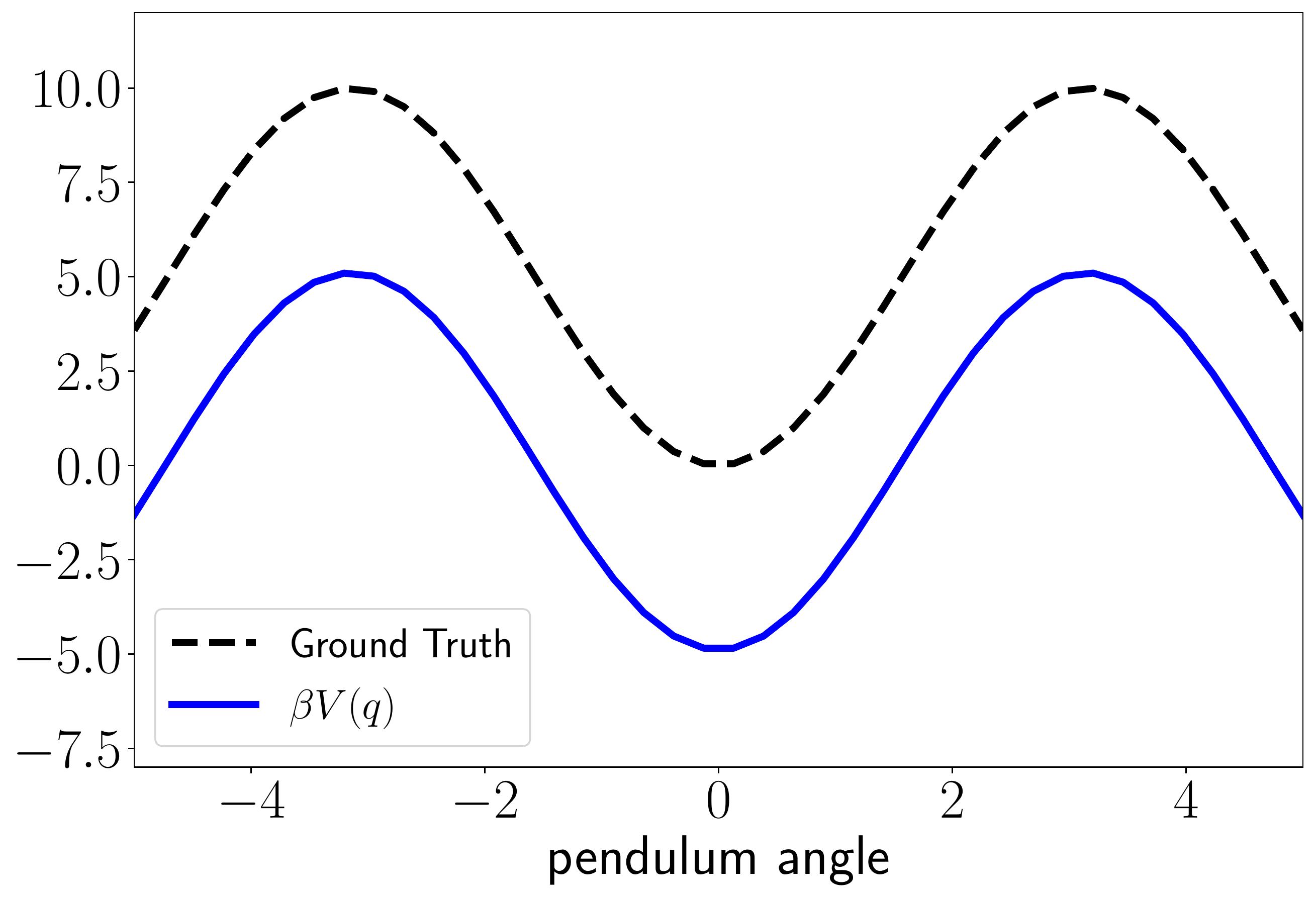}%
        \caption{$\beta V(q)$ versus $\varphi$.}
        \label{fig:pend_Vx}
\end{subfigure}%
\hfill%
\begin{subfigure}[t]{0.248\textwidth}
        \centering
        \includegraphics[width=\textwidth]{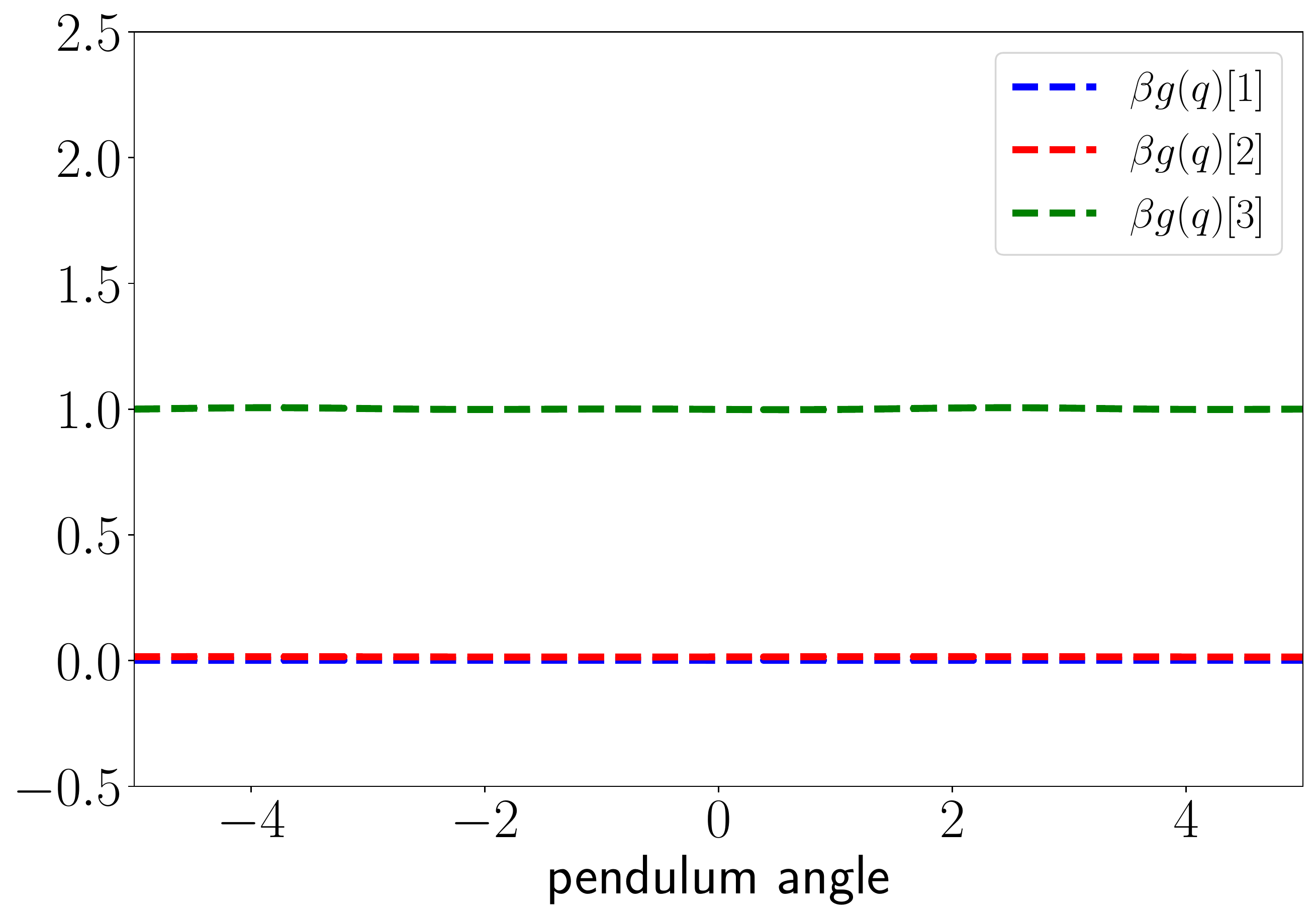}
        \caption{$\beta\bfg(q)$ versus $\varphi$.}
        \label{fig:pend_gx}
\end{subfigure}%

\begin{subfigure}[t]{0.232\textwidth}
        \centering
\includegraphics[width=\textwidth]{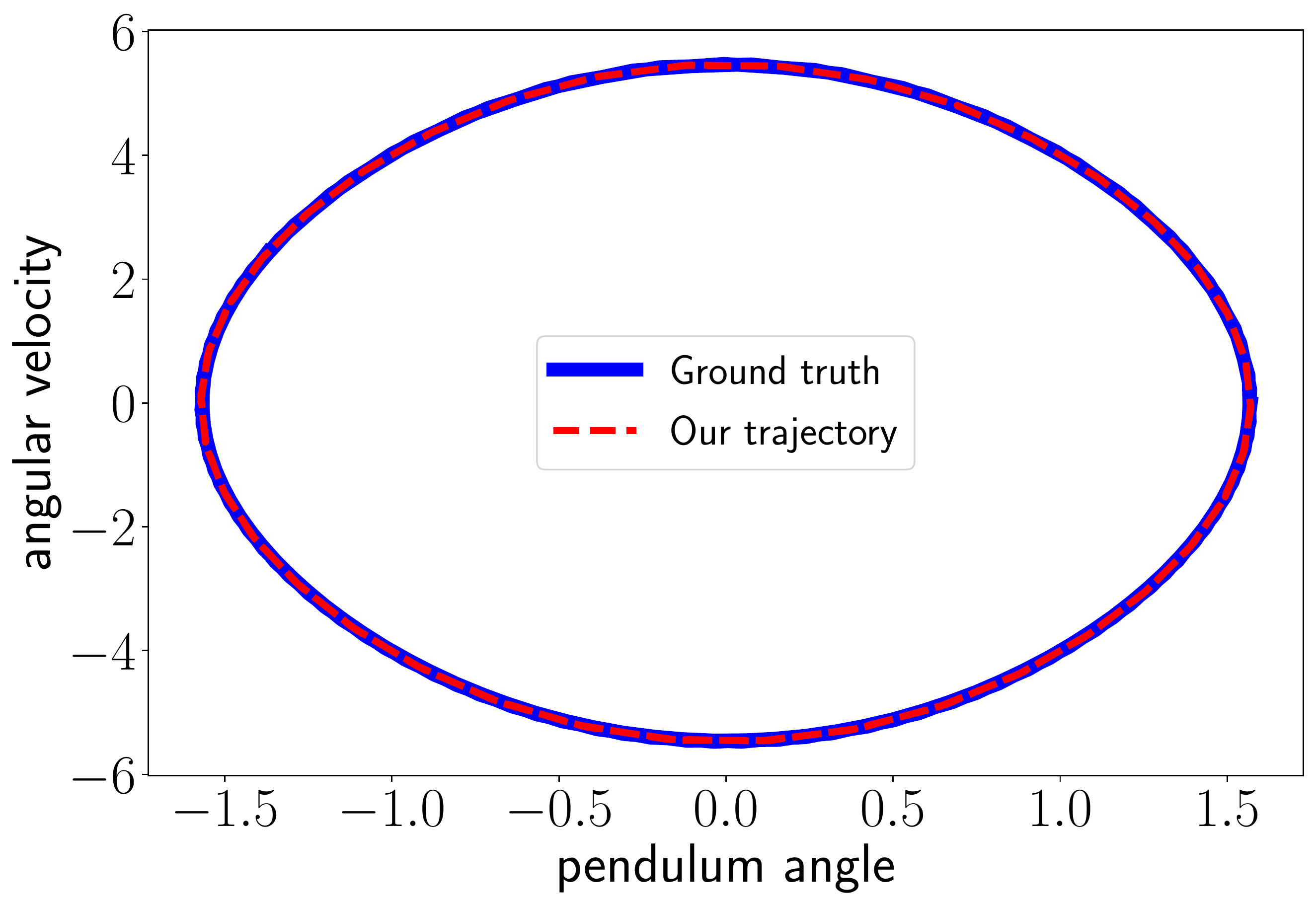}%
        \caption{Portrait of a test trajectory.}
        \label{fig:phase_portrait}
\end{subfigure}%
\hfill%
\begin{subfigure}[t]{0.245\textwidth}
        \centering
\includegraphics[width=\textwidth]{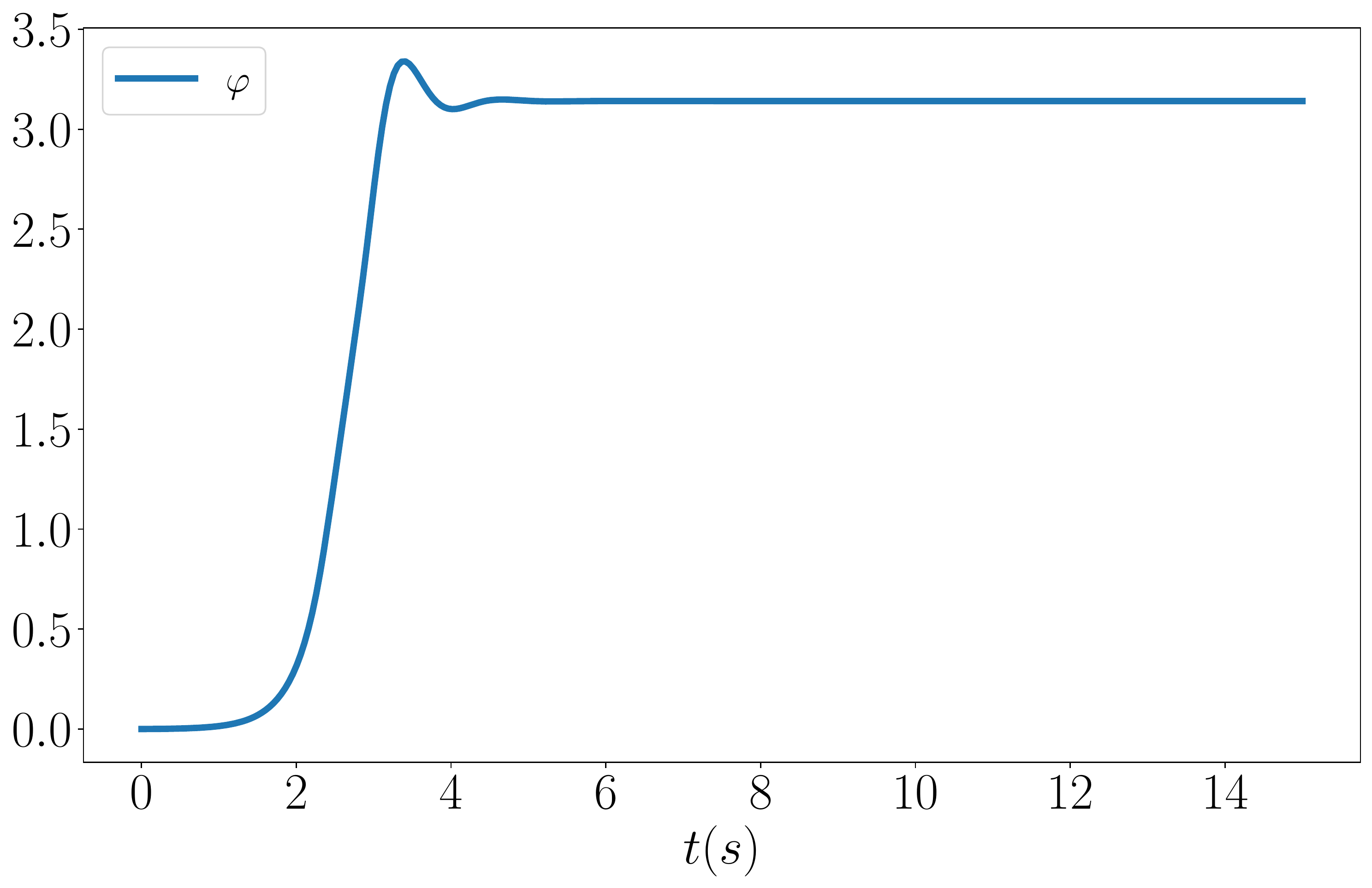}%
        \caption{Controlled angle $\varphi$.}
        \label{fig:control_theta}
\end{subfigure}%
\hfill%
\begin{subfigure}[t]{0.245\textwidth}
        \centering
\includegraphics[width=\textwidth]{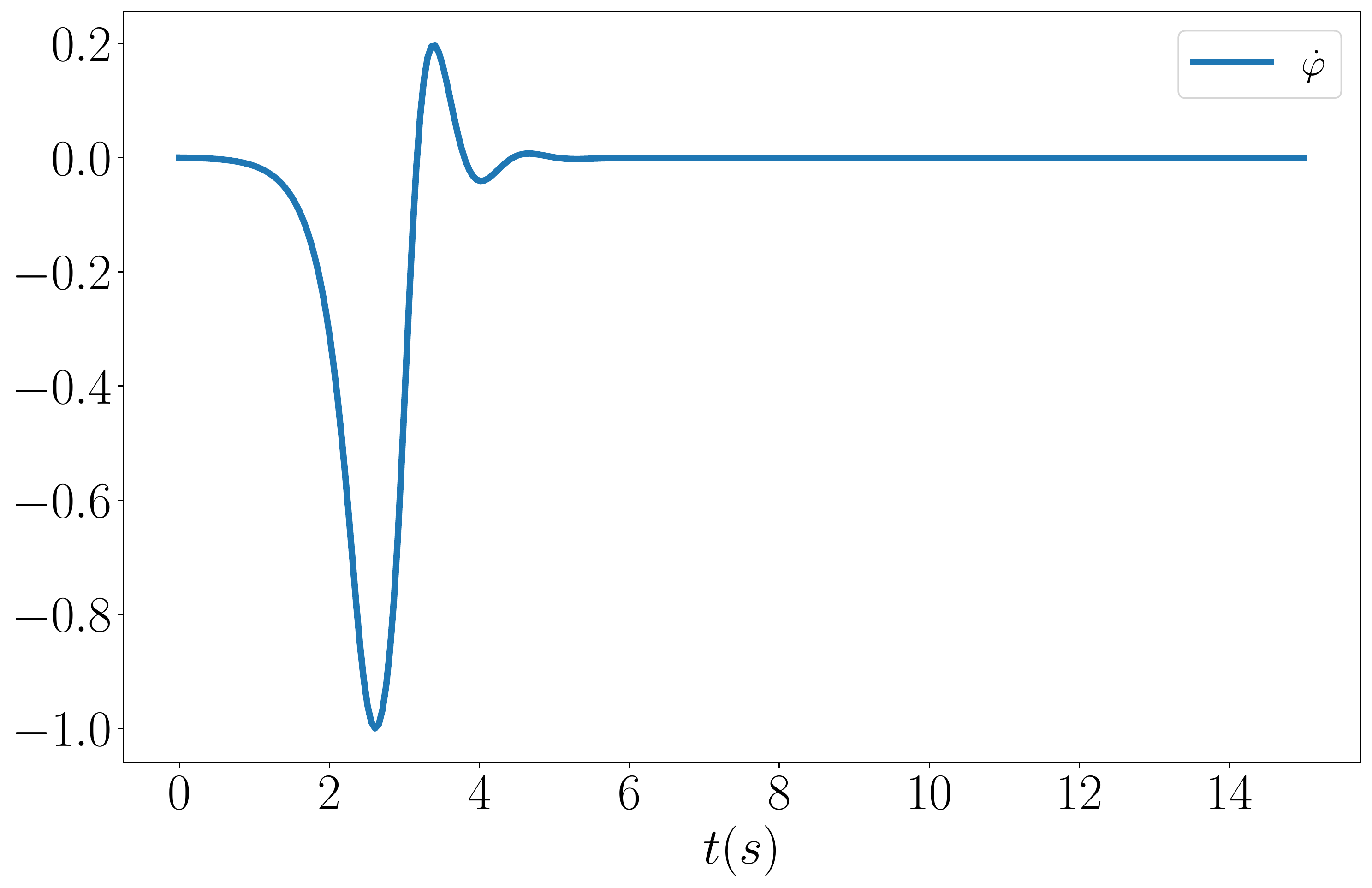}%
        \caption{Controlled angular velocity $\dot{\varphi}$.}
        \label{fig:control_thetadot}
\end{subfigure}%
\hfill%
\begin{subfigure}[t]{0.245\textwidth}
        \centering
\includegraphics[width=\textwidth]{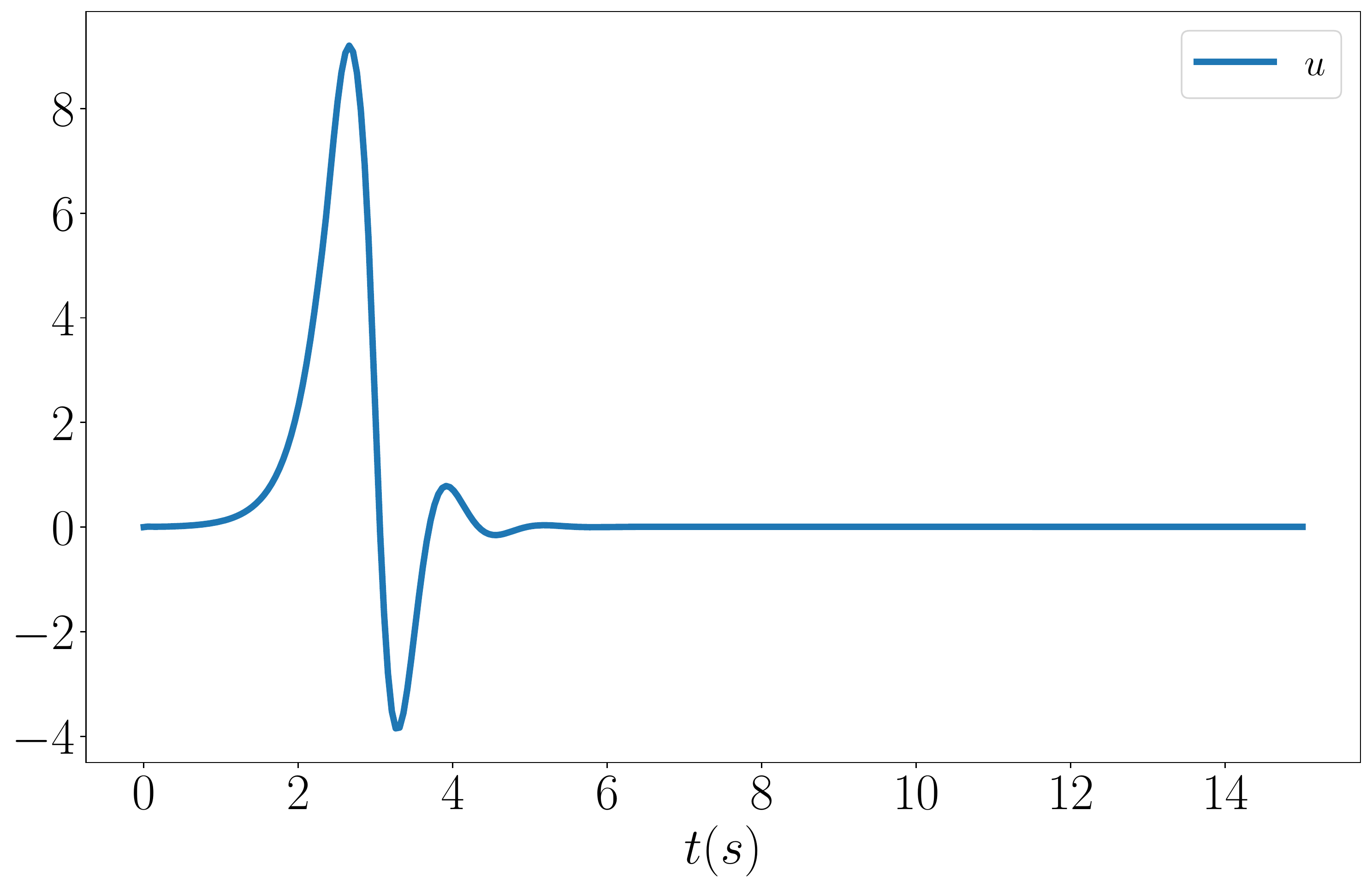}%
        \caption{Control input $u$.}
        \label{fig:control_input}
\end{subfigure}%
\caption{Evaluation of our $SO(3)$ Hamiltonian neural ODE network on a pendulum system with scale factor $\beta = 4.6$.}
\label{fig:pend_exp}
\end{figure*}

\subsection{Pendulum}
\label{subsec:pendulum_so3}
We consider a pendulum with the following dynamics:
%\begin{equation}
%\label{eq:pend_gt_dyn}
%\dot{\mathbf\frakq} = 3p, \quad \dot{\mathbf\frakp} = -5\sin{\mathbf\frakq} + \bfu,
%\end{equation}
\begin{equation}
\label{eq:pend_gt_q_dyn}
\ddot{\varphi} = -15\sin{\varphi} + 3u,
\end{equation}
where $\varphi$ is the angle of the pendulum with respect to its vertically downward position and $u$ is a scalar control input. The ground-truth mass, potential energy, and the input coefficient are: $m = 1/3, V(\varphi) = 5(1-\cos{\varphi})$, and $g(\varphi) = 1$, respectively.
%, leading to the following Hamiltonian function:
%\begin{equation}
%\mathcal{H}(\mathbf\frakq, \mathfrak{p}) = 1.5\mathbf\frakp^2 + 5(1-\cos{\mathbf\frakq}).
%\end{equation}
We collected data of the form $\{(\cos{\varphi}, \sin{\varphi}, \dot{\varphi})\}$ from an OpenAI Gym environment, provided by \cite{zhong2020symplectic}, with the dynamics in \eqref{eq:pend_gt_q_dyn}. To illustrate our manifold-constrained neural ODE learning, we viewed $\varphi$ as a yaw angle and convert $(\cos{\varphi}, \sin{\varphi})$ into a rotation matrix:
\begin{equation}
\bfR = \begin{bmatrix}
\cos{\varphi} & - \sin{\varphi} & 0 \\
\sin{\varphi} & \cos{\varphi} & 0 \\
0 & 0 & 1 
\end{bmatrix}.
\end{equation}
We used $\bfomega = [0, 0, \dot{\varphi}]$ for angular velocity and remove position $\bfp$ and linear velocity $\bfv$ from the Hamiltonian dynamics in \eqref{eq:hamiltonian_pose_pdot}, \eqref{eq:hamiltonian_pose_rdot}, \eqref{eq:hamiltonian_pose_pvdot}, \eqref{eq:hamiltonian_pose_pomegadot}, restricting the system to the $SO(3)$ manifold with generalized coordinates $\mathbf\frakq = [\bfr_1^\top \quad \bfr_2^\top \quad \bfr_3^\top]^\top \in \mathbb{R}^{9}$.

As described in Sec. \ref{subsec:data_gen}, control inputs $\bfu^{(i)}$ were sampled randomly and applied to the pendulum for five time intervals of $0.05s$, forming a dataset $\mathcal{D} = \left\{t_{0:N}^{(i)},\mathbf\frakq_{0:N}^{(i)}, \bfomega_{0:N}^{(i)}, \bfu^{(i)})\right\}_{i =1}^D$ with $N = 5$ and $D=5120$. We trained the $SO(3)$ Hamiltonian neural ODE network as described in Sec. \ref{sec:training} for 1000 iterations with no pre-training. %No network initialization described in Sec. \ref{subsec:network_init} was done.

As noted in \cite{zhong2020symplectic}, since the generalized momenta $\mathbf\frakp$ are not available in the dataset, the dynamics of $\mathbf\frakq$ in \eqref{eq:pend_gt_q_dyn} do not change if $\mathbf\frakp$ is scaled by a factor $\beta > 0$. This is also true in our formulation as scaling $\mathbf\frakp$ leaves the dynamics of $\mathbf\frakq$ in \eqref{eq:hamiltonian_pose_rdot} and \eqref{eq:hamiltonian_zetadot} unchanged. To emphasize this scale invariance, let $\bfM_\beta(\mathbf\frakq) = \beta\bfM(\mathbf\frakq)$, $V_\beta(\mathbf\frakq) = \beta V(\mathbf\frakq)$, $\bfg_\beta(\mathbf\frakq) = \beta\bfg(\mathbf\frakq)$, and:
\begin{equation}
\begin{aligned}
\mathbf\frakp_{\beta} &= \bfM_{\beta}(\mathbf\frakq)\bfomega = \beta \mathbf\frakp,  \quad\qquad \dot{\mathbf\frakp}_\beta = \beta \dot{\mathbf\frakp}, \\
\mathcal{H}_{\beta}(\mathbf\frakq,\mathbf\frakp) &= \frac{1}{2}\mathbf\frakp_{\beta}^\top \bfM_{\beta}^{-1}(\mathbf\frakq) \mathbf\frakp_{\beta} + V_{\beta}(\mathbf\frakq)=\beta\mathcal{H}(\mathbf\frakq,\mathbf\frakp), \\
\frac{\partial{\mathcal{H}_{\beta}(\mathbf\frakq,\mathbf\frakp)}}{\partial \mathbf\frakp_\beta} &= \bfM^{-1}_{\beta}(\mathbf\frakq) \mathbf\frakp_{\beta} = \frac{\partial{H(\mathbf\frakq,\mathbf\frakp)}}{\partial \mathbf\frakp},
\end{aligned}
\end{equation}
guaranteeing that \eqref{eq:hamiltonian_pose_pdot}, \eqref{eq:hamiltonian_pose_rdot}, and \eqref{eq:hamiltonian_zetadot} still hold.

\begin{figure}[t]
\centering
\begin{subfigure}[t]{0.24\textwidth}
        \centering
\includegraphics[width=\textwidth]{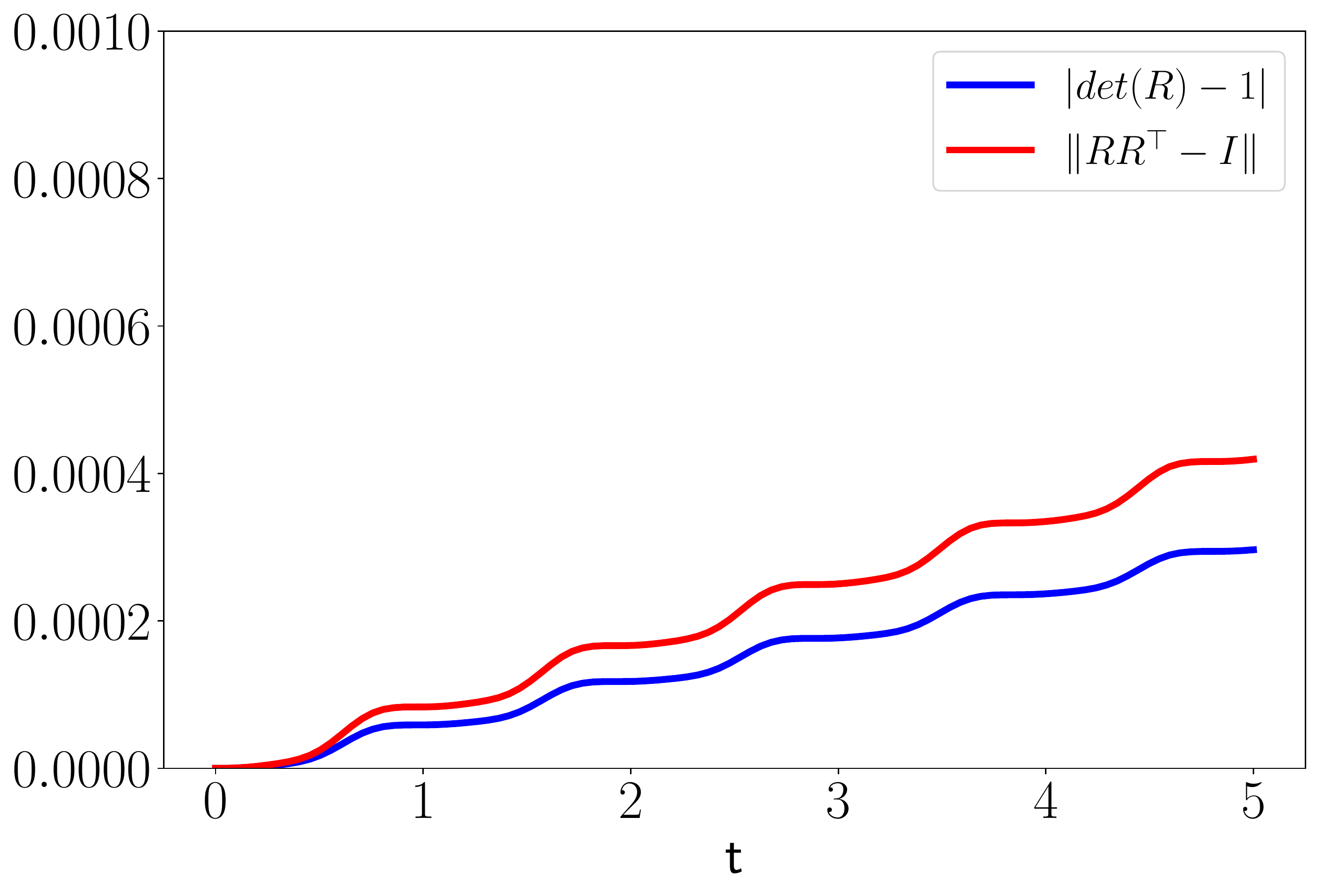}%
        \caption{SO(3) constraints.}
        \label{fig:pend_so3_constraints}
\end{subfigure}%
\begin{subfigure}[t]{0.235\textwidth}
        \centering
\includegraphics[width=\textwidth]{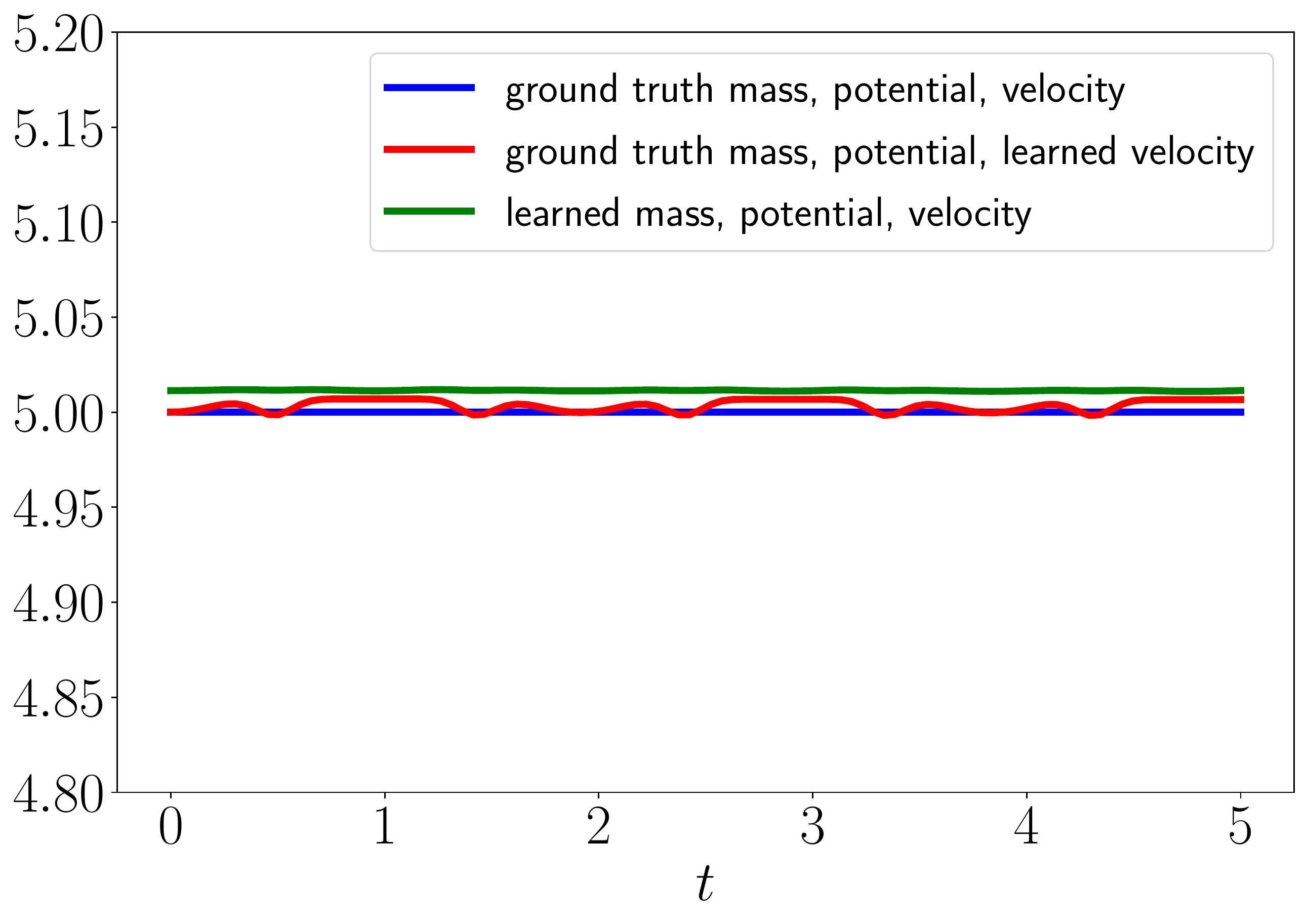}%
        \caption{Total energy.}
        \label{fig:pend_total_energy}
\end{subfigure}%
\caption{$SO(3)$ constraints and total energy along a trajectory rollout from the learned pendulum model, initialized at $\phi = \pi/2$.}
\label{fig:pend_so3_totalenergy}
\end{figure}

Fig. \ref{fig:pend_exp} shows the training and testing behavior of our $SO(3)$ Hamiltonian ODE network.
%t our neural ODE network converged with a scaling factor $\beta = 4.6$. The loss in log scale in Fig. \ref{fig:pend_loss}. As we consider $\bfomega = [0, 0, \dot{\varphi}]$ as the angular velocity, 
Fig. \ref{fig:pend_M_x_all} and \ref{fig:pend_gx} show that the $\brl{\bfM(\mathbf\frakq)^{-1}}_{3,3}$ entry of the mass inverse and the $\brl{\bfg(\mathbf\frakq)}_3$ entry of the input matrix are close to their correct values of $3$ and $1$, respectively, while the other entries are close to zero. Fig. \ref{fig:pend_Vx} indicates a constant gap between the learned and the ground-truth potential energy, which can be explained by the relativity of potential energy. The learned pendulum dynamics are illustrated by the phase portrait of a test trajectory in Fig. \ref{fig:phase_portrait}, which coincides with the ground-truth portrait.

We tested stabilization of the pendulum based on the learned dynamics to the stable equilibrium at the downward position $\varphi = 0$ and to the unstable equilibrium at the upward position $\varphi = \pi$. 
%
% The control input was calculated from \eqref{eq:u_es} and \eqref{eq:u_di}
Since the pendulum is a fully-actuated system and the desired state has zero velocity, potential energy shaping is enough to drive the system to the desired state $(\mathbf\frakq^*, \bf0)$. Our energy-based controller in Sec. \ref{sec:controller_design} achieved this by setting $\bfJ_2 = \mathbf\frakp^\times$, $\bfM_d(\mathbf\frakq) = \bfM(\mathbf\frakq)$ in \eqref{eq:general_u_ES}, leading to:
\begin{equation}
\bfu = \bfu_{ES} + \bfu_{DI} = \bfg^{\dagger}_{\bfomega}(\mathbf\frakq)\sum_{i = 1}^3 \bfr_i \times \frac{\partial \mathcal{H}_a}{\partial\bf\bfr_i} - \bfg^\dagger_{\bfomega}\bfK_\bfd(\mathbf\frakq)\bfomega,
\end{equation}
where the additional energy $\mathcal{H}_a(\mathbf\frakq, \mathbf\frakp)$ was simplified by removing the position error:
\begin{equation}
\mathcal{H}_a(\mathbf\frakq, \mathbf\frakp) = -V(\mathbf\frakq) + \frac{1}{2} \tr(\bfK_{\bfR}(\bfI - \bfR^{*\top}\bfR)).
\end{equation}
The controlled angle $\varphi$ and angular velocity $\dot{\varphi}$ as well as the control inputs $\bfu$ with gains $\bfK_\bfR = \bfI$ and $\bfK_\bfd = 0.4 \bfI$ are shown over time in Fig. \ref{fig:control_theta}, \ref{fig:control_thetadot}, and \ref{fig:control_input}. We can see that the controller was able to smoothly drive the pendulum from $\phi = 0$ to $\phi = \pi$, relying only on the learned dynamics.

Lastly, Fig. \ref{fig:pend_so3_constraints} verifies that the $SO(3)$ constraints are satisfied by plotting $|\det{\bfR} - 1|$ and $\Vert \bfR \bfR^\top - \bfI \Vert$ from the learned model along a $5$-second trajectory rollout with zero input, initialized at $\phi = \pi/2$. To verify the energy conservation as well, we calculated the Hamiltonian via \eqref{eq:hamiltonian_def} along the trajectory using: (1) ground-truth mass, potential energy, and velocity; (2) ground-truth mass and potential energy but velocity rolled out from the learned dynamics; and (3) all mass, potential energy and velocity rolled out from the learned dynamics. The constant Hamiltonian in Fig. \ref{fig:pend_total_energy} verifies that, with no control input, our model obeys the law of energy conservation and remains close to the ground-truth energy after scaling by $\beta$.

\subsection{Fully-actuated Rigid Body}

\begin{figure*}[h]
\begin{subfigure}[t]{0.24\textwidth}
        \centering
        \includegraphics[width=\textwidth]{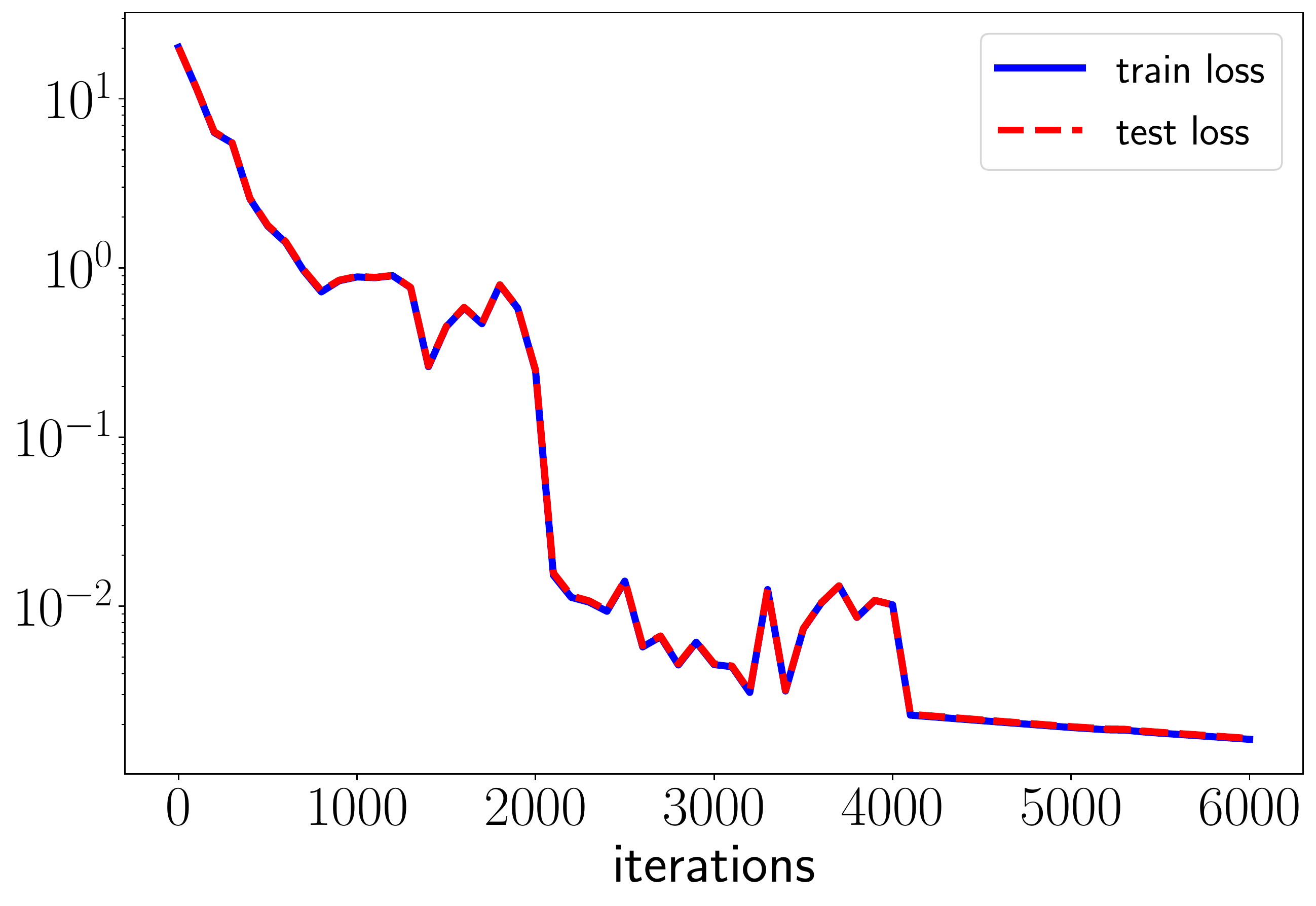}
        \caption{Loss (log scale)}
        \label{fig:rigid_loss}
\end{subfigure}%
\hfill%
\begin{subfigure}[t]{0.238\textwidth}
        \centering
        \includegraphics[width=\textwidth]{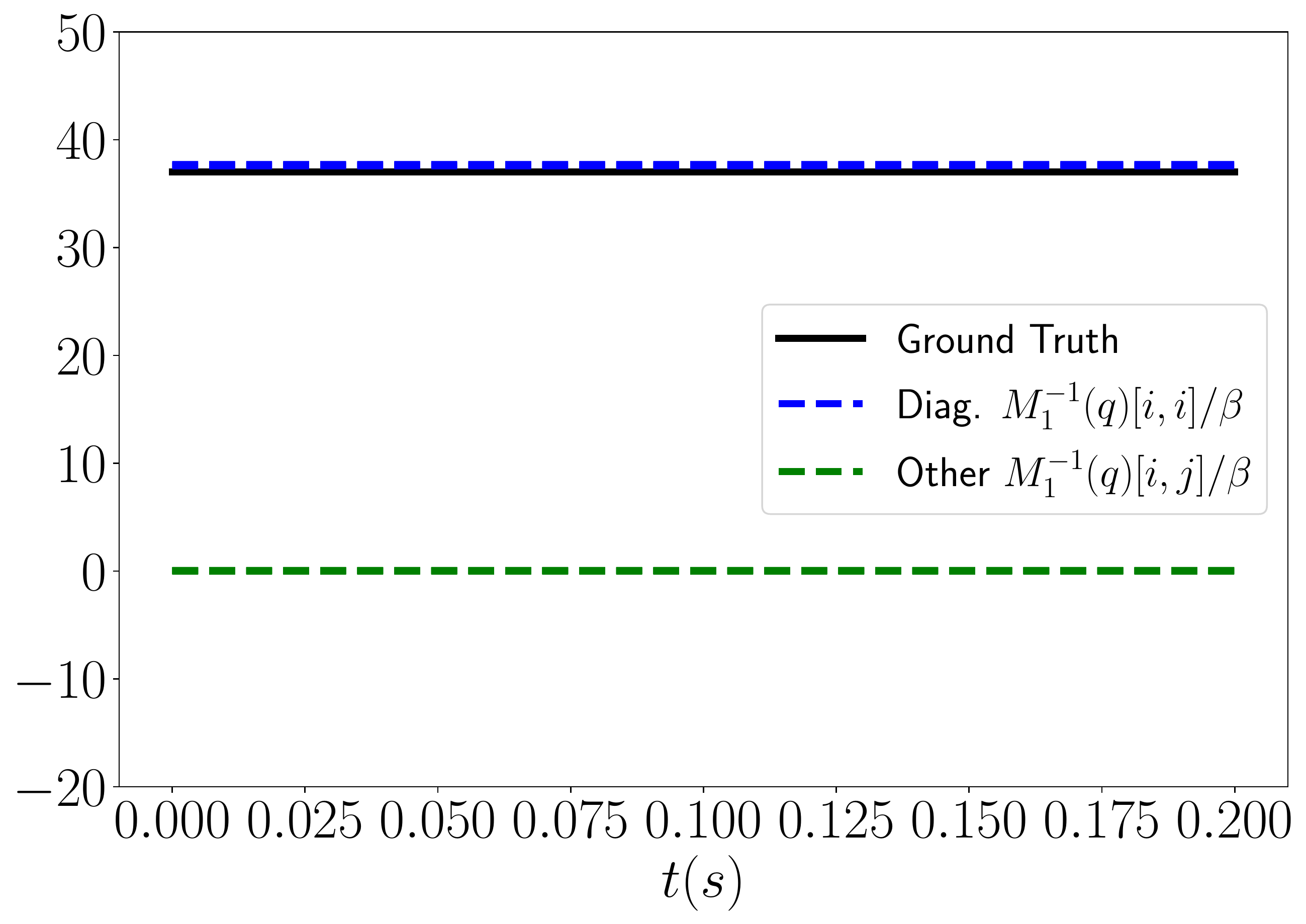}
        \caption{$\bfM_1^{-1}(q)/\beta$ over time.}
        \label{fig:rigid_M1_x_all}
\end{subfigure}%
\hfill%
\begin{subfigure}[t]{0.249\textwidth}
        \centering
		\includegraphics[width=\textwidth]{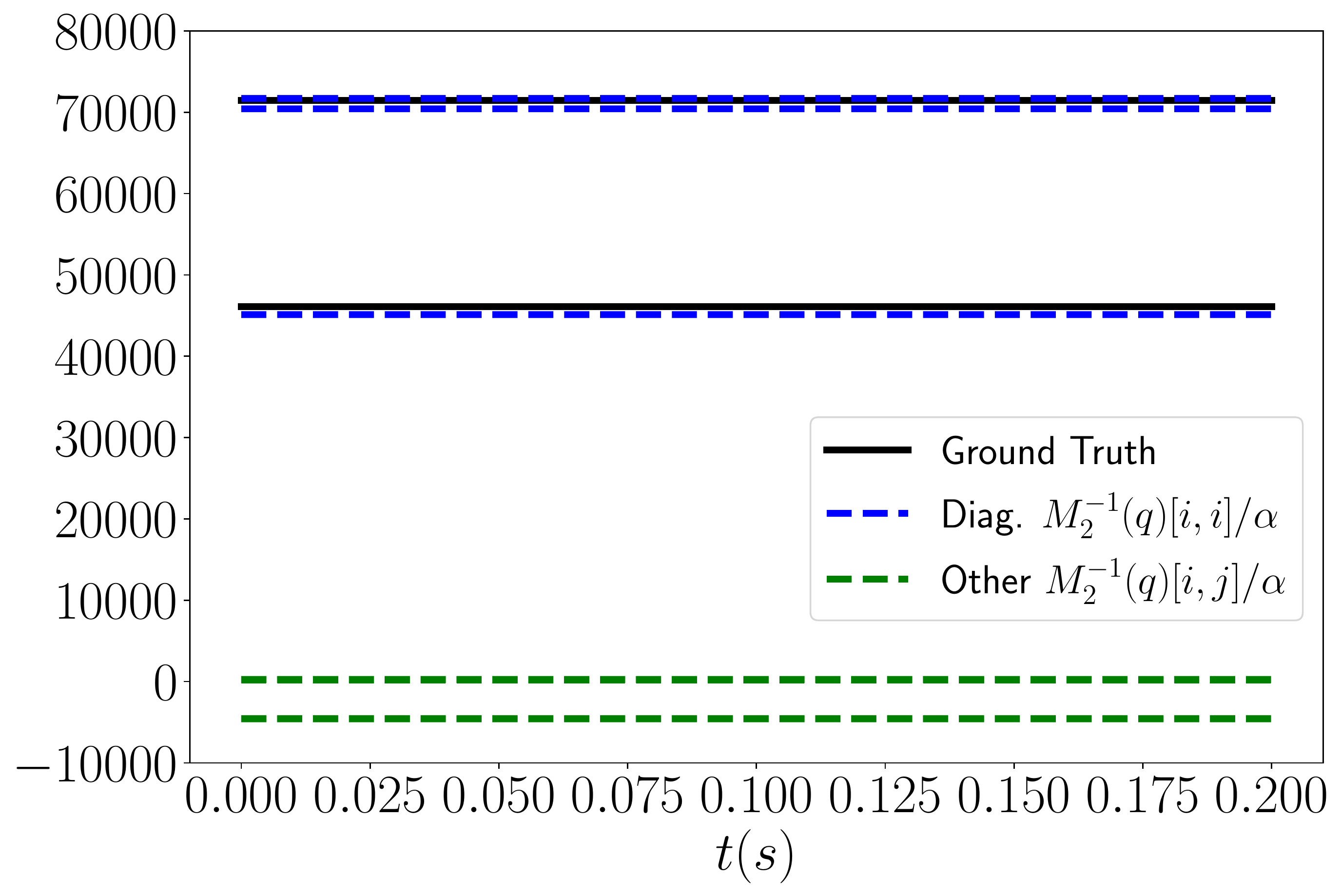}%
        \caption{$\bfM_2^{-1}(q)/\alpha$ over time.}
        \label{fig:rigid_M2_x_all}
\end{subfigure}%
\hfill%
\begin{subfigure}[t]{0.24\textwidth}
        \centering
        \includegraphics[width=\textwidth]{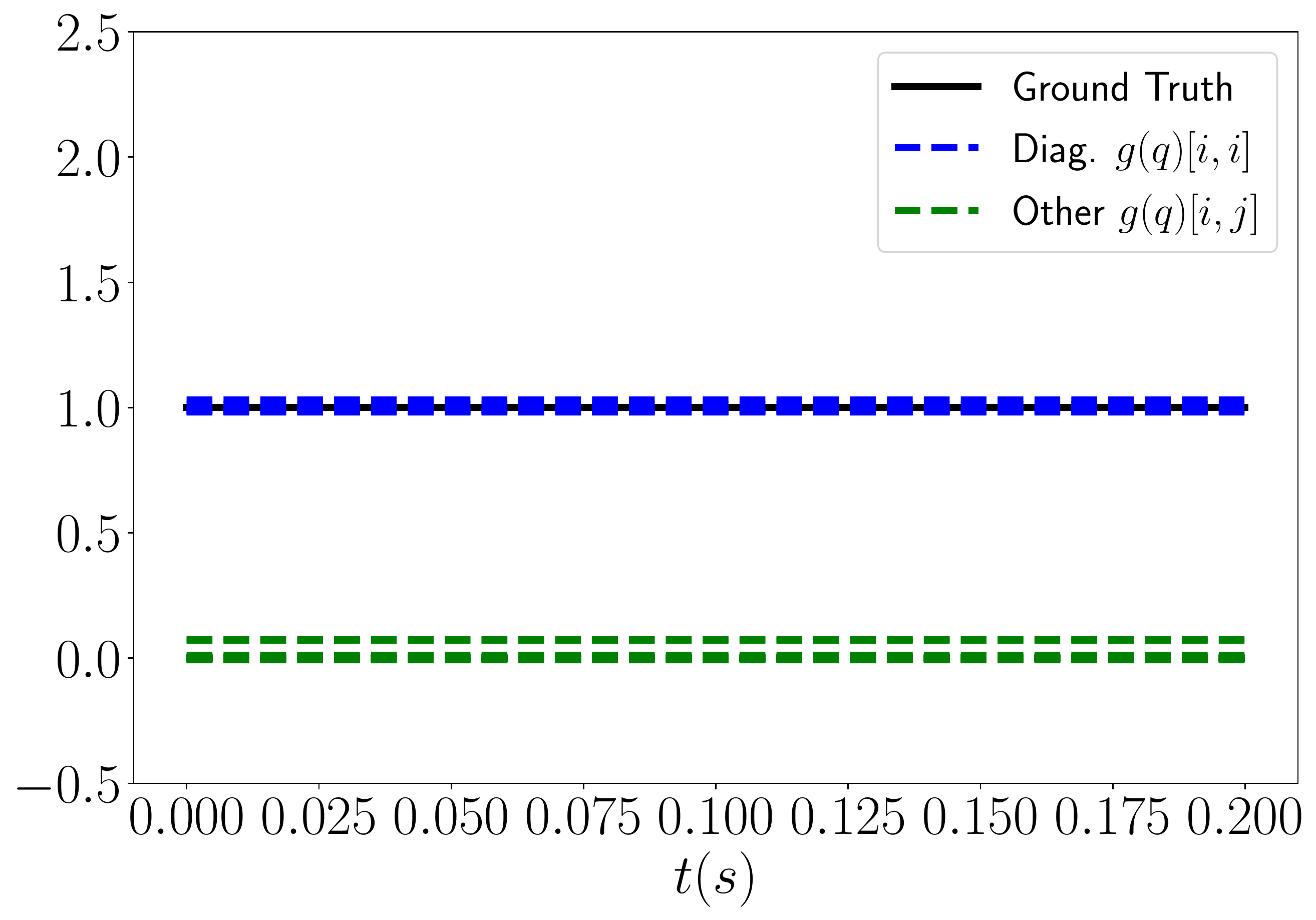}
        \caption{Scaled $\bfg(q)$ over time.}
        \label{fig:rigid_g_x_all}
\end{subfigure}%

\begin{subfigure}[t]{0.228\textwidth}
        \centering
\includegraphics[width=\textwidth]{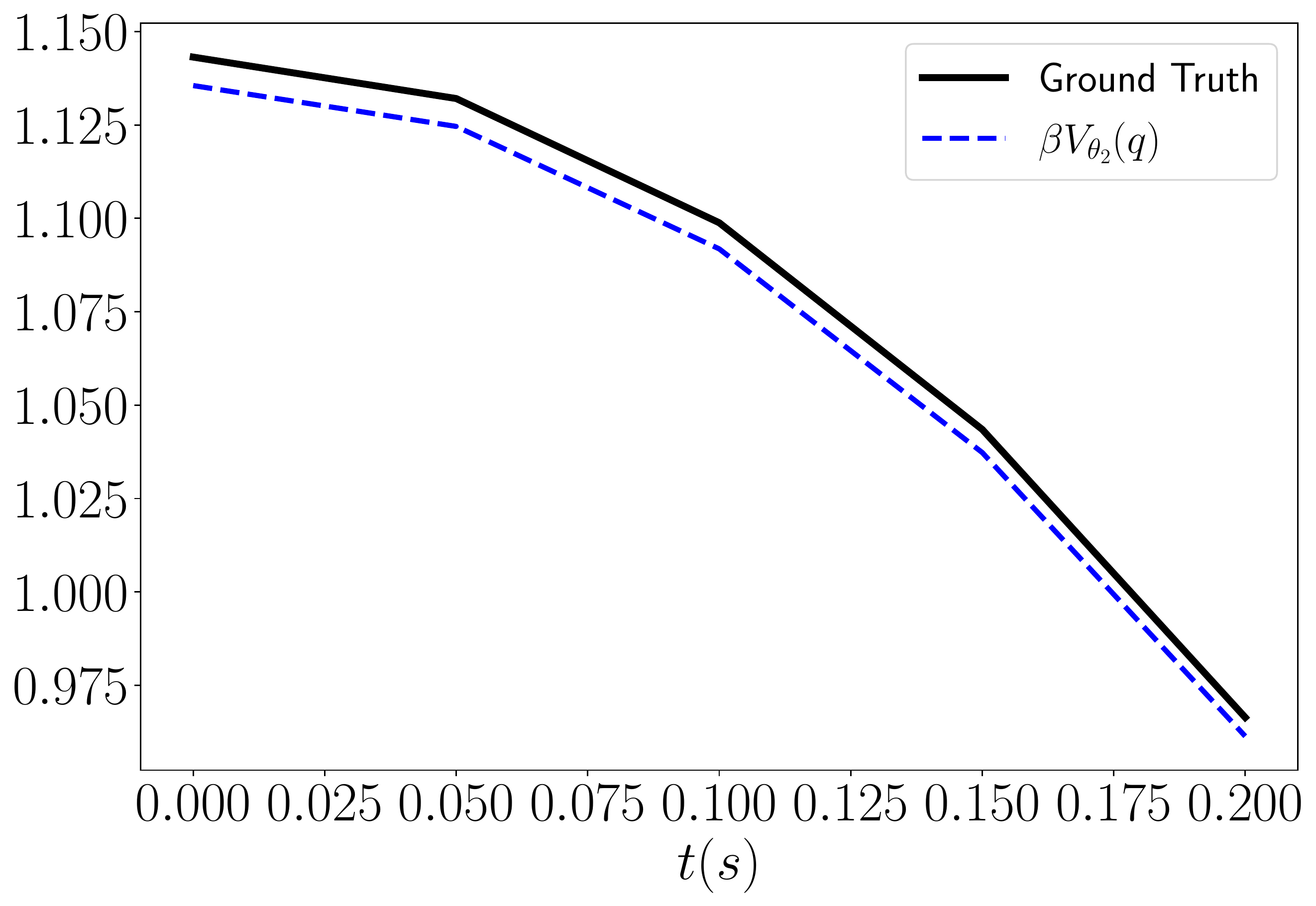}%
        \caption{$\beta V(q)$ over time.}
        \label{fig:rigid_Vx}
\end{subfigure}%
\hfill%
\begin{subfigure}[t]{0.24\textwidth}
        \centering
\includegraphics[width=\textwidth]{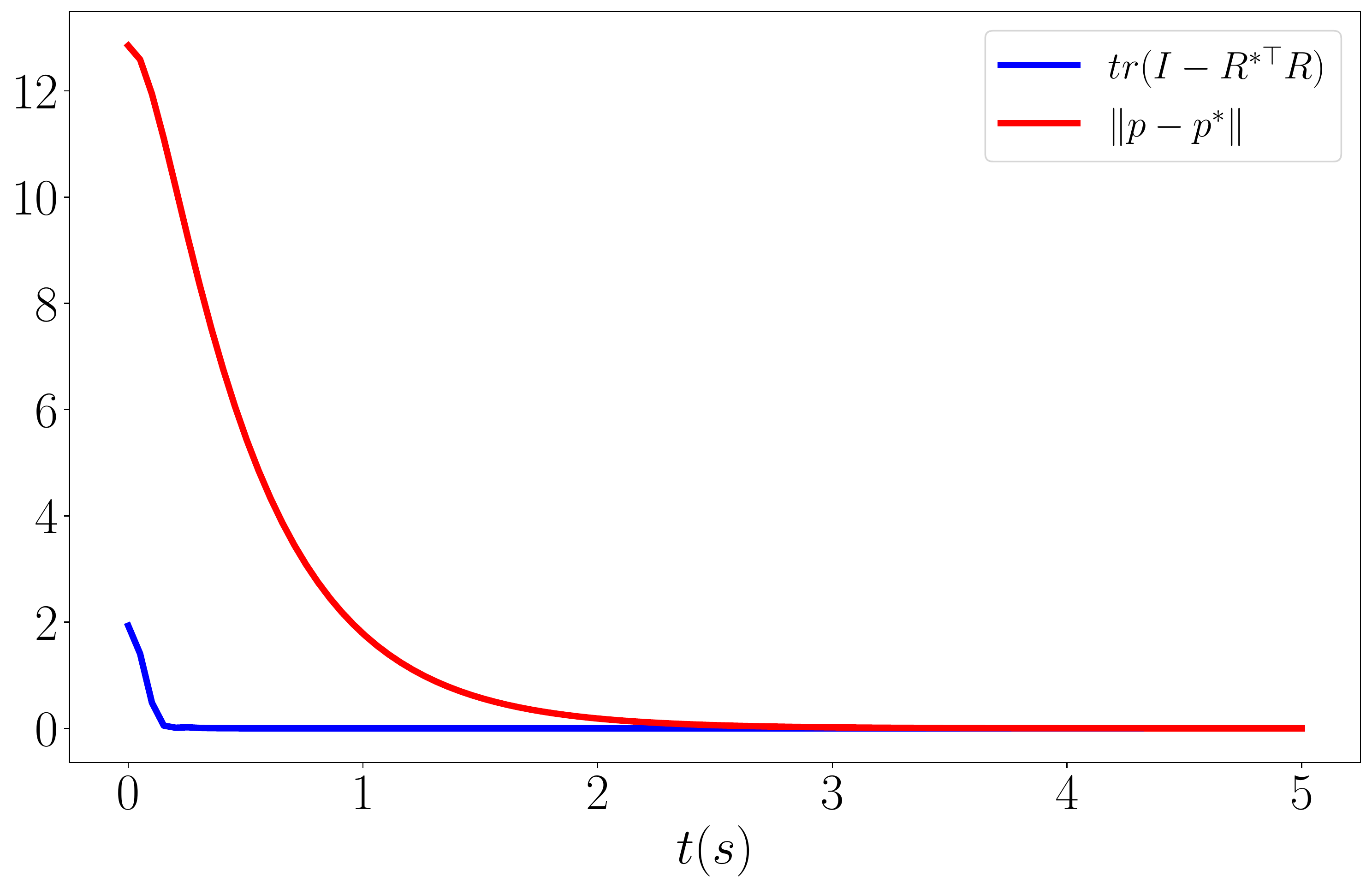}%
        \caption{Controlled state errors.}
        \label{fig:rigid_control_errors}
\end{subfigure}%
\hfill%
\begin{subfigure}[t]{0.24\textwidth}
        \centering
\includegraphics[width=\textwidth]{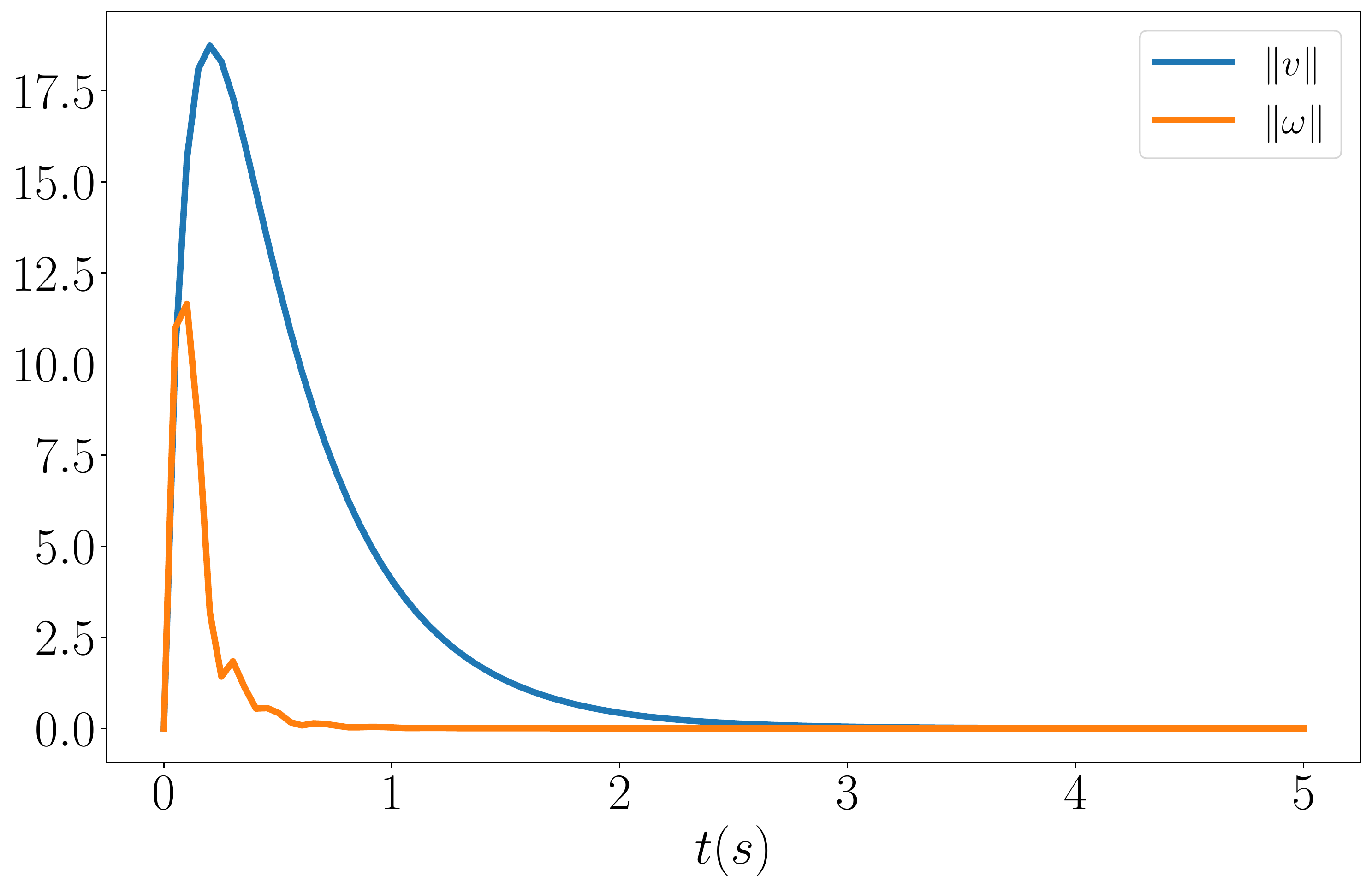}%
        \caption{Controlled velocity.}
        \label{fig:rigid_control_vel}
\end{subfigure}%
\hfill%
\begin{subfigure}[t]{0.24\textwidth}
        \centering
\includegraphics[width=\textwidth]{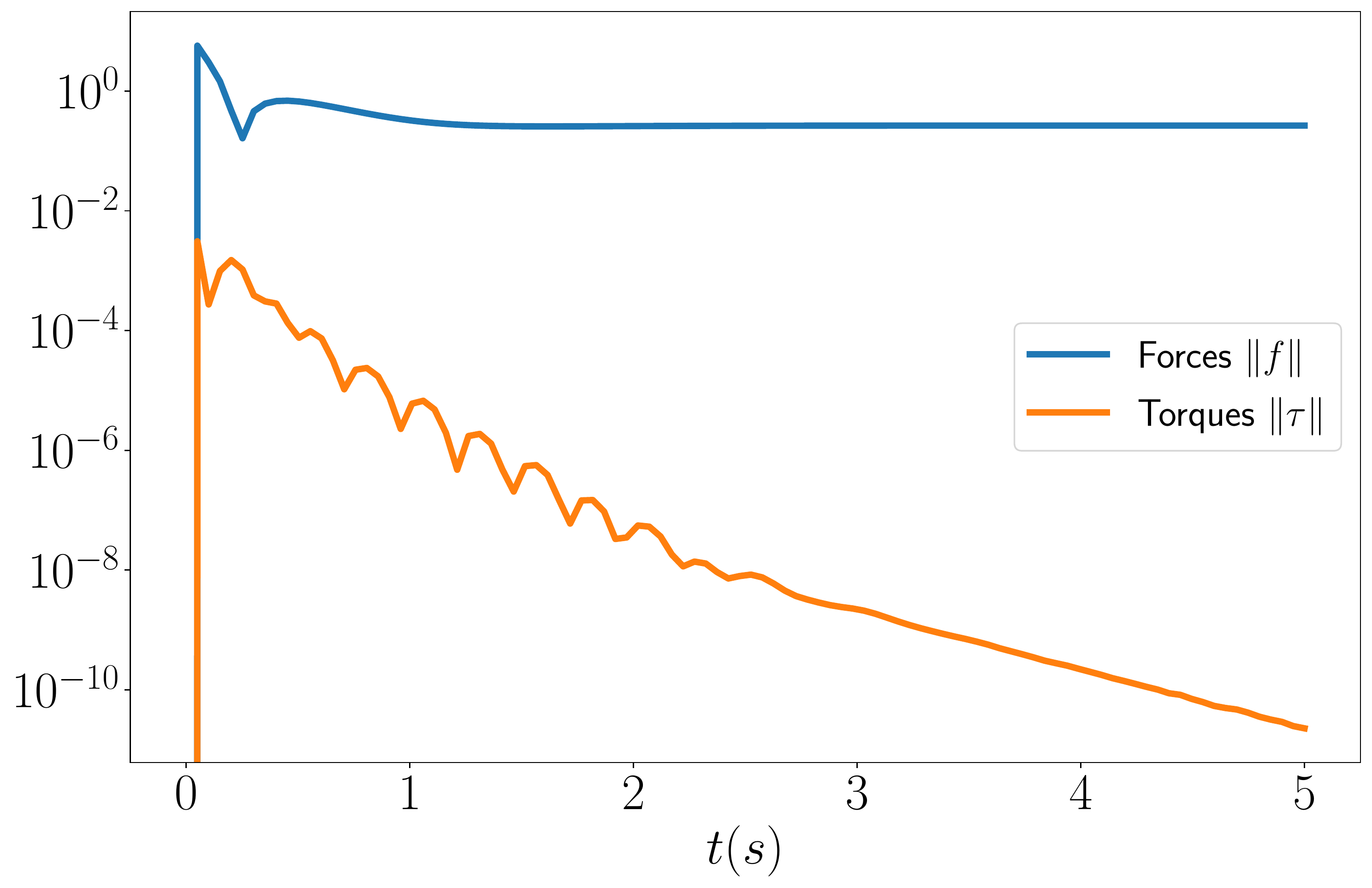}%
        \caption{Control input $\bfu$.}
        \label{fig:rigid_control_input}
\end{subfigure}%
\caption{Evaluation of our $SE(3)$ Hamiltonian neural ODE network on a fully-actuated rigid body with scaling $\alpha = 0.0032$ and $\beta = 0.864$.}
\label{fig:rigid_exp}
\end{figure*}

%In this experiment, we considered a fully-actuated rigid body that has the  weight $m = 0.027$ and inertia matrix $\bfJ = 10^{-5}\text{diag}([1.4, 1.4, 2.17])$, mimicking a Crazyflie drone. 

Next, we consider a fully-actuated rigid body with mass $m = 0.027$ and inertia matrix $\bfJ = 10^{-5}\text{diag}([1.4, 1.4, 2.17])$. This can be viewed as an abstraction of any mobile robot that can be modeled as a single rigid body, such as car-like, fixed-wing, and quadrotor robots. For example, a hexarotor UAV with fixed-tilt rotors pointing in different directions is fully actuated \cite{hexarotor}. The ground-truth dynamics follow \eqref{eq:hamiltonian_pose_pdot}, \eqref{eq:hamiltonian_pose_rdot}, \eqref{eq:hamiltonian_pose_pvdot}, and \eqref{eq:hamiltonian_pose_pomegadot} with generalized coordinates $\mathbf\frakq = [\bfp^\top \quad \bfr_1^\top \quad \bfr_2^\top \quad \bfr_3^\top]^\top \in \mathbb{R}^{12}$, generalized velocity $\bfzeta = [\bfv^\top \quad \bfomega^\top]^\top \in \mathbb{R}^6$, generalized mass $\bfM_1(\mathbf\frakq) = m\bfI$, $\bfM_2(\mathbf\frakq) = \bfJ$, potential energy $V(\mathbf\frakq) = mgz$, and the input matrix $\bfg(\mathbf\frakq) = \bfI$. The control input $\bfu$ is a 6-dimensional wrench (i.e., 3-dimensional force and 3-dimensional torque). 

As described in Sec. \ref{subsec:data_gen}, control inputs $\bfu^{(i)}$ were sampled randomly and applied to the system for one time step of $0.05s$, forming a dataset $\mathcal{D} = \left\{t_{0:N}^{(i)},\mathbf\frakq_{0:N}^{(i)}, \bfzeta_{0:N}^{(i)}, \bfu^{(i)})\right\}_{i = 1}^D$ with $N = 1$ and $D = 11520$. The $SE(3)$ ODE network was trained as described in Sec. \ref{sec:training} for 6000 iterations. The neural networks modeling $\bfM_1(\mathbf\frakq)$ and $\bfM_2(\mathbf\frakq)$ were pre-trained to output an identity matrix.

%Note that we convert the linear velocity in the world frame $\bfv$ to the linear velocity in the body frame using  \eqref{eq:linvel_body_world}. 
\begin{figure}[t]
\centering
\begin{subfigure}[t]{0.247\textwidth}
        \centering
\includegraphics[width=\textwidth]{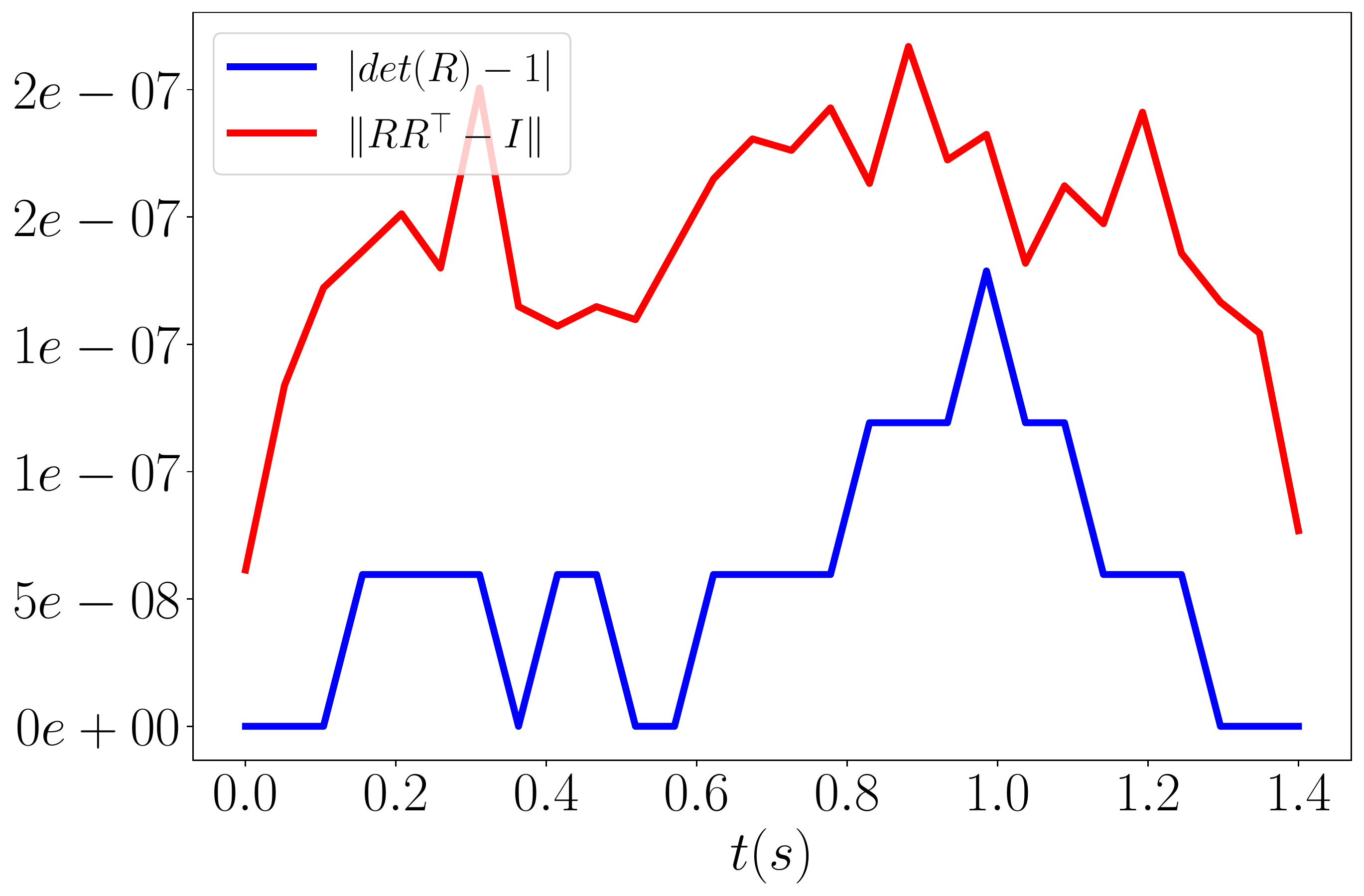}%
        \caption{SO(3) constraints.}
        \label{fig:rigid_so3_constraints}
\end{subfigure}%
\begin{subfigure}[t]{0.23\textwidth}
        \centering
\includegraphics[width=\textwidth]{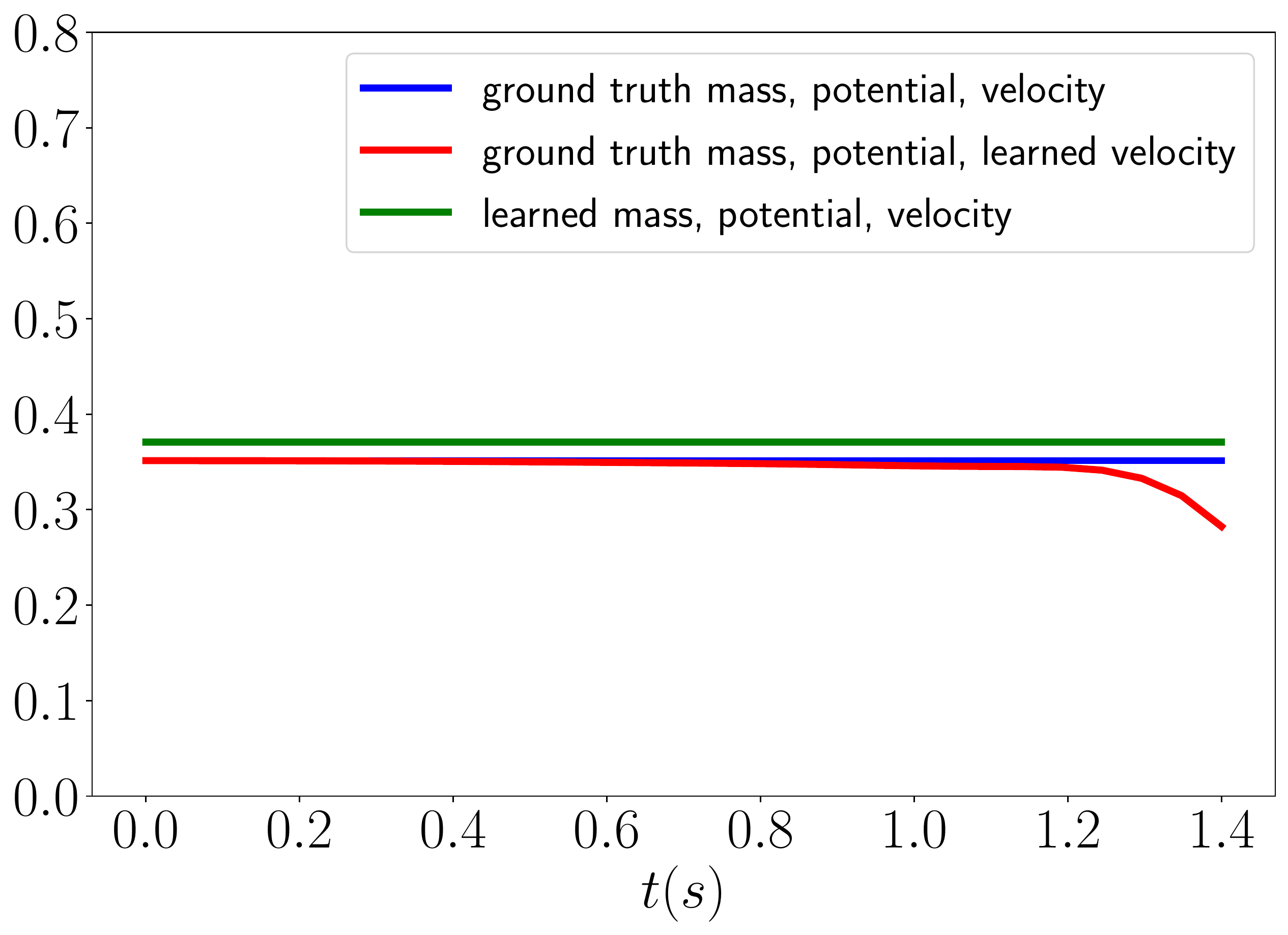}%
        \caption{Total energy.}
        \label{fig:rigid_total_energy}
\end{subfigure}%
\caption{$SO(3)$ constraints and total energy along a trajectory rollout with zero control input from the learned rigid body model.}
\label{fig:rigid_se3_totalenergy}
\end{figure}

Similar to the pendulum system in Sec. \ref{subsec:pendulum_so3}, the dynamics of $\mathbf\frakq$ do not change if $\mathbf\frakp$ is scaled. In fact, as the ground-truth generalized mass is $\bfM_1(\mathbf\frakq) = m\bfI$, and the potential energy $V(\mathbf\frakq) = mgz$ only depends on the position $\bfp$, the generalized momenta $\mathbf\frakp_{\bfomega}$ and $\mathbf\frakp_{\bfv}$ can be scaled, respectively, by two different factors $\alpha >0$ and $\beta >0$. In other words, $\bfM_1(\mathbf\frakq)$, $V(\mathbf\frakq)$, and $\bfg_{\bfv}(\mathbf\frakq)$ can be scaled by $\beta$ and $\bfM_2(\mathbf\frakq)$ and $\bfg_{\bfomega}(\mathbf\frakq)$ can be scaled by $\alpha$ without changing the dynamics of $\mathbf\frakq$ and $\bfzeta$ in \eqref{eq:hamiltonian_pose_pdot}, \eqref{eq:hamiltonian_pose_rdot}, and \eqref{eq:hamiltonian_zetadot}.
%using similar derivation shown for the pendulum experiment in Sec. \ref{subsec:pendulum_so3}. In fact, we have two scaling factors $\alpha$ and $\beta$ for $\mathbf\frakp_{\bfomega}$ and $\mathbf\frakp_{\bfv}$, respectively, in this experiment. 

%The results are shown in Fig. \ref{fig:rigid_exp} with scaling factor $\beta = 0.864$ and $\alpha = 0.0032$. 

Fig.~\ref{fig:rigid_exp} shows the training and testing performance of the $SE(3)$ Hamiltonian ODE network. The training loss is shown in Fig.~\ref{fig:rigid_loss} in log scale. Fig. \eqref{fig:rigid_M1_x_all}, \ref{fig:rigid_M2_x_all}, and \ref{fig:rigid_g_x_all} show that the diagonal entries of the scaled mass matrices $\bfM_1(\mathbf\frakq)$, $\bfM_2(\mathbf\frakq)$ and the input matrix $\bfg(\mathbf\frakq)$ converge to the ground-truth values while the remaining entries are close to $\bf0$. Fig. \ref{fig:rigid_Vx} shows that the learned potential energy closely resembles the ground-truth values up to scaling.

We tested regulation of the fully-actuated rigid body based on its learned dynamics from an initial position $\bfp = \bf0$ and randomized initial orientation to a desired position $\bfp^* = [1, 2, 5]$ and desired orientation $\bfR^* = \bfI$. As for the pendulum, potential energy shaping was enough to drive the system to a desired state $(\mathbf\frakq^*, 0)$ and was achieved by setting $\bfJ_2 = \mathbf\frakp^\times$, $\bfM_d(\mathbf\frakq) = \bfM(\mathbf\frakq)$ in \eqref{eq:general_u_ES}. The additional energy $\mathcal{H}_a(\mathbf\frakq, \mathbf\frakp)$ was simplified to:
\begin{eqnarray}
\mathcal{H}_a(\mathbf\frakq, \mathbf\frakp) &=& -V(\mathbf\frakq) +  \frac{1}{2}(\bfp - \bfp^*)^\top\bfK_\bfp(\bfp - \bfp^*) \nonumber \\
&& + \frac{1}{2} \tr(\bfK_{\bfR}(\bfI - \bfR^{*\top}\bfR)).
\end{eqnarray}

The controller becomes $\bfu = \bfu_{ES} + \bfu_{DI}$ where:
\begin{eqnarray}
\bfu_{ES}(\mathbf\frakq, \mathbf\frakp) &=& \bfg_{\bftheta_3}^{\dagger}(\mathbf\frakq) \begin{bmatrix}
\sum_{i = 1}^3 \bfr_i \times \frac{\partial \mathcal{H}_a}{\partial\bf\bfr_i} \\
- \bfR^\top \frac{\partial \mathcal{H}_a}{\partial\bf\bfp}
\end{bmatrix}, \label{eq:u_es} \\
\bfu_{DI}(\mathbf\frakq, \mathbf\frakp) &=& -\bfg_{\bftheta_3}^\dagger(\mathbf\frakq)\bfK_\bfd\bfzeta. \label{eq:u_di}
\end{eqnarray}
With control gains $\bfK_{\bfp} = \bfK_\bfR = 0.5 \bfI$ and $\bfK_\bfd = 0.25\text{diag}(10^{-3}, 10^{-3}, 10^{-3}, 1.0, 1.0, 1.0)$, the \mbox{errors}~$\text{tr}(\bfI~-~\bfR^{*\top}\bfR)$ and $\Vert \bfp - \bfp^* \Vert$ and the velocities $\bfv$, $\bfomega$ go to $0$ and regulation is achieved successfully, as shown in Fig. \ref{fig:rigid_control_errors} and \ref{fig:rigid_control_vel}. Fig. \ref{fig:rigid_control_input} shows the control input $\bfu$, in which the torque component goes to $\bf0$ while the force component approaches a constant value compensating for gravity.

As for the pendulum, we also verified that predicted orientation trajectories from the learned model satisfy the $SO(3)$ constraints. Fig. \ref{fig:rigid_so3_constraints} shows $|\det{\bfR} - 1|$ and $\Vert \bfR \bfR^\top - \bfI \Vert$ along a trajectory rollout. As before, we also calculated the Hamiltonian via \eqref{eq:hamiltonian_def} along the trajectory with no control input using: 1) ground-truth mass, potential energy and velocity; 2) ground-truth mass and potential energy but velocity rolled out from the learned dynamics; and 3) all mass, potential energy and velocity rolled out from the learned dynamics. Fig. \ref{fig:pend_total_energy} shows that the Hamiltonian stays constant for cases 1) and 3) but not for case 2) after $25$ time steps. This indicates that the learned dynamics, by design, are guaranteed to obey the law of energy conservation with respect to the learned mass and potential energy, rather than the ground-truth ones. This discrepancy can be fixed by increasing the length of the training sequence $N$ (in this experiment, $N = 1$) or by avoiding rollout predictions of more than $25$ steps with the model without additional training.
% or by updating the velocity using a localization algorithm or a motion capture system.

%This means our neural ODE framework for SE(3) manifold never violates the SO(3) constraints for rotation matrices $\bfR$.

\subsection{Crazyflie Quadrotor}

\begin{figure*}[h]
\begin{subfigure}[t]{0.239\textwidth}
        \centering
        \includegraphics[width=\textwidth]{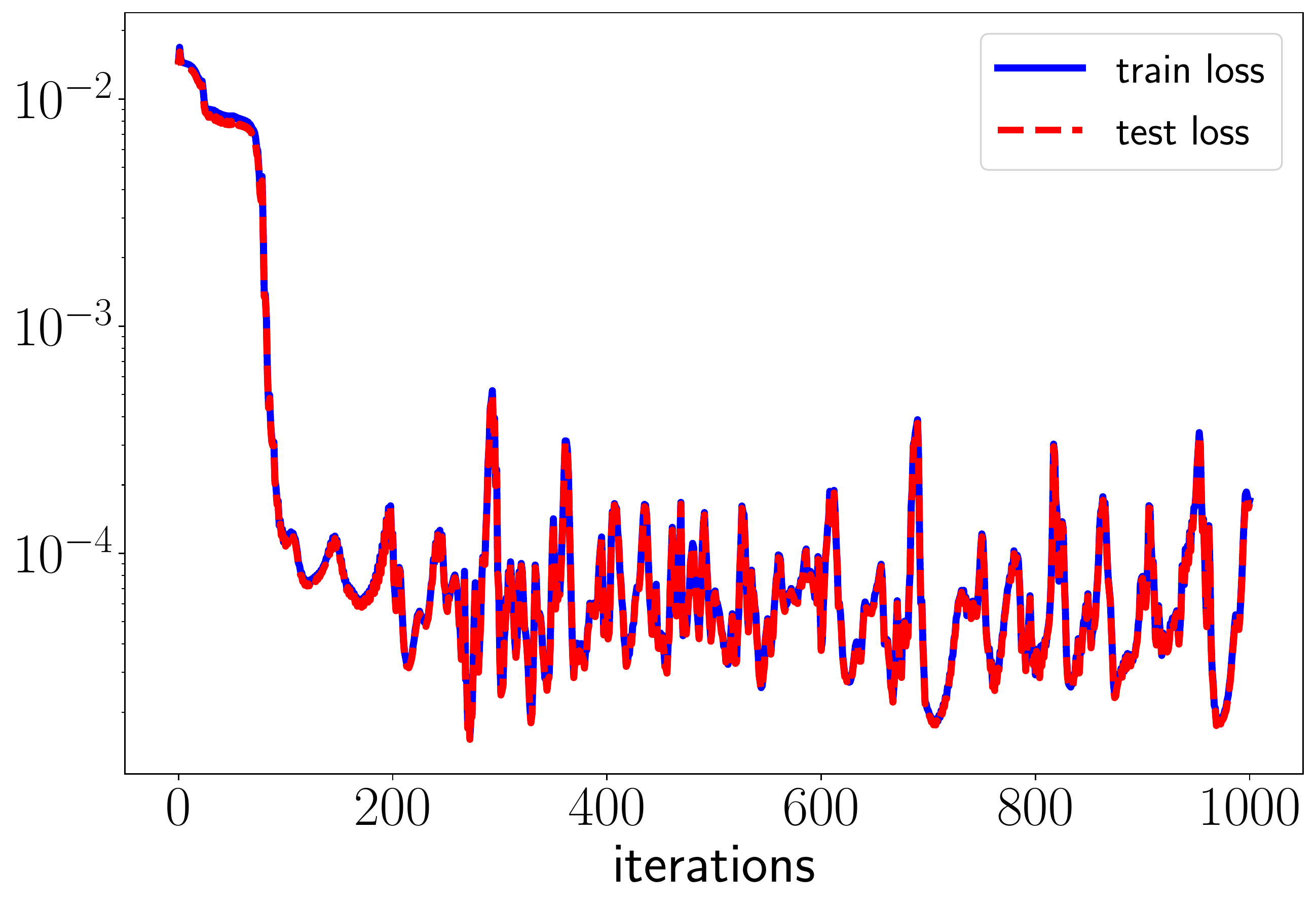}
        \caption{Loss (log scale)}
        \label{fig:pybullet_loss}
\end{subfigure}%
\begin{subfigure}[t]{0.241\textwidth}
        \centering
        \includegraphics[width=\textwidth]{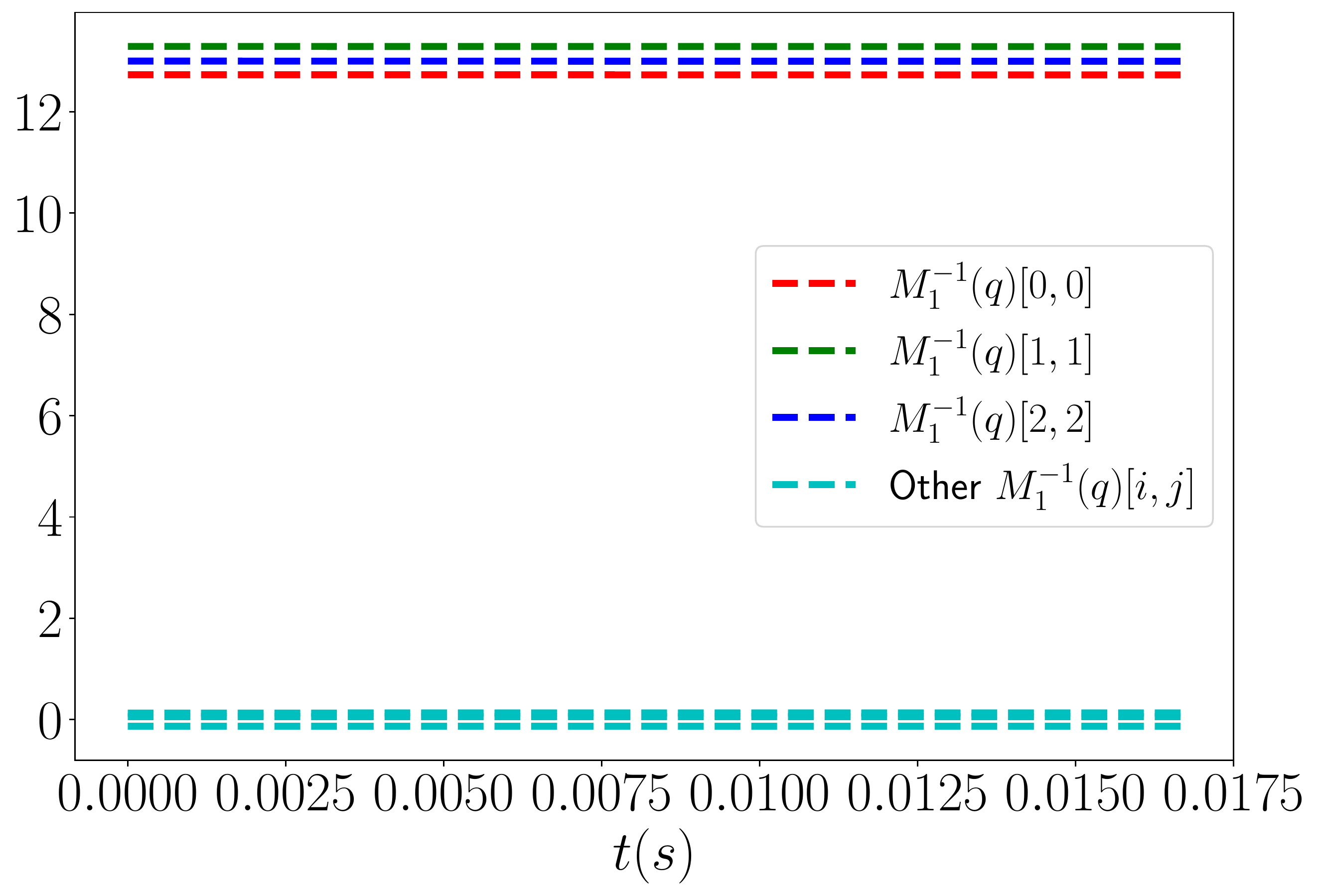}
        \caption{$\bfM_1^{-1}(q)$}
        \label{fig:pybullet_M1_x_all}
\end{subfigure}%
\begin{subfigure}[t]{0.24\textwidth}
        \centering
		\includegraphics[width=\textwidth]{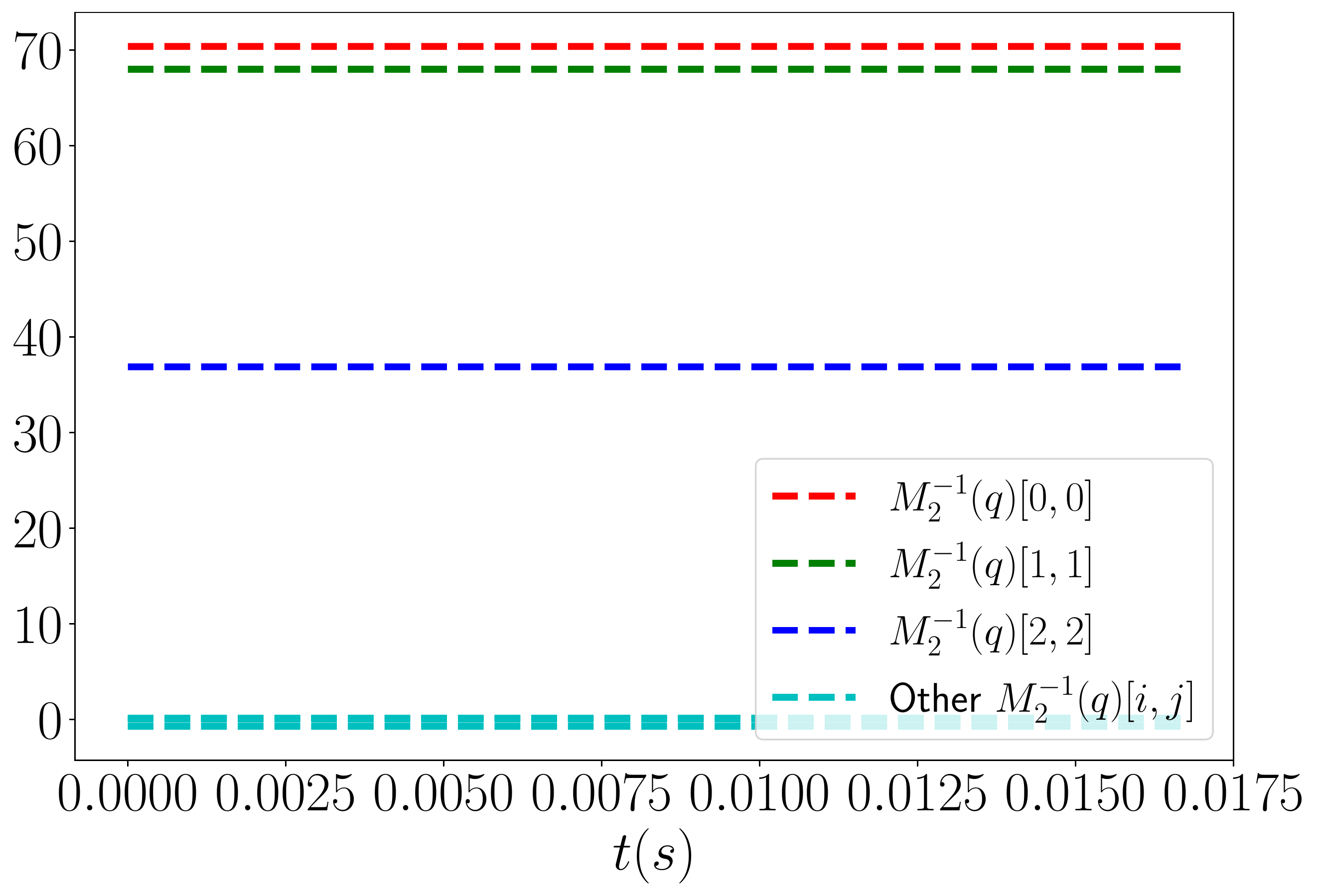}%
        \caption{$\bfM_2^{-1}(q)$}
        \label{fig:pybullet_M2_x_all}
\end{subfigure}%
\begin{subfigure}[t]{0.248\textwidth}
        \centering
\includegraphics[width=\textwidth]{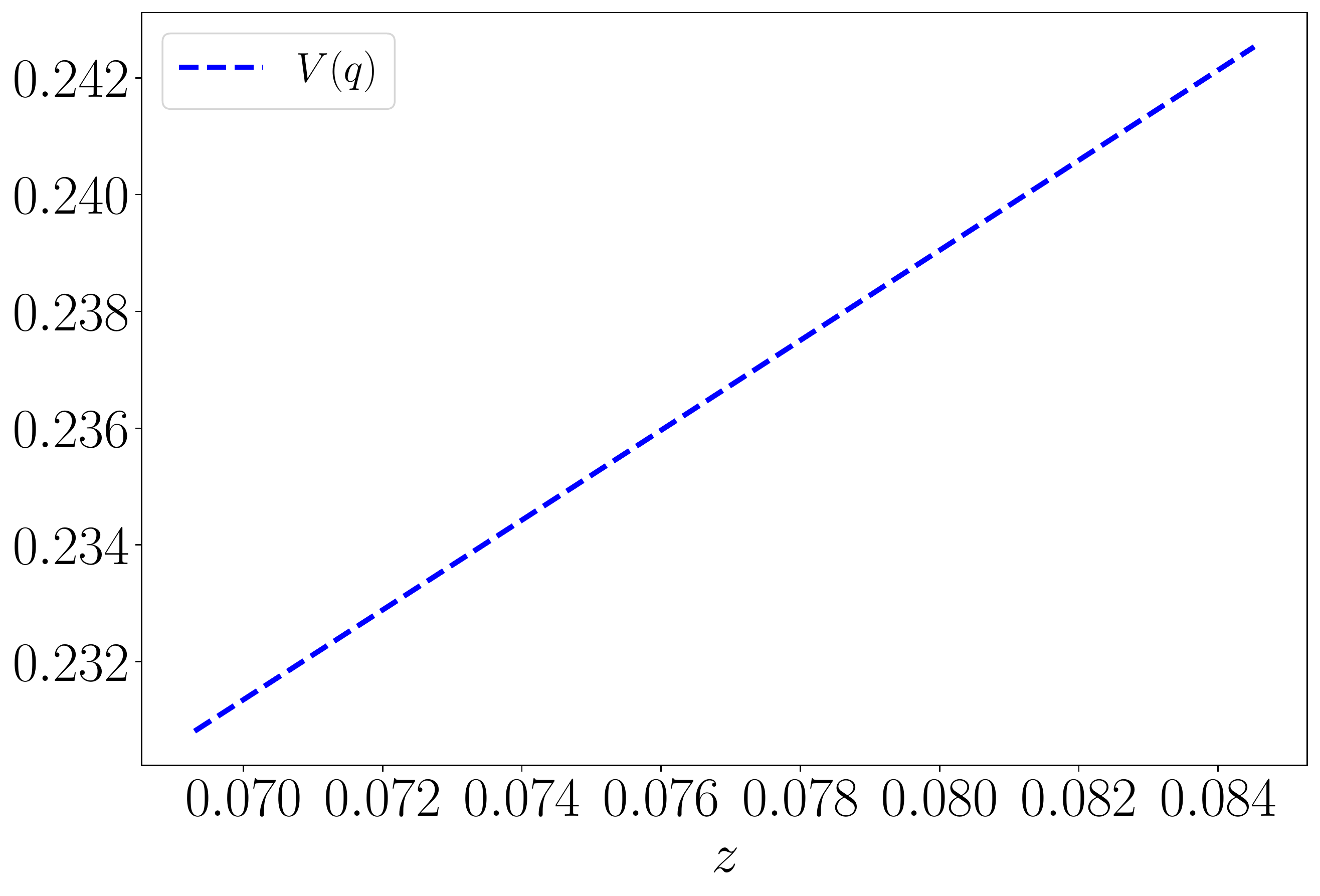}%
        \caption{$V(q)$}
        \label{fig:pybullet_Vx}
\end{subfigure}%

\begin{subfigure}[t]{0.24\textwidth}
        \centering
        \includegraphics[width=\textwidth]{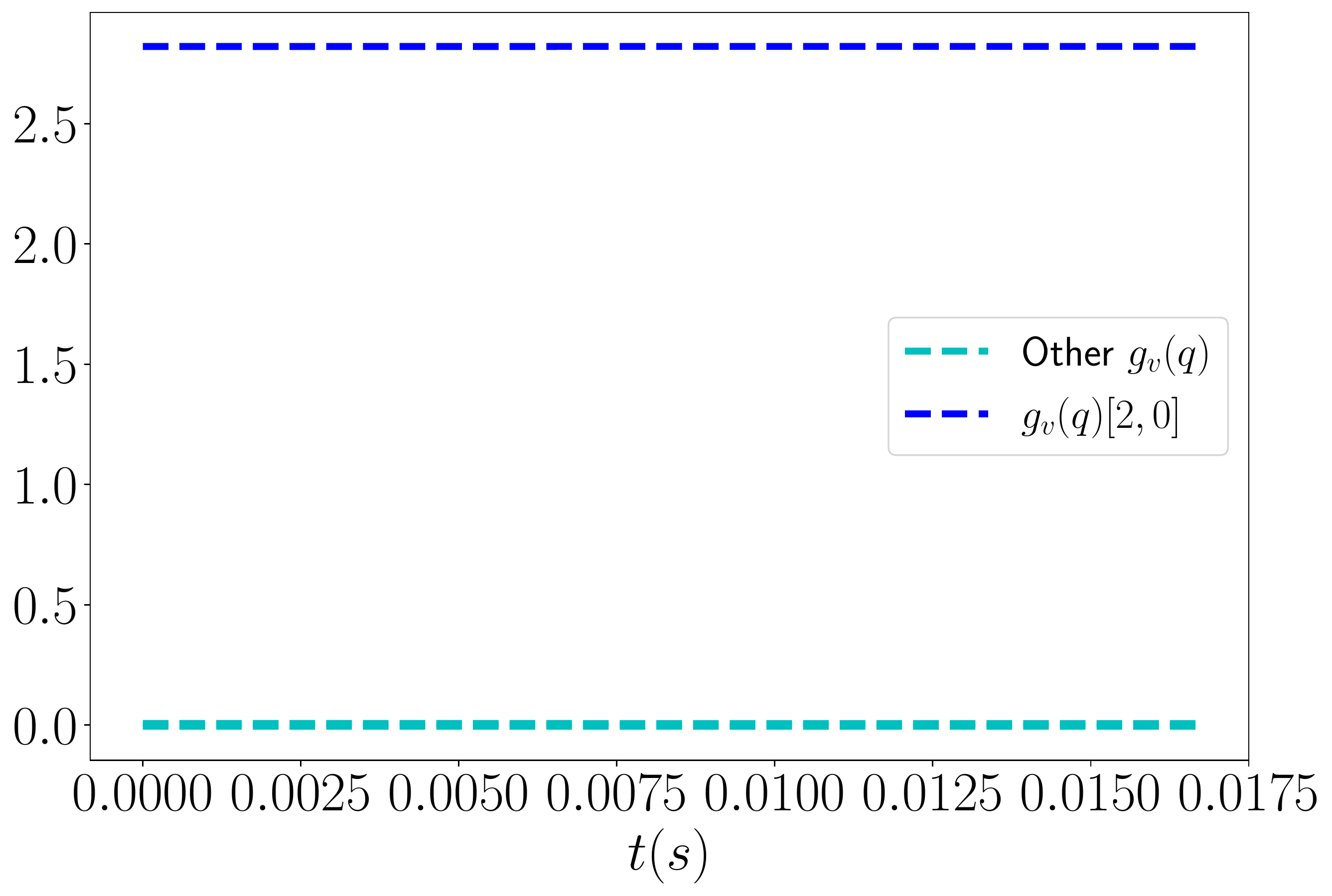}
        \caption{$\bfg_{\bfv}(q)$}
        \label{fig:pybullet_g_v_x_all}
\end{subfigure}%
\begin{subfigure}[t]{0.24\textwidth}
        \centering
        \includegraphics[width=\textwidth]{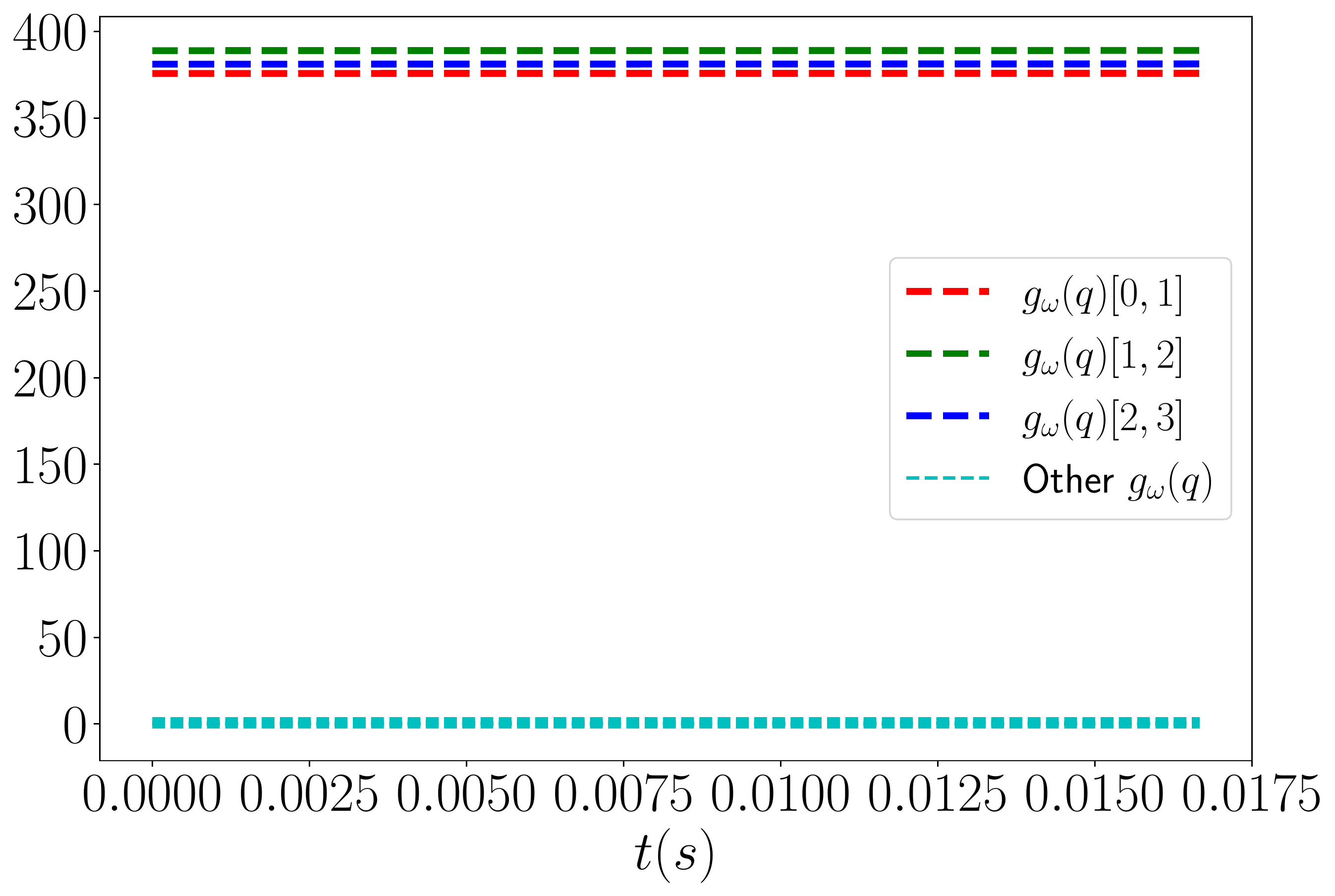}
        \caption{$\bfg_{\bfomega}(q)$}
        \label{fig:pybullet_g_w_x_all}
\end{subfigure}%
\begin{subfigure}[t]{0.24\textwidth}
        \centering
\includegraphics[width=\textwidth]{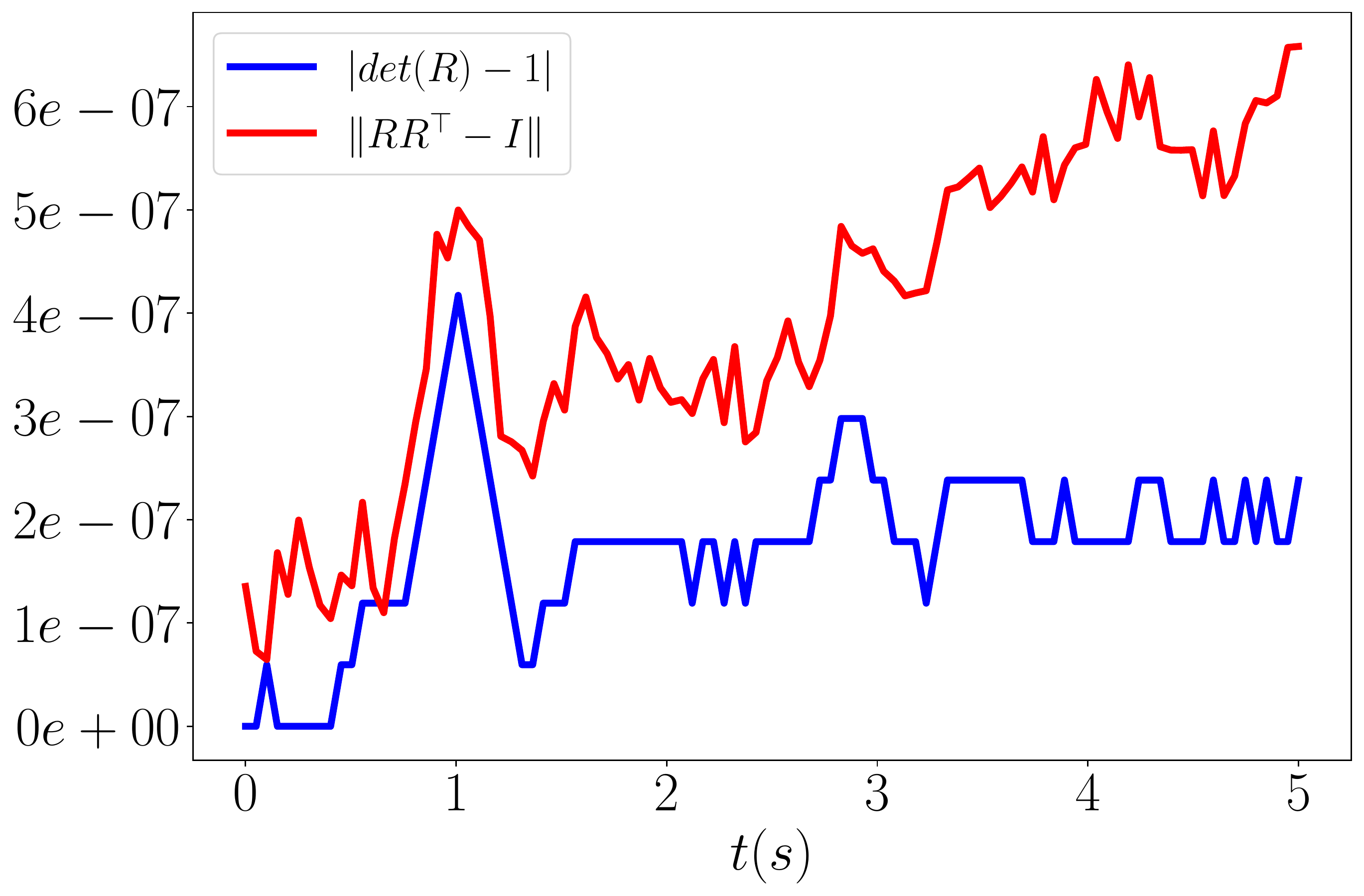}%
        \caption{SO(3) constraints.}
        \label{fig:pybullet_so3_constraints}
\end{subfigure}%
\begin{subfigure}[t]{0.24\textwidth}
        \centering
\includegraphics[width=\textwidth]{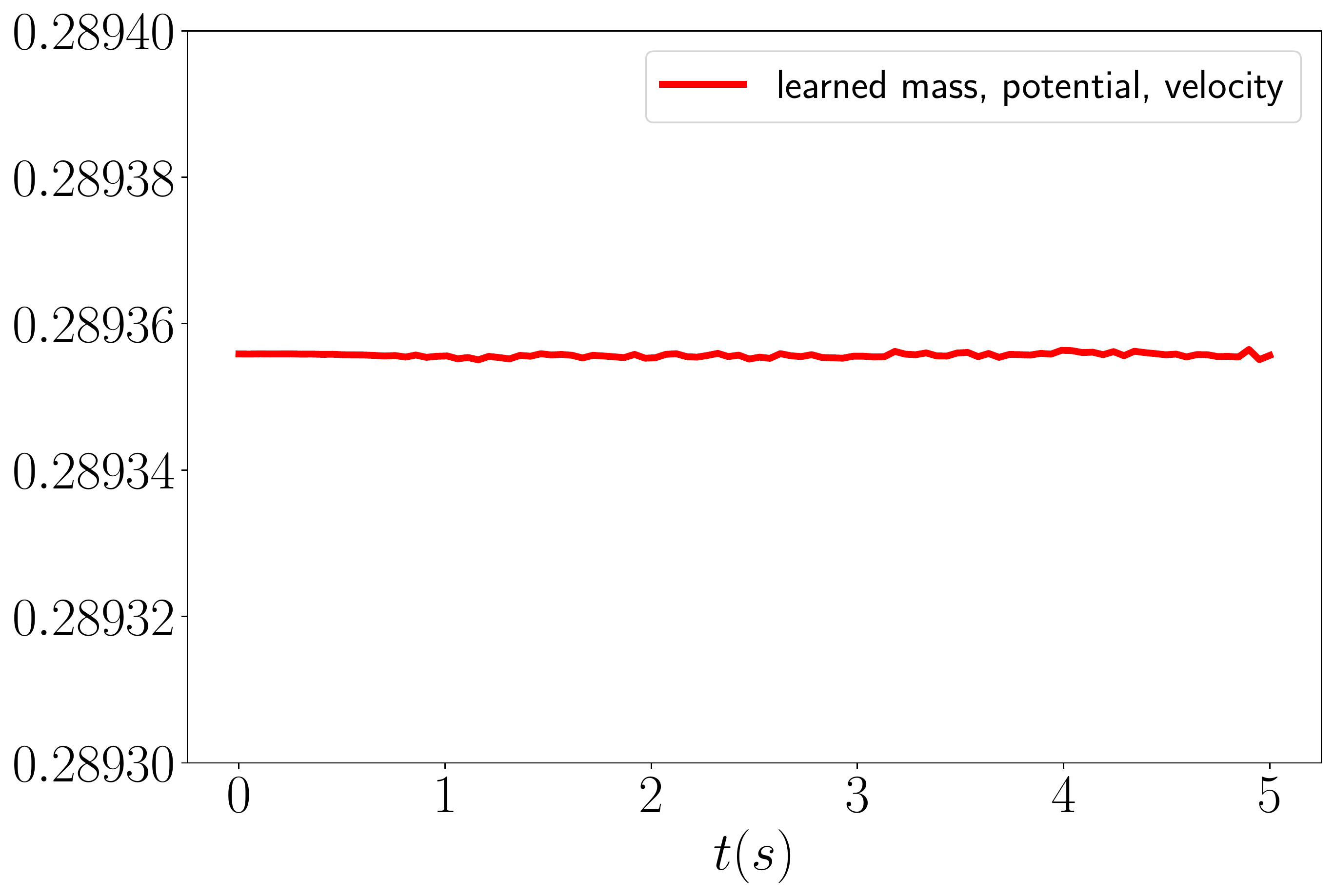}%
        \caption{Total energy.}
        \label{fig:pybullet_total_energy}
\end{subfigure}%
\caption{Evaluation of the $SE(3)$ Hamiltonian neural ODE network on an under-actuated Crazyflie quadrotor in the PyBullet simulator \cite{gym-pybullet-drones2020}.}
\label{fig:pybullet_exp}
\end{figure*}

\begin{figure}[t]
\centering
\includegraphics[width=0.5\textwidth]{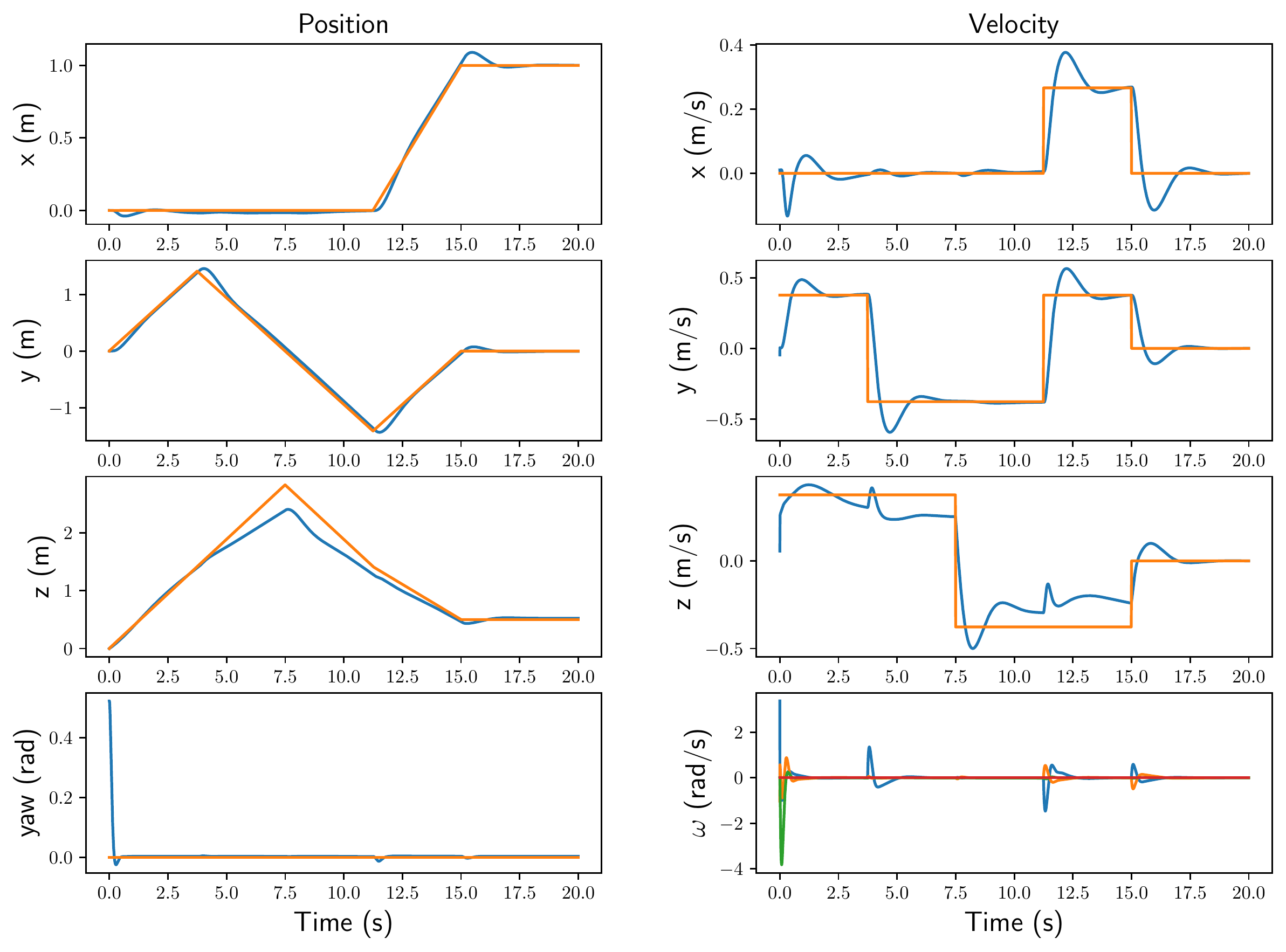}%
\caption{Crazyflie quadrotor trajectory (blue) tracking a desired diamond-shaped trajectory (orange) shown in Fig.~\ref{fig:pybullet_traj_viz}.}
\label{fig:pybullet_tracking_results}
\end{figure}

\begin{figure}[t]
\centering
\begin{subfigure}[t]{0.22\textwidth}
        \centering
\includegraphics[width=\textwidth]{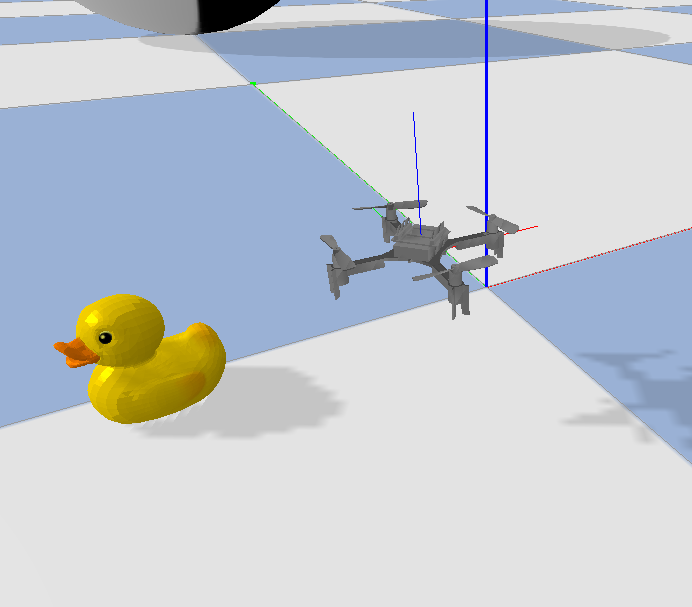}%
        \caption{Crazyflie simulator}
        \label{fig:pybullet_crazyflie}
\end{subfigure}%
\begin{subfigure}[t]{0.26\textwidth}
        \centering
\includegraphics[width=\textwidth]{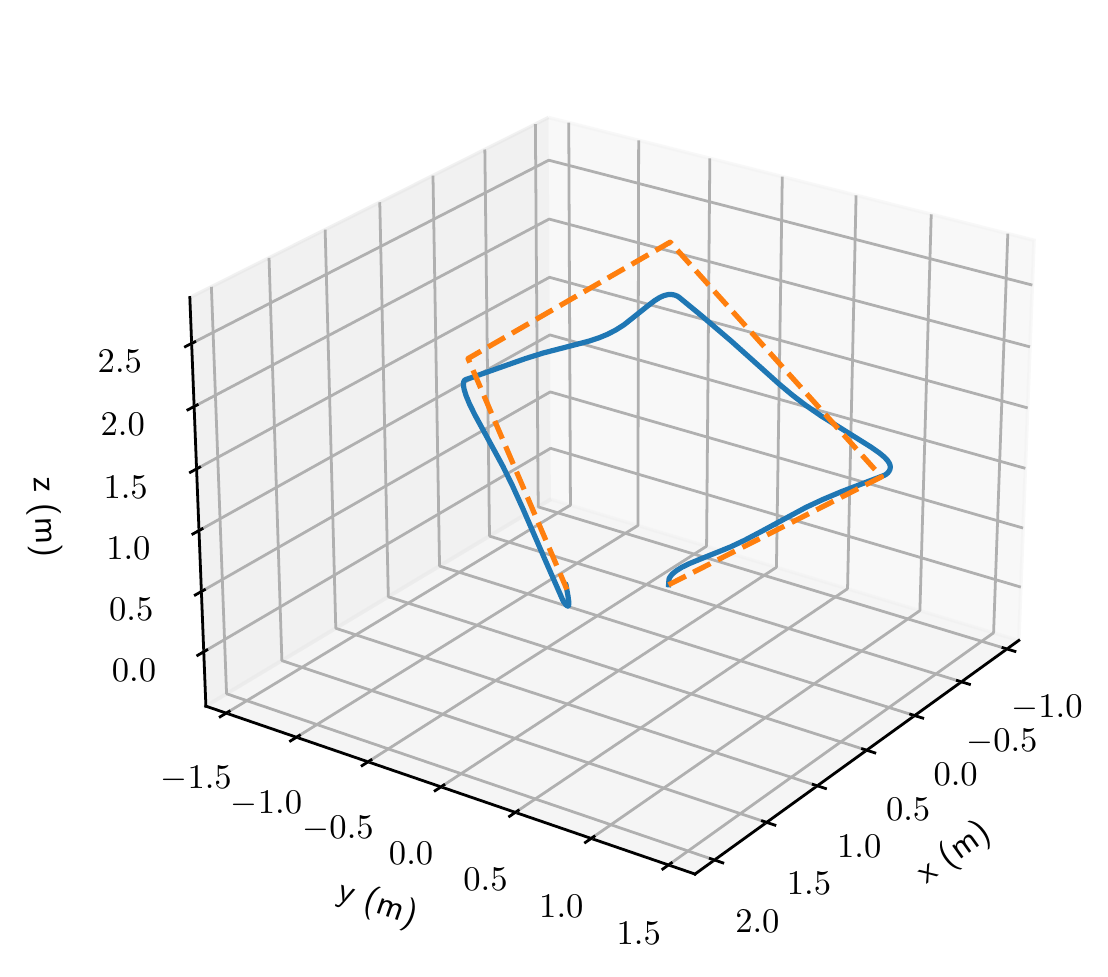}%
        \caption{Trajectory tracking}
        \label{fig:pybullet_trajviz}
\end{subfigure}%
\caption{Trajectory tracking experiment with a Crazyflie quadrotor in the PyBullet simulator \cite{gym-pybullet-drones2020}.}
\label{fig:pybullet_traj_viz}
\end{figure}

% This is an under-actuated system where the control input $\bfu = [f, \bftau]$ includes a thrust $f\in \mathbb{R}_{\geq 0}$ and a torque vector $\bftau \in \mathbb{R}^3$ generated by $4$ rotors.

Finally, we demonstrate that our $SE(3)$ dynamics learning and control approach can achieve trajectory tracking for an under-actuated system. We consider a Crazyflie quadrotor, shown in Fig. \ref{fig:pybullet_crazyflie}, simulated in the physics-based simulator PyBullet \cite{gym-pybullet-drones2020}. The control input $\bfu = [f, \bftau]$ includes a thrust $f\in \mathbb{R}_{\geq 0}$ and a torque vector $\bftau \in \mathbb{R}^3$ generated by the $4$ rotors. The generalized coordinates and velocity are $\mathbf\frakq = [\bfp^\top \quad \bfr_1^\top \quad \bfr_2^\top \quad \bfr_3^\top]^\top \in \mathbb{R}^{12}$ and $\bfzeta = [\bfv^\top \quad \bfomega^\top]^\top \in \mathbb{R}^{6}$ as before.

The quadrotor was controlled from a random starting point to $18$ different desired poses using a PID controller provided by \cite{gym-pybullet-drones2020}, providing $18$ $2.5$-second trajectories. The trajectories were used to generate a dataset $\mathcal{D} = \{t_{0:N}^{(i)},\mathbf\frakq_{0:N}^{(i)}, \bfzeta_{0:N}^{(i)}, \bfu^{(i)})\}_{i=1}^D$ with $N = 5$ and $D = 1080$. The $SE(3)$ Hamiltonian ODE network was trained, as described in Sec. \ref{sec:training}, for 1000 iterations with $\bfM_1(\mathbf\frakq)$ and $\bfM_2(\mathbf\frakq)$ pre-trained to output an identity matrix.

Our training and test results are shown in Fig. \ref{fig:pybullet_exp}. The learned generalized mass and inertia converged to constant diagonal matrices: $\bfM_1^{-1}(\mathbf\frakq) \approx 12.8\bfI$, $\bfM_2^{-1}(\mathbf\frakq) \approx \text{diag}([70, 70, 36])$. The input matrix $\bfg_{\bfv}(\mathbf\frakq)$ converged to a constant matrix whose entry $\begin{bmatrix}\bfg_{\bfv}(\mathbf\frakq)\end{bmatrix}_{2,0} \approx 2.8$ while other entries were closed to 0, consistent with the fact that the thrust only affects the linear velocity along the $z$ axis in the body-fixed frame. The input matrix $\bfg_{\bfomega}(\mathbf\frakq)$ converged to $\sim 380\bfI$ as the torques affects all components of the angular velocity $\bfomega$. The potential energy $V(\mathbf\frakq)$ was linear in the height $z$, agreed with the gravitational potential. 

We verified that predicted orientation trajectories from the learned model satisfy the $SO(3)$ constraints. Fig. \ref{fig:pybullet_so3_constraints} plots two near-zero quantities $|\det{\bfR} - 1|$ and $\Vert \bfR \bfR^\top - \bfI \Vert$, by rolling out our learned dynamics for $5$ seconds. We also calculated the Hamiltonian via \eqref{eq:hamiltonian_def} using the learned generalized mass matrix and the velocity rolled out from the learned dynamics. Fig. \ref{fig:pybullet_total_energy} shows a constant total energy along the $5$-second trajectory without control input, obeying the law of energy conservation.

Finally, we verified our energy-based controller for under-actuated systems in Sec. \ref{sec:controller_design} by driving the drone to track a pre-defined trajectory. We are given the desired position $\bfp^*$ and the desired heading $\bfpsi^*$ by the trajectory and construct an appropriate choice of $\bfR^*$, $\mathbf\frakp^*$ to be used with the energy-based controller in  \eqref{eq:idapbc_pose_twist_tracking}. The desired momenta are constructed as follows:
\begin{equation}
\begin{aligned}
\mathbf\frakp^* &= \bfM \begin{bmatrix} \bfR^\top \dot{\bfp}^* \\ \bfR^\top \bfR^* \bfomega^* \end{bmatrix} = \bfM \begin{bmatrix} \bfR^\top \bfR^* \bfv^* \\ \bfR^\top \bfR^* \bfomega^* \end{bmatrix},\\
\dot{\mathbf\frakp}^* &= \bfM \begin{bmatrix} \bfR^\top \ddot{\bfp}^* - \hat{\bfomega}\bfR^\top \dot{\bfp}^* \\
\bfR^\top \bfR^* \dot{\bfomega}^* - \hat{\bfomega}\bfR^\top\bfR^* \bfomega^* \end{bmatrix}.
\end{aligned}
\end{equation}
The control input \eqref{eq:idapbc_pose_twist_tracking} becomes:
\begin{eqnarray}
\label{eq:idapbc_quadrotor_control1}
\bfu &=& \bfg^{\dagger}(\mathbf\frakq)\left(\mathbf\frakq^{\times\top} \frac{\partial V(\mathbf\frakq)}{\partial \mathbf\frakq} - \mathbf\frakp^{\times}\bfM^{-1}\mathbf\frakp - \bfe(\mathbf\frakq,\mathbf\frakq^*) \right. \nonumber \\
&& \qquad\qquad\qquad - \left. \bfK_\bfd\bfM^{-1}(\mathbf\frakp-\mathbf\frakp^*) + \dot{\mathbf\frakp}^*\right).
\end{eqnarray}
By expanding the terms in  \eqref{eq:idapbc_quadrotor_control1}, we have: 
\begin{eqnarray}
\mathbf\frakp^{\times}\bfM^{-1}\mathbf\frakp &=& \mathbf\frakp^{\times} \bfzeta = \begin{bmatrix}
\hat{\mathbf\frakp}_\bfv \bfomega \\
\hat{\mathbf\frakp}_{\bfomega} \bfomega + \hat{\mathbf\frakp}_\bfv \bfv
\end{bmatrix}, \\
\bfM^{-1}(\mathbf\frakp-\mathbf\frakp^*) &=& \begin{bmatrix}
\bfv - \bfR^{\top}\dot{\bfp}^*  \\
\bfomega - \bfR^\top \bfR^* \bfomega^*
\end{bmatrix}, \\
\mathbf\frakq^{\times\top} \frac{\partial V(\mathbf\frakq)}{\partial \mathbf\frakq} &=& \begin{bmatrix}
\bfR^{\top} \frac{\partial V(\mathbf\frakq)}{\partial \bfp} \\
\sum_{i = 1}^3 \hat{\bfr}_i \frac{\partial V(\mathbf\frakq)}{\partial \bfr_i}
\end{bmatrix}.
\end{eqnarray}
Choosing the control gain $\bfK_\bfd$ of the form $\bfK_\bfd = \begin{bmatrix}
\bfK_v & \bf0\\
\bf0 & \bfK_\omega
\end{bmatrix}$, the control input can be written explicitly as
\begin{equation}
\label{eq:control_quadrotor}
\bfu = \bfg^{\dagger}(\mathbf\frakq) \begin{bmatrix}
\bfb_\bfv \\
\bfb_{\bfomega}
\end{bmatrix},
\end{equation}
where 
\begin{eqnarray}
\bfb_\bfv &=& \bfR^{\top} \frac{\partial V(\mathbf\frakq)}{\partial \bfp}  - \hat{\mathbf\frakp}_\bfv \bfomega - \bfR^\top\bfK_\bfp(\bfp - \bfp^*) \nonumber \\
&& - \bfK_\bfv(\bfv - \bfR^{\top}\dot{\bfp}^*)  + \bfM_1 (\bfR^\top \ddot{\bfp}^* - \hat{\bfomega}\bfR^\top \dot{\bfp}^*),  \\
\bfb_{\bfomega} &=& \sum_{i = 1}^3 \hat{\bfr}_i \frac{\partial V(\mathbf\frakq)}{\partial \bfr_i}- \bfK_{\bfomega}(\bfomega - \bfR^\top \bfR^* \bfomega^*)  \nonumber \\
&&   - (\hat{\mathbf\frakp}_{\bfomega} \bfomega + \hat{\mathbf\frakp}_\bfv \bfv) -\frac{1}{2}\prl{\bfK_{\bfR}\bfR^{*\top}\bfR-\bfR^\top\bfR^{*}\bfK_{\bfR}^\top}^{\vee}   \nonumber \\
&& + \bfM_2(\bfR^\top \bfR^* \dot{\bfomega}^* - \hat{\bfomega}\bfR^\top\bfR^* \bfomega^*).
\end{eqnarray}
%
%When $V(\mathbf\frakq) = mgz, \bfM_1 = mI, \bfM_2 = \bfJ$, we have:
%\begin{eqnarray}
%\bfb_\bfp &=& \bfR^{\top} mg\bfe_3  + m\hat{\bfomega}(\bfv - \bfR^\top \dot{\bfp}^*)  - \bfR^\top\bfK_\bfp(\bfp - \bfp^*) \nonumber \\
%&& - \bfK_d(\bfv - \bfR^{\top}\dot{\bfp}^*) + m\bfR^\top \ddot{\bfp}^* \nonumber  \\
%\bfb_\bfR &=& \bfomega\times \bfJ \bfomega - \frac{1}{2}\prl{\bfK_{\bfR}\bfR^{*\top}\bfR-\bfR^\top\bfR^{*}\bfK_{\bfR}^\top}^{\vee} \nonumber \\
%&&- \bfK_d(\bfomega - \bfR^\top \bfR^* \bfomega^*)  - \bfJ(\hat{\bfomega}\bfR^\top\bfR^* \bfomega^* - \bfR^\top \bfR^* \dot{\bfomega}^*)\nonumber
%\end{eqnarray}
%
Note that $\bfb_\bfv \in \mathbb{R}^3$ is the desired thrust in the body frame that depend only on the desired position $\bfp^*$ and the current pose. It is transformed to the world frame as $\bfR\bfb_\bfv$, representing the thrust in the world frame. Inspired by \cite{lee2010geometric}, the vector  $\bfR\bfb_\bfv$ should be the $z$ axis of the body frame, i.e., the third column $\bfb_3^*$ of the desired rotation matrix $\bfR^*$. The second column $\bfb^*_2$ of the desired rotation matrix $\bfR^*$ can be chosen so that it has the desired yaw angle $\psi^*$ and is perpendicular to $\bfb_3^*$. This can be done by projecting the second column of the yaw's rotation matrix $\bfb_2^\psi = [-\sin{\psi}, \cos{\psi}, 0]$ onto the plane perpendicular to $\bfb_3^*$. We have $\bfR^* = [\bfb_1^* \quad \bfb_2^* \quad \bfb_3^*]$ where:
\begin{equation}
\bfb_3^* = \frac{\bfR\bfb_\bfv}{\Vert \bfR\bfb_\bfv \Vert}, \bfb_1^* = \frac{\bfb_2^\psi \times \bfb^*_3}{\Vert  \bfb_2^\psi \times \bfb^*_3 \Vert}, \bfb_2^* = \bfb_3^* \times \bfb_1^*,
\end{equation}
and $\hat{\bfomega}^* = \bfR^{*\top}\dot{\bfR}^*$. The derivative $\dot{\bfR}^*$ is calculated as follows.
\begin{eqnarray}
\dot{\bfb}_3^* &=& \bfb_3^* \times \frac{\dot{\bfR\bfb_\bfv}}{\Vert \bfR\bfb_\bfv \Vert} \times \bfb_3^*,   \\
\dot{\bfb}_1^* &=& \bfb_1^* \times \frac{\dot{\bfb}_2^\psi \times \bfb_3^* + \bfb_2^\psi \times \dot{\bfb}_3^*}{\Vert \bfb_2^\psi \times \bfb_3^* \Vert} \times \bfb_1^*,  \\
\dot{\bfb}_2^* &=& \dot{\bfb}_3^* \times \bfb_1^* + \bfb_3^* \times \dot{\bfb}_1^*.
\end{eqnarray} 
By plugging $\bfR^*$ and $\bfomega^*$ back in $\bfb_{\bfomega}$, we obtain the complete control input $\bfu$ in  \eqref{eq:control_quadrotor}.

Fig. \ref{fig:pybullet_trajviz} qualitatively shows that the drone controlled by our energy-based controller successfully finished the task. Since the learned generalized mass $\bfM_1$ and inertia $\bfM_2$ converged to constant diagonal matrices, the control gains were chosen as follows: $\bfK_\bfp = 5\bfM_1, \bfK_\bfv = 2.5\bfM_1, \bfK_{\bfR} = 250\bfM_2, \bfK_{\bfomega} = 20\bfM_2$. Fig. \ref{fig:pybullet_exp} quantitatively plots the tracking errors for position, yaw angles, linear and angular velocity. Our controller's computation time was $3.5ms$ per control input, including forward passes of the learned neural networks, showing that it is suitable for fast real-time applications.

%The generalized mass inverse $\bfM_1^{-1}(\mathbf\frakq) \in\mathbb{R}^{3\times 3}$, inertia $\bfM_2^{-1}(\mathbf\frakq)\in\mathbb{R}^{3\times 3}$, the control input matrices $\bfg_{\bfv}(\mathbf\frakq)\in\mathbb{R}^{3\times 4}$, $\bfg_{\bfomega}(\mathbf\frakq) \in\mathbb{R}^{3\times 4}$  in in Fi.g \ref{fig:pybullet_M1_x_all}, \ref{fig:pybullet_M2_x_all}, \ref{fig:pybullet_g_v_x_all}, and \ref{fig:pybullet_g_w_x_all}, respectively.

%% file: tex/Conclusion.tex
\section{Conclusion}
\label{sec:conclusion}

This paper proposes a neural ODE network design for rigid body dynamics learning that captures $SE(3)$ kinematics and Hamiltonian dynamics constraints by construction. It also developed a general control approach for under-actuated trajectory tracking based on the learned $SE(3)$ Hamiltonian dynamics. The learning and control designs are not system-specific and thus can be applied to any mobile robot, whose state evolves on $SE(3)$. These techniques have the potential to enable mobile robots to adapt their models online, in response to changing operational conditions or structural damage, and, yet, maintain stability and autonomous operation. Future work will focus on comparison with other dynamics learning approaches, extending our formulation to allow learning the kinematic and dynamic structure of multi-rigid body systems and provide safe and stable adaptive control, in the presence of noise and disturbances.

%% file: tex/Appendix.tex
\section{Appendix}
\label{sec:appendix}
\subsection{Implementation Details}
\label{subsec:implement_details}

We used fully-connected neural networks whose architecture is shown below. The first number is the input dimension while the last number is the output dimension. The numbers in between are the hidden layers' dimensions and activation functions. The value of $\varepsilon$ in \eqref{eq:M_cholesky} is set to $0.01$.
\begin{enumerate}
	\item Pendulum:
	\begin{itemize}
		\item Input dimension: $9$. Action dimension: $1$.
		\item $\bfL(\mathbf\frakq)$:\\ 9 - 300 Tanh - 300 Tanh - 300 Tanh - 300 Linear - 6.
		\item $V(\mathbf\frakq)$: 9 - 50 Tanh - 50 Tanh - 50 Linear - 1.
		\item $\bfg(\mathbf\frakq)$: 9 - 300 Tanh - 300 Tanh - 300 Linear - 3.
	\end{itemize}
	\item Rigid body:
		\begin{itemize}
		\item Input dimension: $12$. Action dimension: $6$.
		\item $\bfL_1(\mathbf\frakq)$ only takes the position $\bfp \in \mathbb{R}^3$ as input:\\ 3 - 400 Tanh - 400 Tanh - 400 Tanh - 400 Linear - 6.
		\item $\bfL_2(\mathbf\frakq)$ only takes the rotation matrix $\bfR \in \mathbb{R}^{3\times 3}$ as input:\\ 9 - 400 Tanh - 400 Tanh - 400 Tanh - 400 Linear - 6.
		\item $V(\mathbf\frakq)$: 12 - 400 Tanh - 400 Tanh - 400 Linear - 1.
		\item $\bfg(\mathbf\frakq)$: 12 - 400 Tanh - 400 Tanh - 400 Linear - 36.
	\end{itemize}
	\item Quadrotor:
		\begin{itemize}
		\item Input dimension: $12$. Action dimension: $4$.
		\item $\bfL_1(\mathbf\frakq)$ only takes the position $\bfp \in \mathbb{R}^3$ as input:\\ 3 - 400 Tanh - 400 Tanh - 400 Tanh - 400 Linear - 6.
		\item $\bfL_2(\mathbf\frakq)$ only takes the rotation matrix $\bfR \in \mathbb{R}^{3\times 3}$ as input:\\ 9 - 400 Tanh - 400 Tanh - 400 Tanh - 400 Linear - 6.
		\item $V(\mathbf\frakq)$: 12 - 400 Tanh - 400 Tanh - 400 Linear - 1.
		\item $\bfg(\mathbf\frakq)$: 12 - 400 Tanh - 400 Tanh - 400 Linear - 24.
	\end{itemize}
\end{enumerate}